\documentclass[10pt,twocolumn,letterpaper]{article}

\usepackage{cvpr}              %

\usepackage{array}
\usepackage{algorithm}
\usepackage{algpseudocode}
\usepackage{amsmath}
\usepackage{booktabs}
\usepackage{tabularx}
\usepackage{adjustbox}
\usepackage{multirow}
\usepackage{graphicx}
\usepackage{float}
\usepackage[table]{xcolor}
\usepackage{colortbl}
\usepackage{makecell}
\usepackage{caption}
\usepackage{relsize}
\usepackage{marvosym}
\usepackage{arydshln}

\definecolor{lightyellow}{RGB}{255,255,200}
\definecolor{lightorange}{RGB}{255,230,200}
\definecolor{lightred}{RGB}{255,200,200}

\newcommand{\tbest}[1]{\cellcolor{lightyellow}#1}
\newcommand{\sbest}[1]{\cellcolor{lightorange}#1}
\newcommand{\best}[1]{\cellcolor{lightred}#1}

\newcommand{\method}{Gen3R}
\newcommand{\boldparagraph}[1]{\vspace{0.2cm}\noindent{\bf #1.} }

\newcolumntype{P}[1]{>{\centering\arraybackslash}m{#1}}

\definecolor{cvprblue}{rgb}{0.21,0.49,0.74}
\usepackage[pagebackref,breaklinks,colorlinks,allcolors=cvprblue]{hyperref}

\title{Gen3R: 3D Scene Generation Meets Feed-Forward Reconstruction}

\vspace{-0.25cm}

\author{
Jiaxin Huang$^{1}$,
Yuanbo Yang$^{1}$,
Bangbang Yang$^{2}$,
Lin Ma$^{2}$,
Yuewen Ma$^{2}$,
Yiyi Liao$^{1}\textsuperscript{\Letter}$
\vspace{0.05cm}
\\
{\normalsize $^{1}$ Zhejiang University\quad$^{2}$ ByteDance}
\\
\small{Project Page: \url{https://xdimlab.github.io/Gen3R/}}
}

\vspace{-0.1cm}

\begin{document}
\twocolumn[{%
\renewcommand\twocolumn[1][]{#1}%
\maketitle
\begin{center}
  {
  \vspace{-25pt}
  \captionsetup{type=figure}
  \centering
  \includegraphics[width=0.95\linewidth]{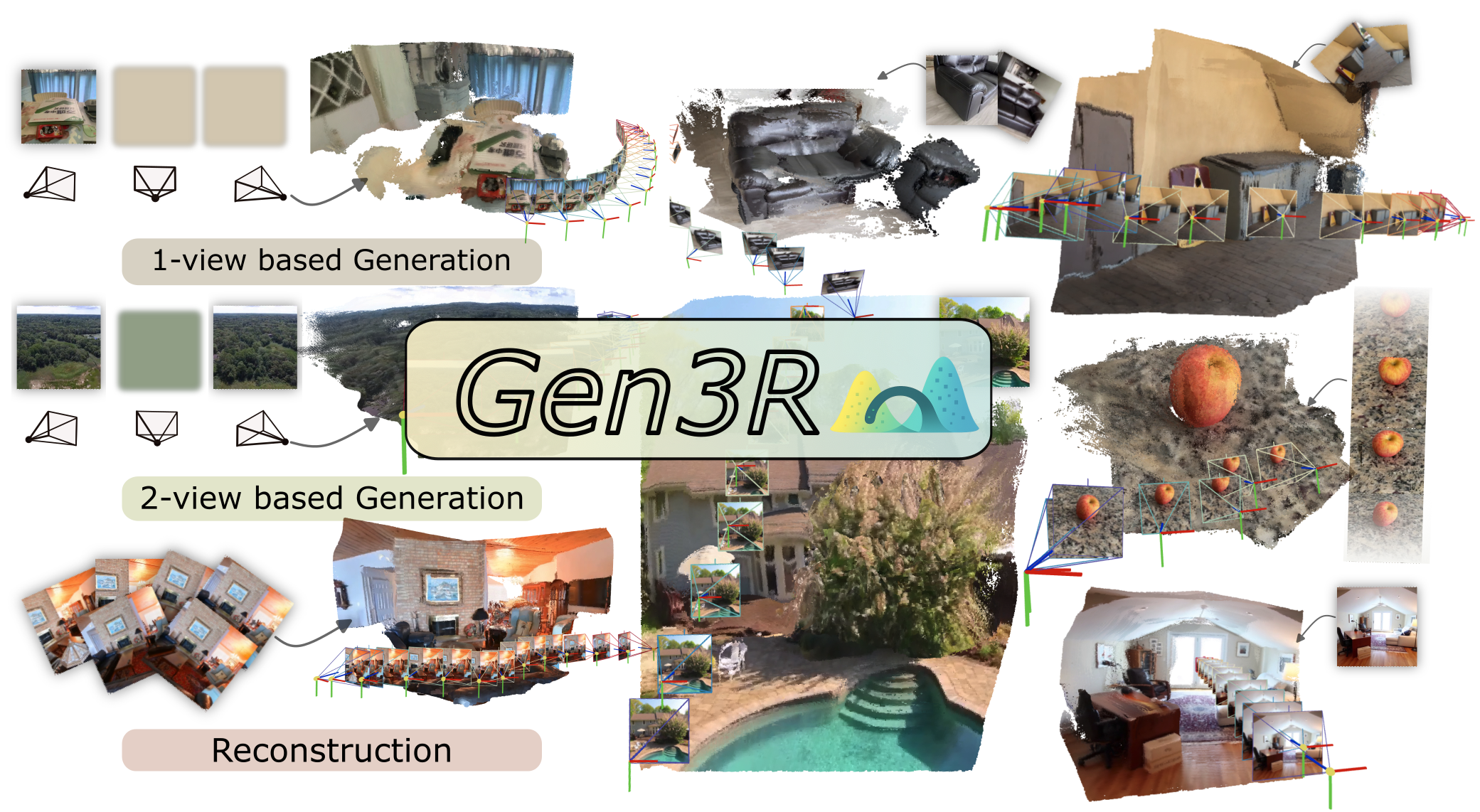}
  \vspace{-10pt}
  \captionof{figure}{
  \textbf{\method{}} bridges foundational reconstruction models with 2D video diffusion, enabling the joint generation of 2D videos and their corresponding geometry in various settings.
  }
  \label{fig:teaser}
}
\end{center}
}]

\renewcommand{\thefootnote}{}
\footnotetext[2]{$\textsuperscript{\Letter}$ Corresponding author.}
\renewcommand{\thefootnote}{\arabic{footnote}}

\begin{abstract}
We present \method, a method that bridges the strong priors of foundational reconstruction models and video diffusion models for scene-level 3D generation. We repurpose the VGGT reconstruction model to produce geometric latents by training an adapter on its tokens, which are regularized to align with the appearance latents of pre-trained video diffusion models. By jointly generating these disentangled yet aligned latents, \method{} produces both RGB videos and corresponding 3D geometry, including camera poses, depth maps, and global point clouds. Experiments demonstrate that our approach achieves state-of-the-art results in single- and multi-image conditioned 3D scene generation. Additionally, our method can enhance the robustness of reconstruction by leveraging generative priors, demonstrating the mutual benefit of tightly coupling reconstruction and generative models.
\end{abstract}

\section{Introduction}
\label{sec: introduction}

3D scene generation has become a fundamental problem in computer vision and graphics, with wide applications in simulation, gaming, robotics, and virtual reality. A method capable of producing photorealistic and geometrically consistent 3D scenes would enable the creation of immersive environments at scale, serving as essential training data and providing new tools for creative content design. 

Prior methods attempt to extend 2D generative models via score distillation~\cite{prolificdreamer, dreamfusion, dreamgaussian, magic3d}, incremental outpainting~\cite{luciddreamer, wonderworld, wonderjourney, scenescape}, or multi-view synthesis followed by reconstruction~\cite{cat3d, syncdreamer, zeronvs, mvdream, reconfusion, gaussvideodreamer}. Despite promising results, these methods often suffer from poor geometric structure or high optimization cost.
More recently, several works~\cite{prometheus, bolt3d, splatflow, director3d, wvd, matrix3d} have extended video diffusion frameworks to \textit{feed-forward 3D scene generation} for improved efficiency. These approaches typically follow the Latent Diffusion Model paradigm, training a VAE to learn a compact latent space for 3D scenes and applying diffusion within that space. However, the scarcity of large-scale 3D ground truth makes learning geometry-centric VAEs highly challenging. One line of methods trains a VAE to reconstruct geometry from RGB inputs while simultaneously learning a compressed latent representation~\cite{wildfusion, prometheus, director3d}. Yet this is inherently difficult, especially when supervision is limited to 2D signals, which often results in suboptimal geometry and constrained generation quality.

In parallel, transformer-based feed-forward reconstruction models, such as Dust3R~\cite{dust3r} and VGGT~\cite{vggt}, have shown strong reconstruction ability from 2D images. 
Recent works attempt to build better VAEs by compressing their 3D output~\cite{wvd, bolt3d}, but overlook a key fact: these reconstruction models already operate in a spatially compact token space that encodes rich multi-view geometric information, including depth, camera pose, and global structure. This observation raises a central question: 
Can the intrinsic latent manifold learned by reconstruction models be used to fully exploit reconstruction priors for 3D scene generation?

Building on this insight, we introduce \method, a 3D-aware scene generation method that unifies advanced reconstruction and generation models for jointly generating controllable video and globally consistent 3D point clouds. Our key idea is to recast a feed-forward reconstruction model, VGGT~\cite{vggt}, as a VAE-like provider of geometric latents and combine these with appearance latents from a pre-trained video diffusion model for joint generation. This allows us to marry the rich geometric priors learned by reconstruction models over multiple 3D quantities with the strong RGB priors of video diffusion models, effectively combining the strengths of both.
To achieve so, we first project the reconstruction model's intermediate tokens to match the spatial-temporal resolution of the appearance latents using a learned adapter.
Notably, simply compressing the tokens is insufficient as their distribution significantly differs from the corresponding appearance latents. We therefore propose to align the two latent spaces, followed by fine-tuning a video diffusion model for joint generation. By keeping geometric and appearance latents disentangled while aligning their distributions, \method~demonstrates that the latent manifold learned by reconstruction models can indeed serve as a strong foundation for high-fidelity 3D scene generation.

Our framework supports flexible conditioning, enabling generation from single or multiple input views, with or without camera cues, as well as feed-forward scene reconstruction within one unified model. It produces temporally coherent RGB videos and globally aligned point clouds across diverse configurations.

Our contributions are threefold:
1) A novel framework integrating video diffusion models with geometric foundation models, combining strong RGB priors with rich geometric priors for 3D scene generation.
2) A disentangled yet aligned appearance and geometry latent space, enabling controllable and multi-view consistent scene synthesis.
3) A flexible pipeline capable of handling various input settings, producing high-fidelity videos and globally consistent 3D point clouds.

\section{Related Work}
\label{sec: related_work}

\boldparagraph{3D Scene Generation fom 2D Priors}
A common strategy for 3D scene generation is to leverage pretrained 2D generative models~\cite{sd} to provide RGB priors. One line of work employs score distillation sampling~(SDS)~\cite{dreamfusion, magic3d, prolificdreamer, dreamgaussian}, directly optimizing a 3D representation such as NeRF~\cite{nerf, mipnerf, mipnerf360, zipnerf} and  3DGS~\cite{3dgs, 3dgsreview, rtg_slam, ges} to align with the distribution of a 2D diffusion model. Another line of methods first synthesizes multi-view images using pretrained 2D diffusion models, followed by 3D reconstruction through multi-view synthesis~\cite{cat3d, syncdreamer, zeronvs, mvdream, reconfusion, gaussvideodreamer, dimensionx, reconx, mvsplat360, genxd} or incremental outpainting~\cite{scenescape, luciddreamer, wonderjourney, wonderworld,3drecipe}. 
Both paradigms leverage the strong RGB priors of 2D models but are limited by the lack of explicit 3D reasoning, often resulting in inconsistent geometry, weak multi-view fidelity, and high computational cost. Our method tackles this challenge by bridging rich geometric priors of a reconstruction foundation model with a 2D generative model.

\begin{figure*}[t]
    \centering
    \includegraphics[width=\textwidth]{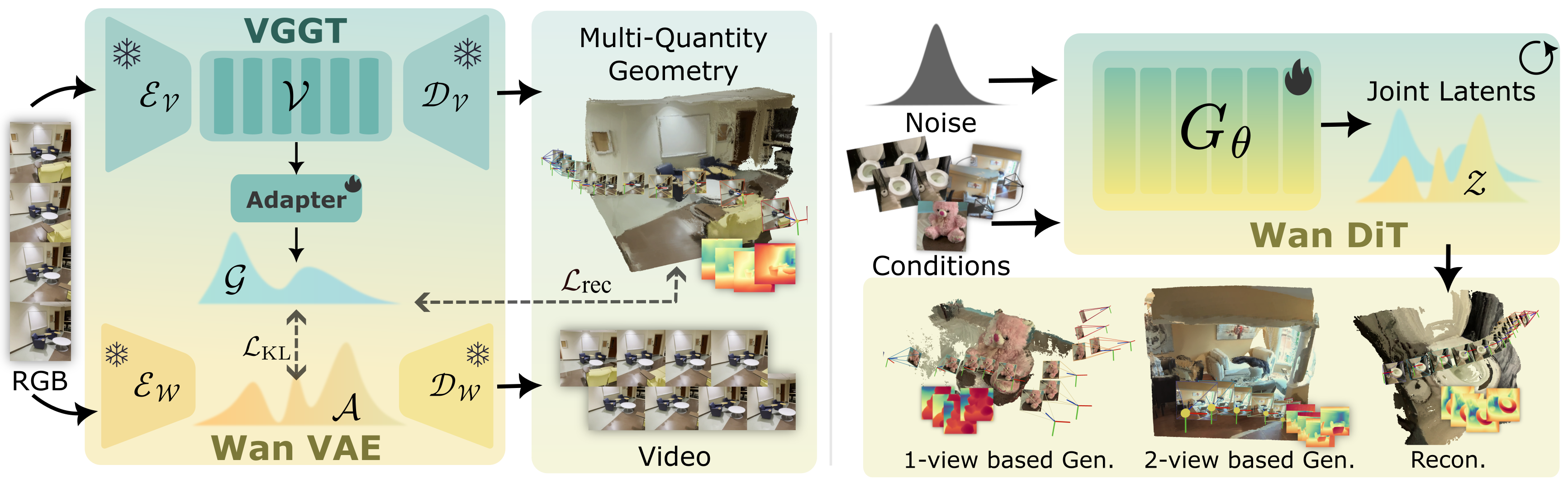}
    \vspace{-0.65cm}
    \caption{\textbf{Method.} \textit{Left:} We recast an advanced transformer-based feed-forward reconstruction model, VGGT, as a VAE to produce geometry latents $\mathcal{G}$ by training an adapter on its latent tokens. The training is supervised with a reconstruction loss $\mathcal{L}_{\mathrm{rec}}$, along with a regularization term $\mathcal{L}_{\mathrm{KL}}$ that aligns $\mathcal{G}$ with the appearance latent $\mathcal{A}$, which is obtained from the VAE of a pre-trained video diffusion model, WAN.
    \textit{Right:}  We fine-tune the video diffusion model to jointly generate geometry and appearance latents, $\mathcal{Z}=[\mathcal{A}; \mathcal{G}]$, under various conditioning signals. At inference, varying the conditioning enables the generation of RGB videos and multiple 3D quantities, including global point clouds, depth maps, and camera parameters, from a single or multiple frames, as well as performing reconstruction.}
   \label{fig: pipeline}
   \vspace{-0.45cm}
\end{figure*}

\boldparagraph{Feed-Forward 3D Scene Generation} 
Object-level feed-forward 3D generation methods~\cite{trellis, zero12g, l3dg, diffsplat, clay} have gained great success thanks to the large-scale 3D ground truth datasets~\cite{objaverse, objaversepp}. However, extending this success to full-scene 3D generation is challenging because high-quality scene-level data is difficult to obtain. A practical alternative is to synthesize a 3D representation in a feed-forward manner and train it using only 2D supervision. Recent works~\cite{diffrf, splatflow, director3d, prometheus, ggs, latentsplat, wonderland} follow this strategy by generating Gaussians and using differentiable rendering to train directly on 2D images, thereby avoiding costly 3D data collection. However, these methods often struggle with intricate geometric details and multi-view consistency due to the lack of explicit 3D supervision.  Other approaches~\cite{wvd, aether, bolt3d, housecrafter} address this limitation by leveraging off-the-shelf dense reconstruction models~\cite{dust3r, mast3r} or Unreal Engine to obtain 3D data for training.
In contrast to methods directly compressing the 3D output of reconstruction models~\cite{wvd, aether},
our approach treats the reconstruction model as an asymmetric VAE that encodes images into geometry latents, %
allowing us to inherit the strong geometric priors across multiple 3D quantities and the high-level scene understanding from the foundation geometry model.

\boldparagraph{Feed-forward 3D Scene Reconstruction}
Traditional 3D scene reconstruction pipelines~\cite{sfm, orbslam, orbslam3, droidslam} have recently been complemented by learning-based methods~\cite{dust3r, mast3r, cut3r, monst3r, dens3r, lrm, gslrm, lgm, vggsfm}, which leverage neural architectures to capture structural regularities of the world. Among these, seminal works such as~\cite{dust3r, mast3r} demonstrated the ability to infer geometrically consistent point clouds from uncalibrated images. Recent advances, exemplified by~\cite{vggt, mapanything}, provide unified frameworks that jointly estimate camera parameters, dense geometry and point tracks. Subsequent studies have further extended VGGT to new scene representations~\cite{anysplat, 4realvideo2, iggt, vggtx} or addressed its inherent limitations~\cite{pi3, fastvggt, streamvggt}.

Our method integrates the geometric prior of such feed-forward reconstruction models~\cite{vggt} with a generative diffusion model~\cite{wan, cogvideo, cogvideox}. Different from prior reconstruction approaches~\cite{vggt, anysplat, fastvggt, pi3, dust3r}, our method is inherently generative, capable of synthesizing coherent 3D scenes from 1 or 2 views. 
Moreover, our method can also be used for performing reconstruction and is able to mitigate errors of the original reconstruction model. 

\section{Method}
\label{sec: method}

Our goal is to generate high-fidelity 3D scenes with consistent geometry and controllable cameras given one or more images. To achieve this, we propose \method, a 3D-aware latent diffusion method bridging foundational reconstruction models with pre-trained video diffusion models.

Specifically, we first design a unified latent space for appearance and geometry by recasting the geometry features of the feed-forward reconstruction model, VGGT, into the latent space of a video diffusion model~(\cref{sec: latent_space}).
We then fine-tune the video diffusion model to jointly generate the appearance and geometry latents under various conditions~(\cref{sec: dit}). Finally, these latents are decoded separately into RGB frames and scene geometry, including global point clouds, depth maps and camera parameters
~(\cref{sec: decoding_and_merging}). \cref{fig: pipeline} illustrates our overall architecture.

\begin{figure*}[htbp]
    \centering
    \newcommand{\imgwidth}{0.132\textwidth}
    \def\mywidth{2.2cm}
    
    \setlength{\tabcolsep}{3pt}
    \setlength\dashlinedash{1.5pt}
    \setlength\dashlinegap{1pt}
    \setlength\arrayrulewidth{1.5pt}

    \begin{tabular}{P{0.4cm}P{\mywidth}P{\mywidth}P{\mywidth}P{\mywidth}P{\mywidth}P{\mywidth}P{\mywidth}}
        \rotatebox{90}{Input} &
        \includegraphics[width=0.08\textwidth]{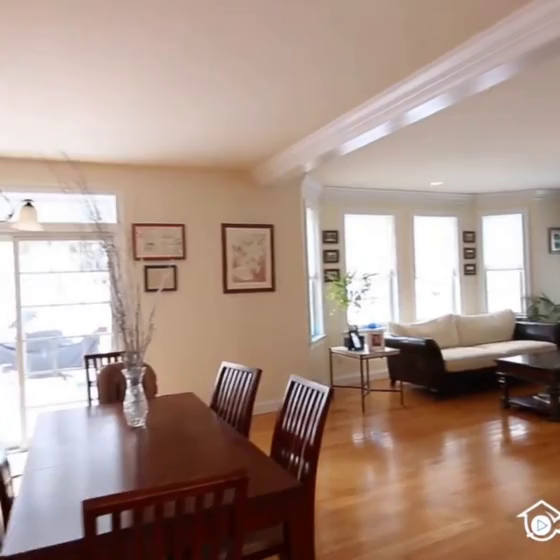} &
        \includegraphics[width=0.08\textwidth]{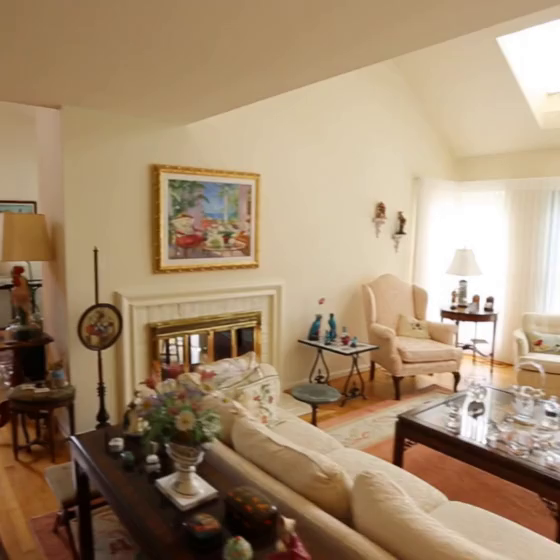} &
        \includegraphics[width=0.08\textwidth]{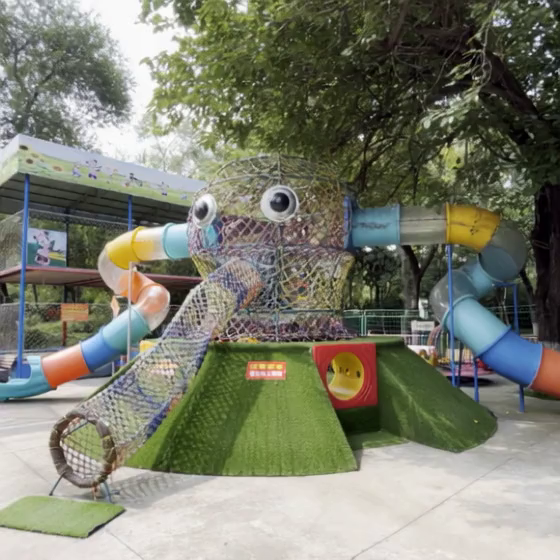} &
        \includegraphics[width=0.08\textwidth]{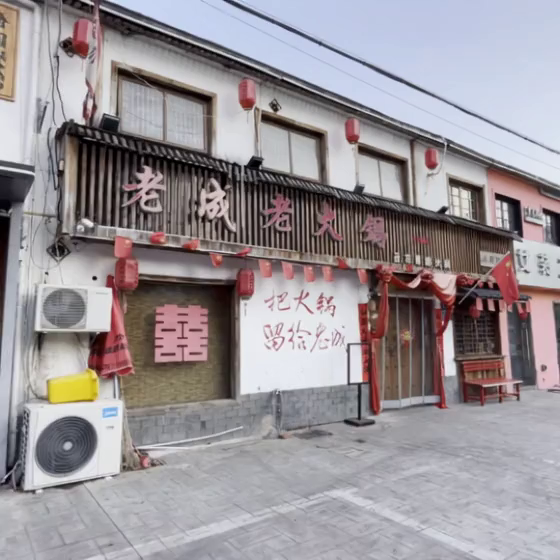} &
        \includegraphics[width=0.08\textwidth]{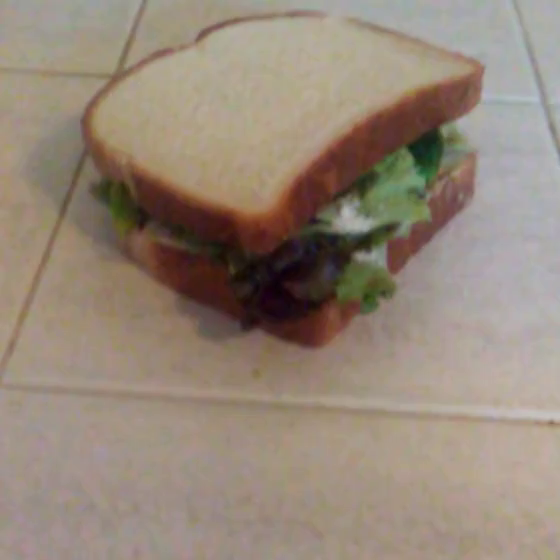} &
        \includegraphics[width=0.08\textwidth]{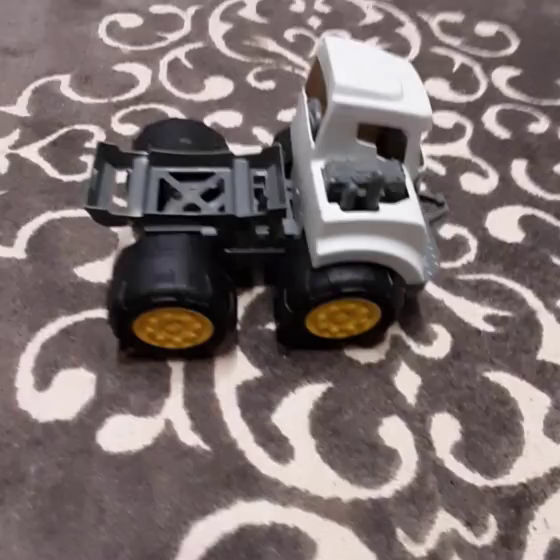} &
        \includegraphics[width=0.08\textwidth]{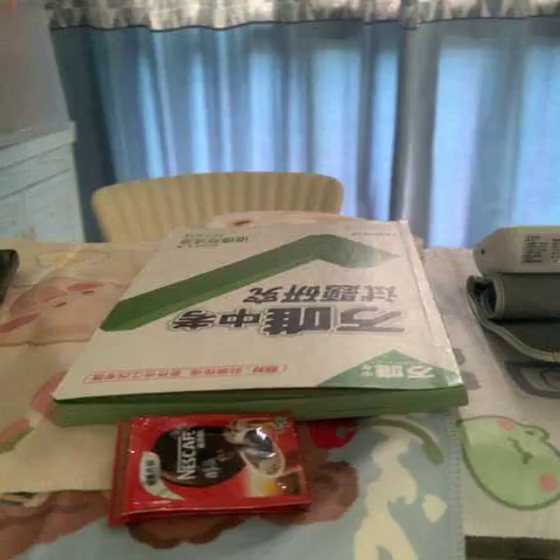} \\ [-4pt]
        \rotatebox{90}{Aether~\cite{aether}} &
        \raisebox{0.2\height}{\includegraphics[width=\imgwidth]{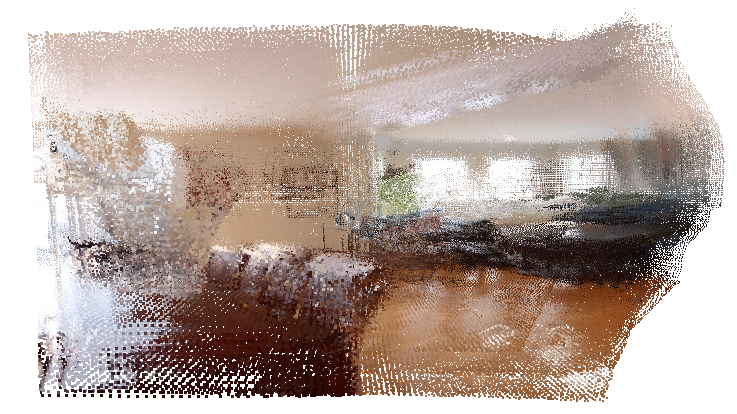}} &
        \raisebox{0.2\height}{\includegraphics[width=\imgwidth]{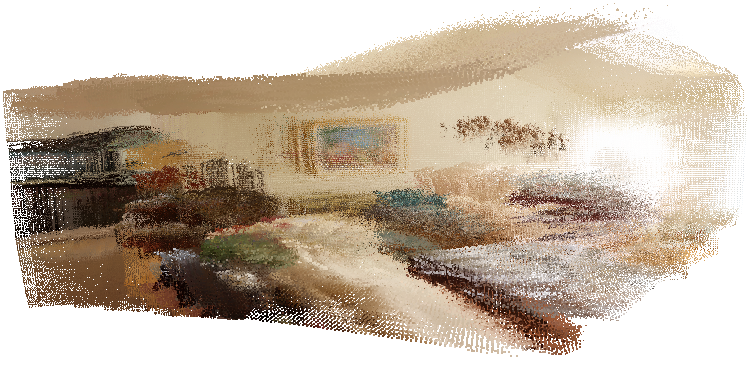}} &
        \raisebox{0.25\height}{\includegraphics[width=\imgwidth]{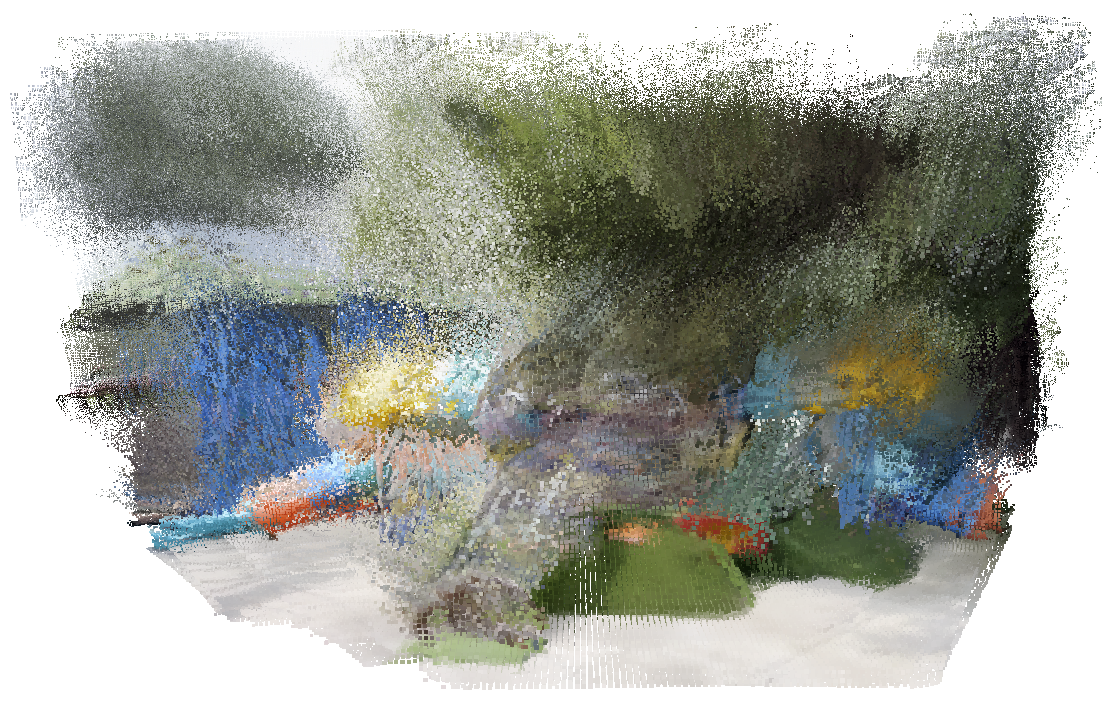}} &
        \raisebox{0.15\height}{\includegraphics[width=\imgwidth]{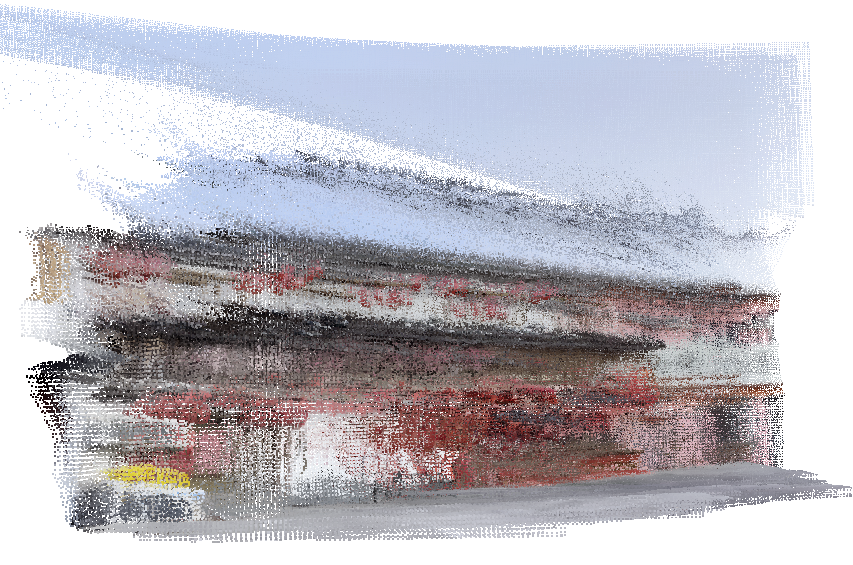}} &
        \raisebox{0.1\height}{\includegraphics[width=\imgwidth]{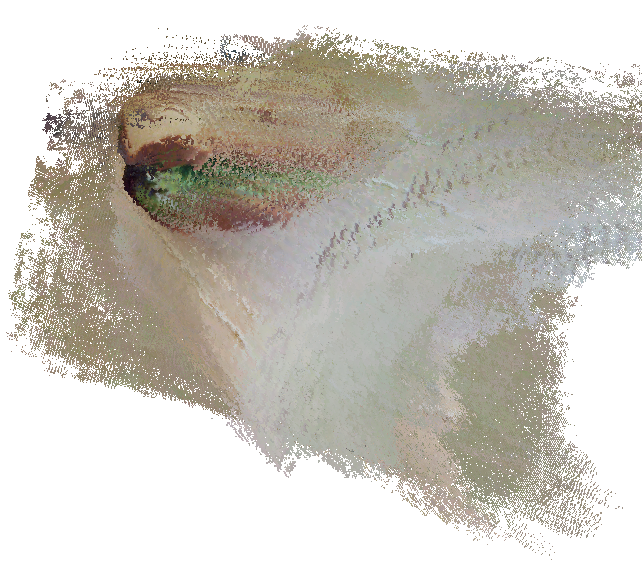}} &
        \raisebox{0.05\height}{\includegraphics[width=\imgwidth]{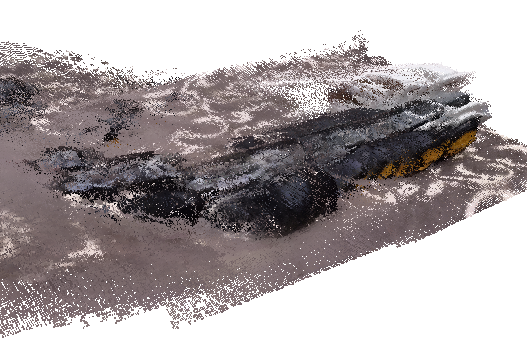}} &
        \includegraphics[width=\imgwidth]{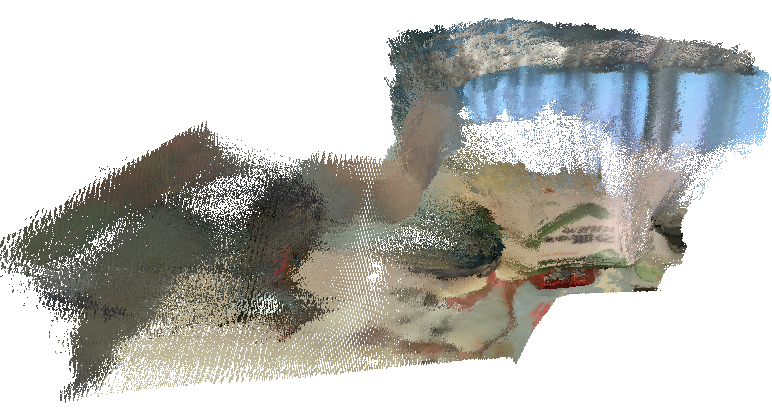} \\ [-20pt]
        \rotatebox{90}{WVD~\cite{wvd}} &
        \raisebox{0.1\height}{\includegraphics[width=\imgwidth]{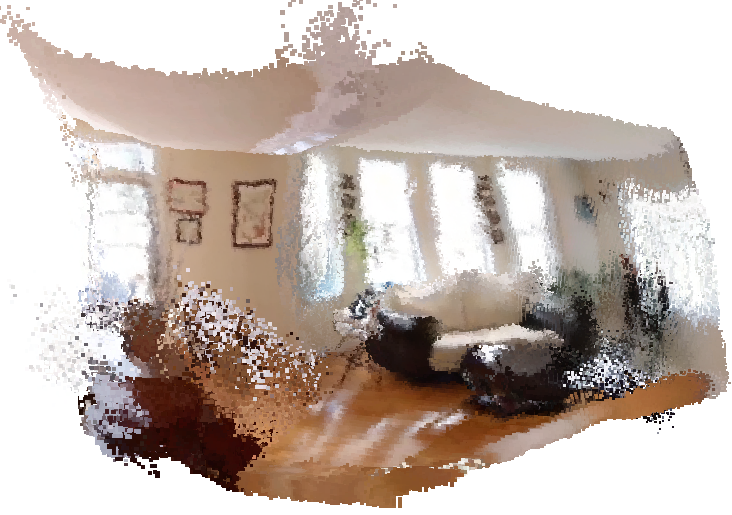}} &
        \raisebox{0.1\height}{\includegraphics[width=\imgwidth]{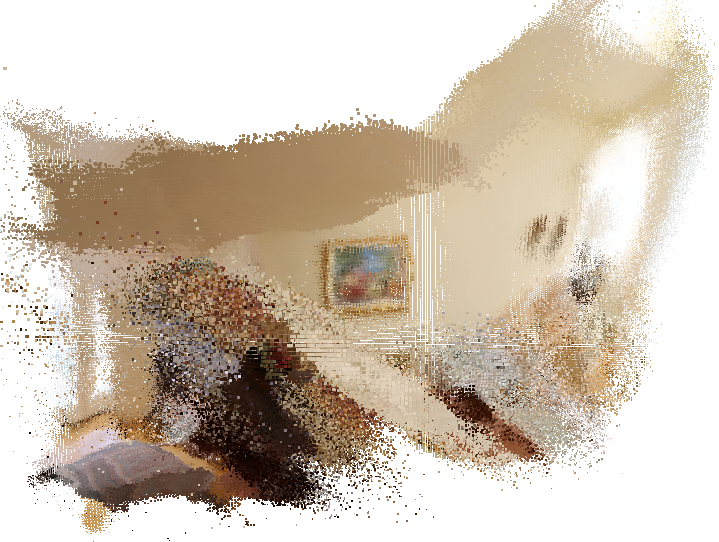}} &
        \raisebox{0.15\height}{\includegraphics[width=\imgwidth]{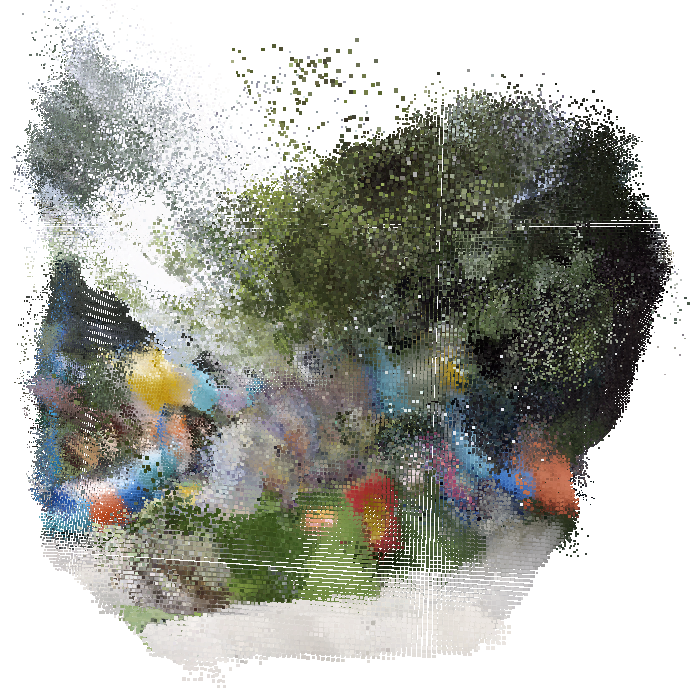}} &
        \includegraphics[width=\imgwidth]{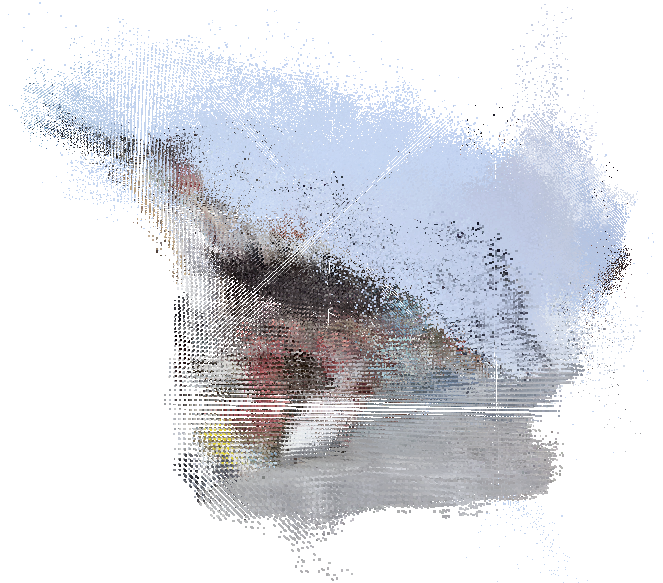} &
        \raisebox{0.1\height}{\includegraphics[width=\imgwidth]{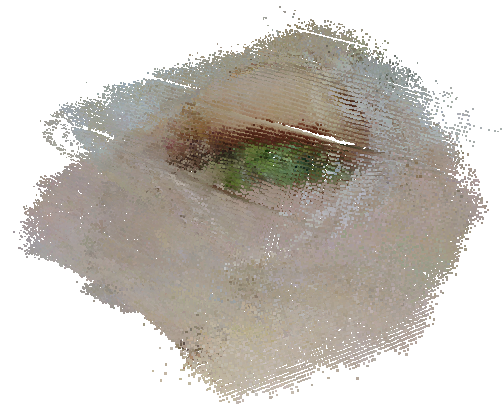}} &
        \raisebox{0.1\height}{\includegraphics[width=\imgwidth]{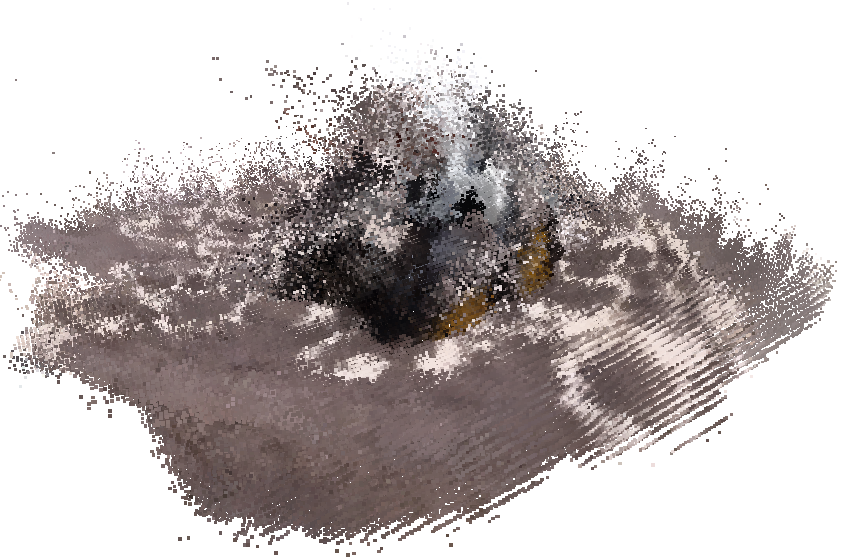}} &
        \includegraphics[width=\imgwidth]{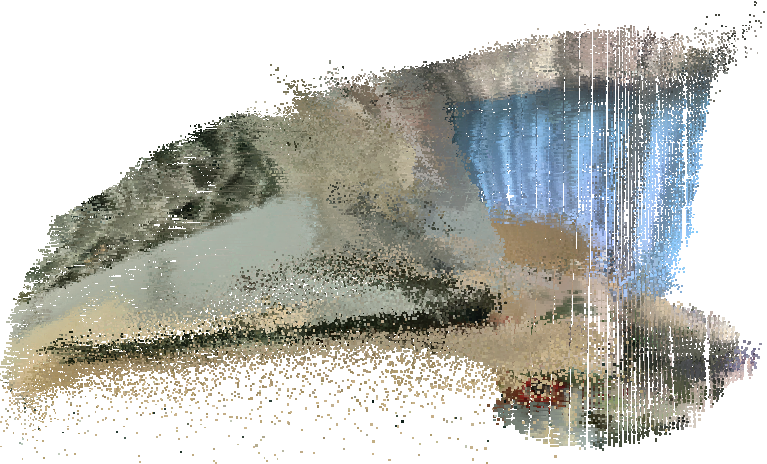} \\ [-15pt]
        \rotatebox{90}{Ours} &
        \raisebox{0.1\height}{\includegraphics[width=0.11\textwidth, height=0.105\textwidth]{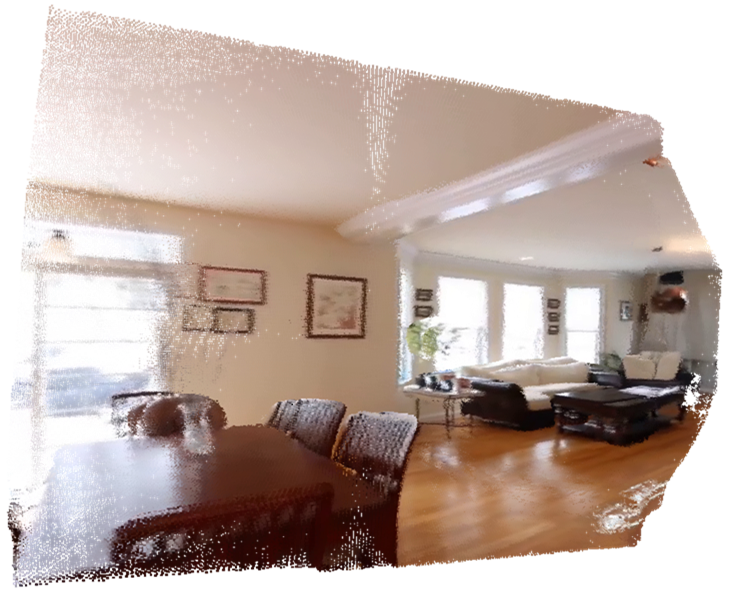}} &
        \raisebox{0.1\height}{\includegraphics[width=\imgwidth]{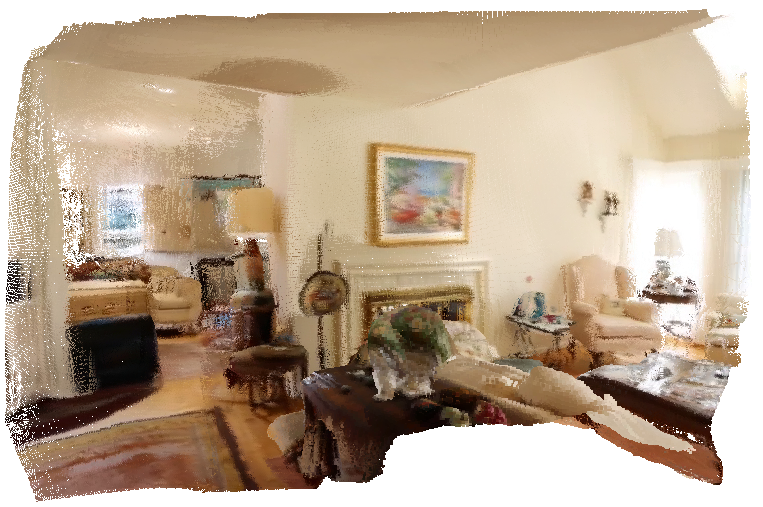}} &
        \raisebox{0.2\height}{\includegraphics[width=\imgwidth]{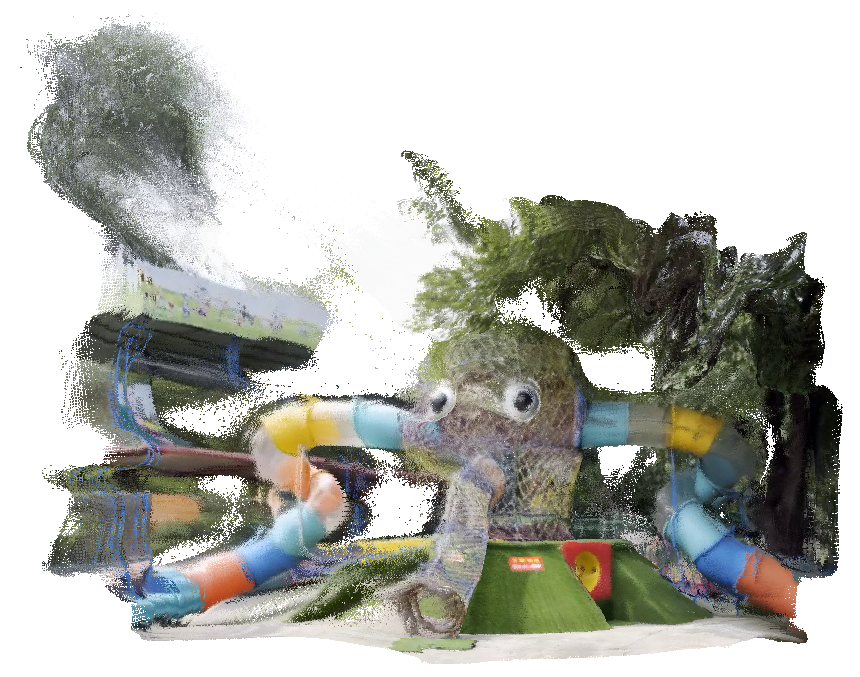}} &
        \includegraphics[width=\imgwidth]{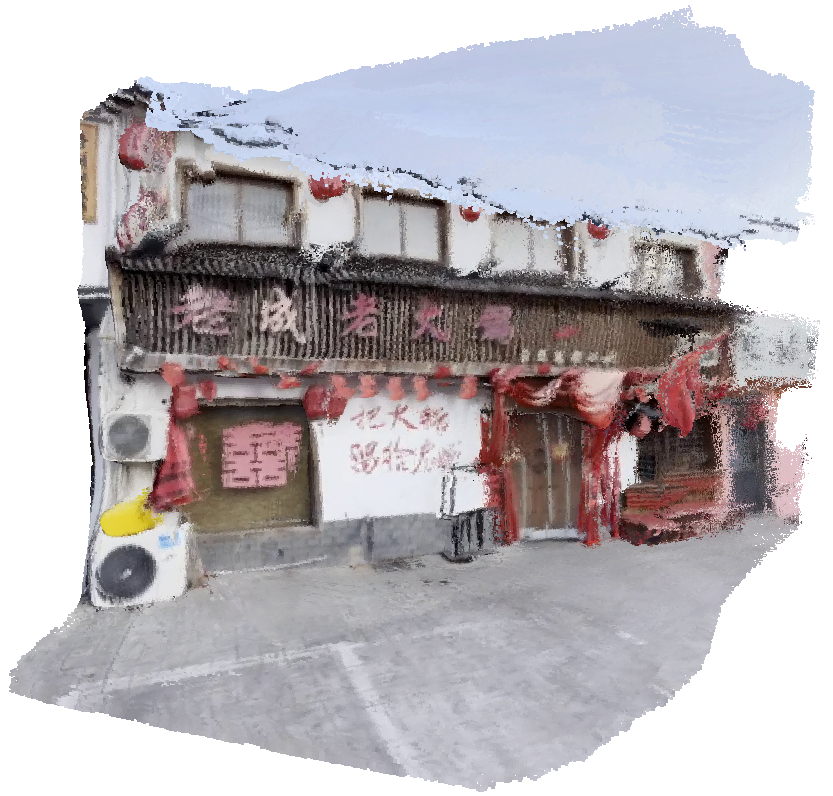} &
        \raisebox{0.1\height}{\includegraphics[width=\imgwidth]{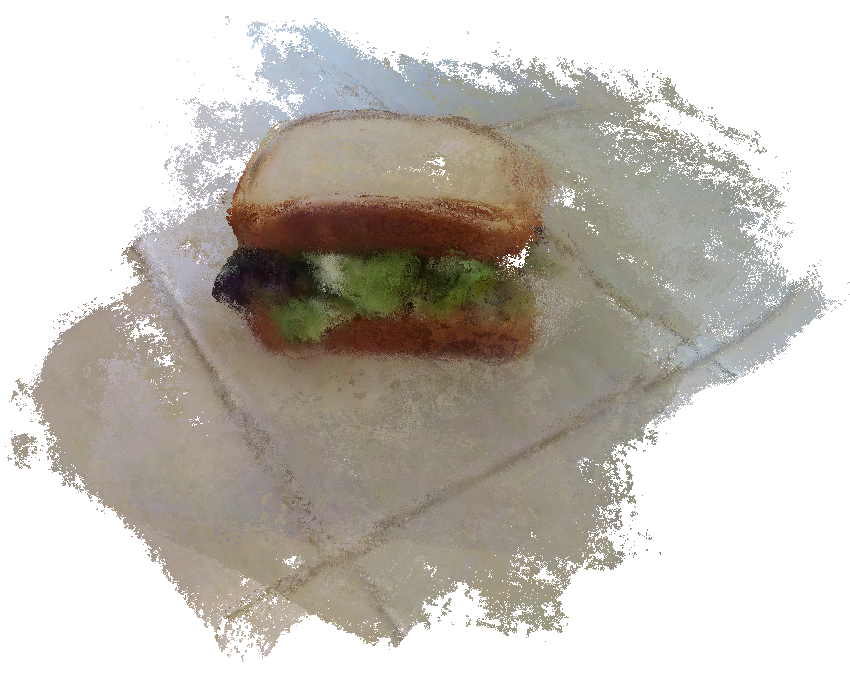}} &
        \raisebox{0.05\height}{\includegraphics[width=\imgwidth]{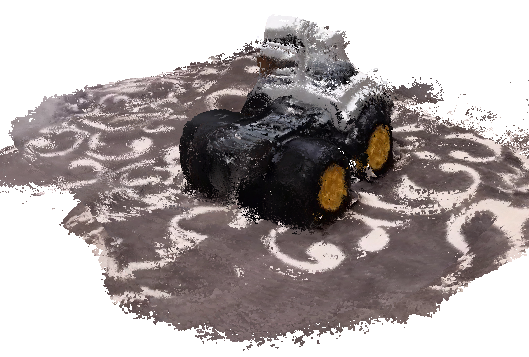}} &
        \raisebox{-0.5\height}{\includegraphics[width=\imgwidth]{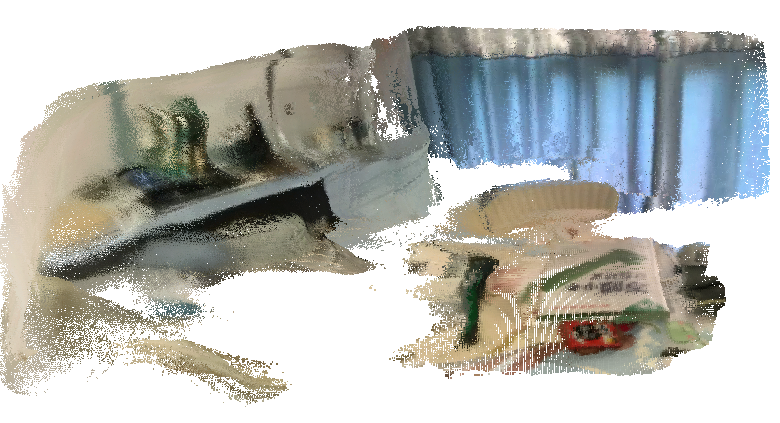}} \\ [2pt]
    \end{tabular}
    \vspace{-0.45cm}
    \caption{\textbf{Qualitative Comparison of Geometry Generation} in the 1-view based setting.}
    \label{fig: single_to_3d_rgb_transposed}
    \vspace{-0.55cm}
\end{figure*}

\subsection{Geometry Adapter for Unified Latent Space}
\label{sec: latent_space}

\boldparagraph{Preliminary} 
VGGT~\cite{vggt}, denoted as \(\mathcal{F}\), is a transformer-based feed-forward reconstruction model that directly infers multiple key 3D quantities of a scene from observed views. It takes \(N\) input images \(\mathcal{I} \in \mathbb{R}^{N \times H \times W \times 3}\) and encodes them into high-dimensional geometry tokens \(\mathcal{V} \in \mathbb{R}^{N \times h_v \times w_v \times C}\) by its encoder \(\mathcal{E}_\mathcal{V}\), which consists of \(24\) attention blocks:
\begin{equation}
    \mathcal{E}_\mathcal{V}: \mathcal{I} \rightarrow \mathcal{V} \in \mathbb{R}^{N \times L \times h_v \times w_v \times C},
\end{equation}
where \(h_v\times w_v\) is the token's spatial resolution, \(C=2048\), and \(L=4\) is the number of intermediate transformer tokens---specifically those from the \(4\)th, \(11\)th, \(17\)th and \(23\)rd blocks~\cite{depthanythingv2}---for subsequent decoding. For simplicity, we omit the camera tokens of VGGT in the description; please refer to Supplementary for details.

The geometry tokens \(\mathcal{V}\) are then decoded by several individual DPT heads~\cite{dpt} \(\mathcal{D}_\mathcal{V}\) into multi-modal dense predictions, such as point clouds \(\mathcal{P} \in \mathbb{R}^{N \times H \times W \times 3}\), depth maps \(\mathcal{D} \in \mathbb{R}^{N \times H \times W \times 1}\) and camera parameters \(\mathcal{T} \in \mathbb{R}^{N \times 9}\):
\begin{equation}
    \mathcal{D}_\mathcal{V}: \mathcal{V} \rightarrow (\mathcal{P}, \mathcal{D}, \mathcal{T}).
    \vspace{-0.2cm}
\end{equation}

\boldparagraph{Token-to-Latent Adapter} We aim to recast VGGT~\cite{vggt} as an asymmetric geometry VAE that takes as input \(N\) RGB images \(\mathcal{I} \in \mathbb{R}^{N \times H \times W \times 3}\), produces geometric latents \(\mathcal{G} \in \mathbb{R}^{n \times h \times w \times c}\) for diffusion-based generation, and decodes them into multi-modal geometric outputs, including globally consistent point clouds \(\hat{\mathcal{P}} \in \mathbb{R}^{N \times H \times W \times 3}\), per-view depth maps \(\hat{\mathcal{D}} \in \mathbb{R}^{N \times H \times W \times 1}\) and camera parameters \(\hat{\mathcal{T}} \in \mathbb{R}^{N \times 9}\).
Since the latent space of a video diffusion model exhibits a different spatial-temporal resolution from VGGT tokens and operates in a substantially lower-dimensional feature space~(e.g., \(c=16\)~\cite{wan}), we train an adapter \((\mathcal{E}_\mathrm{adp}, \mathcal{D}_\mathrm{adp})\) to bridge the two feature spaces by mapping the geometric tokens \(\mathcal{V}\) into the latent space of the video diffusion model and project them back:
\begin{align}
    \mathcal{E}_\mathrm{adp}&: \mathcal{V} \rightarrow \mathcal{G} \in \mathbb{R}^{n \times h \times w \times c}, \\
    \mathcal{D}_\mathrm{adp}&: \mathcal{G} \rightarrow \mathcal{V} \in \mathbb{R}^{N \times L \times h_v \times w_v \times C},
    \vspace{-0.1cm}
\end{align}
where \(n\times h \times w\) is the downsampled resolution.

The resulting geometric latents \(\mathcal{G}\) share the same spatial-temporal resolution and feature dimension as those of the video diffusion model, enabling joint generation of appearance and geometry within a unified latent space.

\begin{table*}[htbp]
    \centering
    \footnotesize
    \begin{adjustbox}{max width=\linewidth}
    \begin{tabular}{c|lcccccccccccc}
    \toprule
    \multirow{2}{*}{\rotatebox[origin=c]{90}{\scriptsize \textbf{Cond.}}} & \multirow{2}{*}{Method} &
    \multicolumn{6}{c}{RealEstate10K} &
    \multicolumn{6}{c}{DL3DV-10K} \\
    \cmidrule(lr){3-8} \cmidrule(lr){9-14}
    & & PSNR~$\uparrow$ & SSIM~$\uparrow$ & LPIPS~$\downarrow$ & I2V Subj.~$\uparrow$ & I2V BG~$\uparrow$ & I.Q.~$\uparrow$ & PSNR~$\uparrow$ & SSIM~$\uparrow$ & LPIPS~$\downarrow$ & I2V Subj.~$\uparrow$ & I2V BG~$\uparrow$ & I.Q.~$\uparrow$ \\
    \midrule
    \multirow{6}{*}{\rotatebox[origin=c]{90}{\textbf{1-view}}} &
    LVSM~\cite{lvsm} & \tbest{18.97} & \tbest{0.7161} & \tbest{0.2992} & \sbest{0.9946} & \sbest{0.9933} & 0.4923 & \tbest{15.61} & \tbest{0.5384} & \sbest{0.4434} & \sbest{0.9635} & \sbest{0.9675} & 0.4262 \\
    & Gen3C~\cite{gen3c} & \sbest{20.26} & \sbest{{0.7186}} & \sbest{0.2302} & 0.9931 & 0.9927 & 0.5200 & \sbest{16.21} & \sbest{0.5557} & \tbest{0.4575} & 0.9427 & 0.9547 & 0.4204 \\
    & GF~\cite{geometryforcing} & 16.32 & 0.5434 & 0.3803 & 0.9882 & 0.9789 & \tbest{0.5614} & 12.05 & 0.3458 & 0.5801 & 0.9335 & 0.9307 & \tbest{0.5410} \\
    & Aether~\cite{aether} & 16.57 & 0.6374 & 0.3808 & 0.9927 & 0.9910 & 0.5419 & 13.82 & 0.5167 & 0.5272 & \tbest{0.9589} & \tbest{0.9653} & 0.4571 \\ 
    & WVD~\cite{wvd} & 17.62 & 0.6658 & 0.3300 & \tbest{0.9935} & \tbest{0.9932} & \sbest{0.5847} & 14.25 & 0.4848 & 0.5063 & 0.9531 & 0.9613 & \sbest{0.5466} \\
    & Ours & \best{\textbf{20.51}} & \best{\textbf{0.7388}} & \best{\textbf{0.2281}} & \best{\textbf{0.9951}} & \best{\textbf{0.9952}} & \best{\textbf{0.5993}} & \best{\textbf{16.38}} & \best{\textbf{0.5821}} & \best{\textbf{0.4234}} & \best{\textbf{0.9657}} & \best{\textbf{0.9715}} & \best{\textbf{0.5497}} \\
    \midrule
    \multirow{7}{*}{\rotatebox[origin=c]{90}{\textbf{2-view}}} &
    DepthSplat~\cite{depthsplat} & \tbest{26.67} & \tbest{0.8711} & \tbest{0.1742} & 0.9909 & 0.9867 & 0.4379 & 16.83 & 0.6094 & \tbest{0.3971} & 0.9505 & 0.9532 & 0.3855 \\
    & LVSM~\cite{lvsm} & \best{\textbf{29.58}} & \best{\textbf{0.9197}} & \best{\textbf{0.1060}} & \best{\textbf{0.9954}} & \sbest{0.9943} & 0.5173 & \best{\textbf{18.80}} & \best{\textbf{0.6404}} & \sbest{0.3575} & \best{\textbf{0.9704}} & \best{\textbf{0.9736}} & 0.4616 \\
    & Gen3C~\cite{gen3c} & 23.83 & 0.8340 & 0.1947 & \tbest{0.9936} & \tbest{0.9930} & 0.5191 & \tbest{17.91} & \tbest{0.6120} & 0.4207 & 0.9470 & 0.9566 & 0.4239 \\
    & GF~\cite{geometryforcing} & 23.28 & 0.7426 & 0.2098 & 0.9893 & 0.9798 & \tbest{0.5614} & 14.39 & 0.4152 & 0.5160 & 0.9050 & 0.9110 & \tbest{0.4911} \\
    & Aether~\cite{aether} & 21.77 & 0.7645 & 0.2241 & 0.9919 & 0.9901 & 0.5258 & 15.68 & 0.5565 & 0.4555 & \tbest{0.9619} & \tbest{0.9676} & 0.4708 \\
    & WVD~\cite{wvd} & 23.78 & 0.7948 & 0.1949 & 0.9935 & 0.9926 & \sbest{0.5795} & 15.72 & 0.5522 & 0.4510 & 0.9520 & 0.9597 & \sbest{0.5584} \\
    & Ours & \sbest{27.05} & \sbest{0.8732} & \sbest{0.1352} & \sbest{0.9948} & \best{\textbf{0.9946}} & \best{\textbf{0.6025}} & \sbest{18.59} & \sbest{0.6149} & \best{\textbf{0.3416}} & \sbest{0.9685} & \sbest{0.9725} & \best{\textbf{0.5623}} \\
    \bottomrule
    \end{tabular}
    \end{adjustbox}
    \vspace{-0.25cm}
    \caption{\textbf{Quantitative Comparison of Appearance Generation.} We compare both 1-view and 2-view based settings.}
    \label{tab: 3d_generation_rgb}
    \vspace{-0.5cm}
\end{table*}

\boldparagraph{Training of the Adapter} %
Our adapter is trained with a reconstruction loss and a distribution alignment loss wrt. the appearance latents:
\vspace{-0.1cm}
\begin{equation}
    \mathcal{L} = \lambda_1 \mathcal{L}_\mathrm{rec} + \lambda_2 \mathcal{L}_\mathrm{KL}.
    \vspace{-0.1cm}
\end{equation}

Specifically, the reconstruction loss enforces the reconstructed geometry tokens \(\hat{\mathcal{V}}=\mathcal{D}_\mathrm{adp}(\mathcal{G})\) to match the original tokens \(\mathcal{V}\), and further regularize the consistency between the decoded outputs $(\hat{\mathcal{P}}, \hat{\mathcal{D}}, \hat{\mathcal{T}})$ and those derived from the original tokens $(\mathcal{P}, \mathcal{D}, \mathcal{T})$ by the pretrained DPT heads: 
\begin{align}
    \mathcal{L}_\mathrm{rec} = & \mathbb{E}\big[\|\hat{\mathcal{V}}-\mathcal{V}\|^2\big] 
            + \mathbb{E}\big[\|\hat{\mathcal{T}}-\mathcal{T}\|_1\big] \nonumber \\
           + & \mathbb{E}\big[\|\hat{\mathcal{D}}-\mathcal{D}\|^2\big]
           + \mathbb{E}\big[\|\hat{\mathcal{P}}-\mathcal{P}\|^2\big].
\end{align}
Furthermore, we observed in practice that although this supervision alone effectively compresses the geometry tokens, it does not constrain the mapped latent space, which hinders diffusion training from converging and degrades generation quality. While most LDMs use VAE or VQ-VAE to constrain the latents, we propose to directly regularize our latent space by aligning it with the pretrained appearance latent distribution. Specifically,
we impose a KL loss on the geometry adapter, encouraging its latent distribution \(q_\mathcal{G}\) to align with the pretrained RGB latent distribution \(q_{\mathcal{A}}\):
\vspace{-0.07cm}
\begin{equation}
    \vspace{-0.1cm}
    \mathcal{L}_\mathrm{KL} = D_{\mathrm{KL}}(q_{\mathcal{G}} \,\|\, q_{\mathcal{A}}).
\end{equation}
This constraint ensures compatibility between the two latent spaces and facilitates simultaneous modeling of both appearance and geometry distributions.

\subsection{Geometry-Aware Joint Latent Generation}
\label{sec: dit}

\boldparagraph{Design of the Joint Latent Space}
After training of the adapter, we establish a compact latent space where geometry and appearance latents can be jointly processed. We then fine-tune a video diffusion model~\cite{wan}, denoted as \(G_\theta\), to generate both modalities of latents within this unified space. 

Specifically, we aim to generate latent codes \(\mathcal{Z}\) consisting of two components: appearance latents \(\mathcal{A} = \mathcal{E}_\mathcal{W}(\mathcal{I}) \in \mathbb{R}^{n \times h \times w \times c}\) and geometry latents \(\mathcal{G} \in \mathbb{R}^{n \times h \times w \times c}\). To avoid introducing additional trainable parameters and to preserve the pretrained video diffusion model's generative capability, we concatenate the two latents along the width dimension~\cite{4dnex} to form a unified latent representation:
\begin{equation}
    \mathcal{Z} = [\mathcal{A}; \mathcal{G}] \in \mathbb{R}^{n \times h \times 2w \times c},
    \vspace{-0.1cm}
\end{equation}
where \([\cdot; \cdot]\) denotes concatenation in the width dimension.

\boldparagraph{Training of the Diffusion Model}
To enhance controllability, we incorporate multiple condition signals into the diffusion process, including a text prompt \(\mathbf{y}\), 
a condition image sequence \(\mathcal{I}_{cond}\) 
with a flexible number of available frames~(where missing images are set to zero),
corresponding binary masks \(\mathcal{M}\) and optional per-view camera conditions \(\mathcal{T}_{cond}\). 
The overall diffusion process is defined as:
\begin{equation}
    \vspace{-0.1cm}
    G_\theta: (\mathcal{Z}_t; t, \mathbf{y}, \mathcal{I}_{cond}, \mathcal{M}, \mathcal{T}_{cond}) \rightarrow \hat{\mathcal{Z}}_{t-1},
    \vspace{-0.05cm}
\end{equation}
where \(\mathcal{Z}_t\) is the noised latent at timestep \(t\), \(\hat{\mathcal{Z}}_{t-1}\) is the predicted latent at \(t-1\). 

Note that we do \textbf{not} provide geometric latents as condition signals, allowing the model to handle diverse tasks from input images only.
During training, we uniformly sample conditions from the following options:~(1) the first frame~(1-view based),~(2) the first and last frames~(2-view based),~(3) all frames, and adjust the binary masks correspondingly. We also randomly drop camera conditions to ensure they can be omitted during inference.

\boldparagraph{Inference} 
Practically, we evaluate on three conditioning settings:
~(1) 1-view-based generation,~(2) 2-view-based generation, and~(3) feed-forward reconstruction with a image sequence. 
Each setting can be performed with or without camera conditions. For fairness, we remove the camera conditions in the feed-forward reconstruction experiments. 

\subsection{Decoding Latents into Scene Attributes}
\label{sec: decoding_and_merging}

Based on the pipeline described above, we achieve feed-forward 3D scene generation by sampling unified latents from noise using \(G_\theta\), and decoding them into RGB frames and geometry attributes using \textit{separate} decoders. 

The appearance latents \(\mathcal{A} \in \mathbb{R}^{n \times h \times w \times c}\) are decoded by the pretrained RGB VAE \(\mathcal{D}_\mathcal{W}\) to synthesize photorealistic video frames \(\mathcal{I}\).
Similarly, the geometry latents \(\mathcal{G} \in \mathbb{R}^{n \times h \times w \times c}\) are mapped by the geometry adapter \(\mathcal{D}_\mathrm{adp}\) to recover geometry tokens \(\mathcal{V}\).
These tokens are then decoded by pretrained VGGT heads \(\mathcal{D}_\mathcal{V}\) to obtain scene attributes, including globally consistent point clouds \(\mathcal{P}\), per-view depth maps \(\mathcal{D}\) and camera parameters \(\mathcal{T}\). Following VGGT, we unproject the depth maps using the generated camera parameters as the final geometry results.

\begin{table*}[htbp]
    \centering
    \footnotesize
    \begin{adjustbox}{max width=\linewidth}
    \begin{tabular}{c|lccccccccc}
    \toprule
    \multirow{2}{*}{\raisebox{-0.2\totalheight}{\rotatebox[origin=c]{90}{\textbf{Cond.}}}} & \multirow{2}{*}{Method} &
    \multicolumn{3}{c}{Co3Dv2} &
    \multicolumn{3}{c}{WildRGB-D} &
    \multicolumn{3}{c}{TartanAir} \\
    \cmidrule(lr){3-5} \cmidrule(lr){6-8} \cmidrule(lr){9-11}
    & & Accuracy~$\downarrow$ & Completeness~$\downarrow$ & CD~$\downarrow$ & Accuracy~$\downarrow$ & Completeness~$\downarrow$ & CD~$\downarrow$ & Accuracy~$\downarrow$ & Completeness~$\downarrow$ & CD~$\downarrow$ \\
    \midrule
    \multirow{4}{*}{\raisebox{-0.1\totalheight}{\rotatebox[origin=c]{90}{\textbf{1-view}}}} &
    Aether~\cite{aether} & \tbest{1.2630} & \best{2.6366} & \tbest{1.9498} & 0.3181 & \tbest{0.2951} & \tbest{0.3066} & \tbest{3.1547} & \tbest{4.5366} & 3.8457 \\
    & WVD~\cite{wvd} & 1.8038 & \sbest{1.4237} & \sbest{1.6137} & \tbest{0.2708} & \sbest{0.2562} & \sbest{0.2635} & 4.3944 & \sbest{3.0660} & \tbest{3.7302} \\
    & VGGT~\cite{vggt} & \best{\textbf{0.3291}} & 4.3830 & 2.3561 & \best{\textbf{0.0346}} & 0.6723 & 0.3534 & \best{\textbf{0.7379}} & 5.3595 & \sbest{3.0487} \\
    & Ours & \sbest{0.8284} & \best{\textbf{1.3811}} & \best{\textbf{1.1047}} & \best{\sbest{0.1581}} & \best{\textbf{0.2402}} & \best{\textbf{0.1992}} & \sbest{3.0250} & \best{\textbf{2.5367}} & \best{\textbf{2.7809}} \\
    \midrule
    \multirow{4}{*}{\rotatebox[origin=c]{90}{\textbf{2-view}}} &
    Aether~\cite{aether} & \tbest{0.9664} & \tbest{2.1376} & \tbest{1.5520} & 0.3536 & \tbest{0.2540} & 0.3038 & \tbest{2.7745} & 3.4420 & \tbest{3.1082} \\
    & WVD~\cite{wvd} & 2.1153 & \sbest{1.3009} & 1.7081 & \tbest{0.2483} & \sbest{0.1813} & \tbest{0.2148} & 4.3794 & \sbest{2.5268} & 3.4531 \\
    & VGGT~\cite{vggt} & \best{\textbf{0.3951}} & 2.1566 & \sbest{1.2759} & \best{\textbf{0.0276}} & 0.2650 & \sbest{0.1463} & \best{\textbf{0.9201}} & \tbest{3.2554} & \sbest{2.0877} \\
    & Ours & \sbest{0.7237} & \best{\textbf{1.2298}} & \best{\textbf{0.9767}} & \sbest{0.1109} & \best{\textbf{0.1744}} & \best{\textbf{0.1426}} & \sbest{2.2825} & \best{\textbf{1.6643}} & \best{\textbf{1.9734}} \\
    \bottomrule
    \end{tabular}
    \end{adjustbox}
    \vspace{-0.25cm}
    \caption{\textbf{Quantitative Comparison of Geometry Generation.} We compare both 1-view and 2-view based settings.}
    \label{tab: 3d_generation_geo}
    \vspace{-0.1cm}
\end{table*}

\begin{figure*}[t]
    \centering

    \def\mywidth{2.3cm}
    \begin{tabular}{P{0.5cm}P{\mywidth}P{\mywidth}P{\mywidth}P{\mywidth}P{\mywidth}P{\mywidth}}

        \begin{minipage}[c]{0.1\textwidth}
            \includegraphics[height=0.49\linewidth]{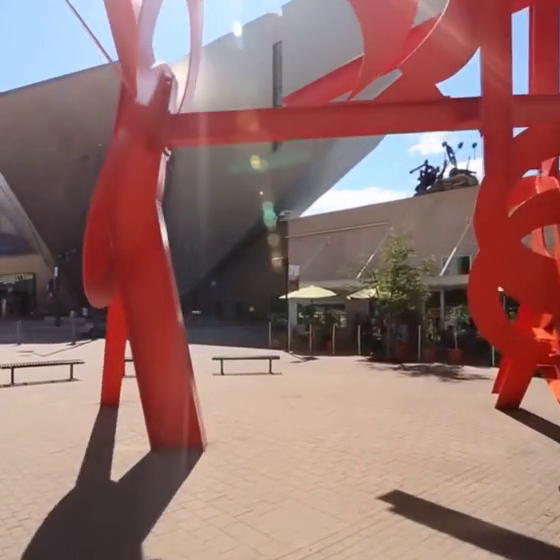} \\
            \includegraphics[height=0.49\linewidth]{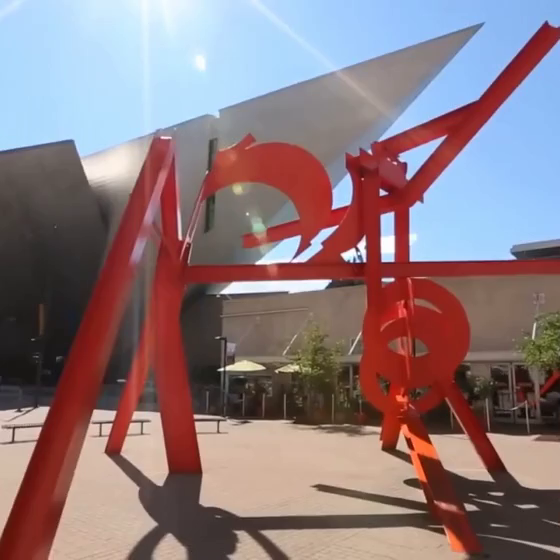}
        \end{minipage}
        &
        \includegraphics[width=1.17\linewidth]{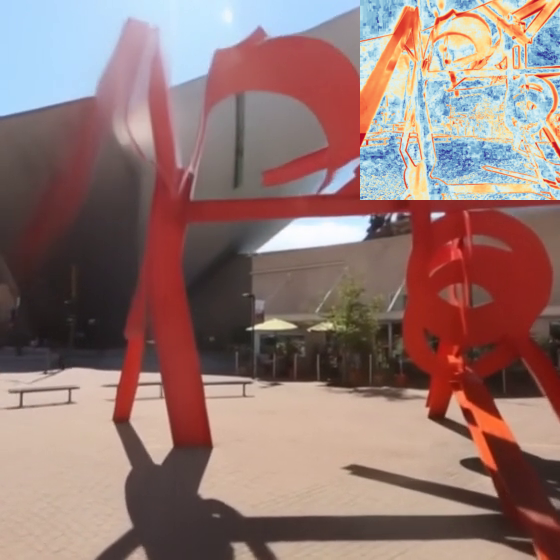} &
        \includegraphics[width=1.17\linewidth]{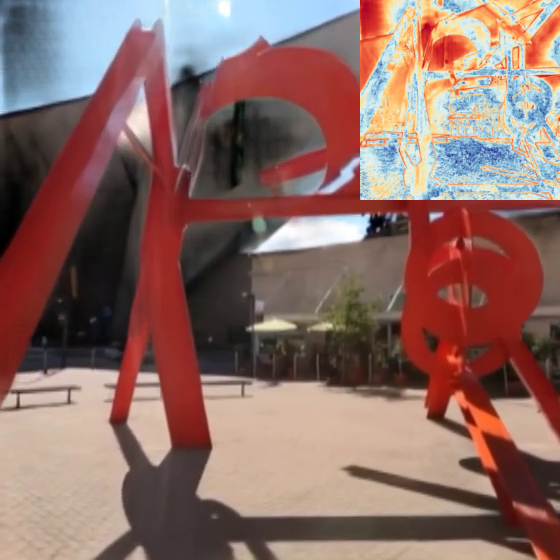} &
        \includegraphics[width=1.17\linewidth]{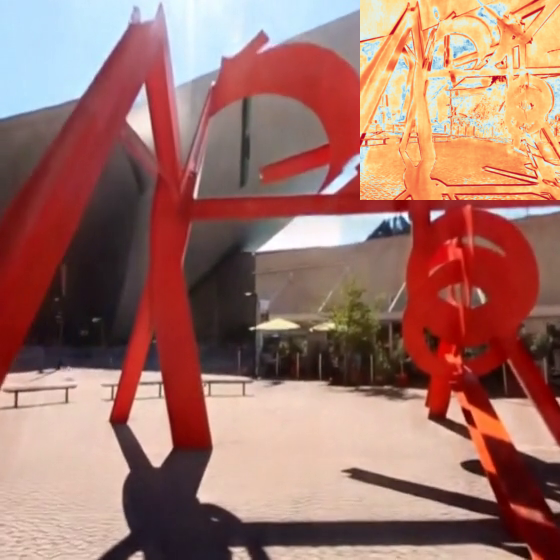} &
        \includegraphics[width=1.17\linewidth]{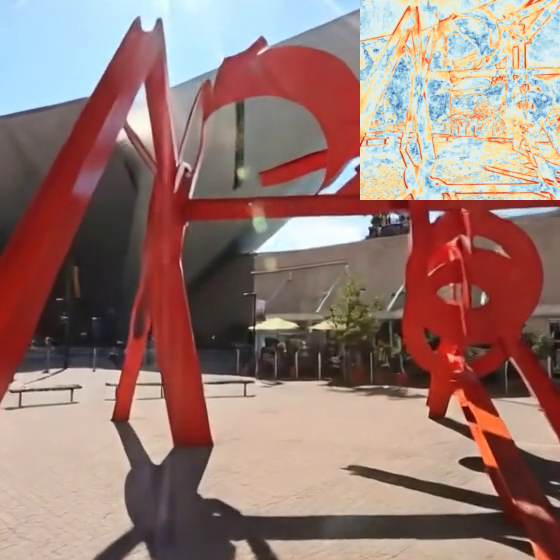} &
        \includegraphics[width=1.17\linewidth]{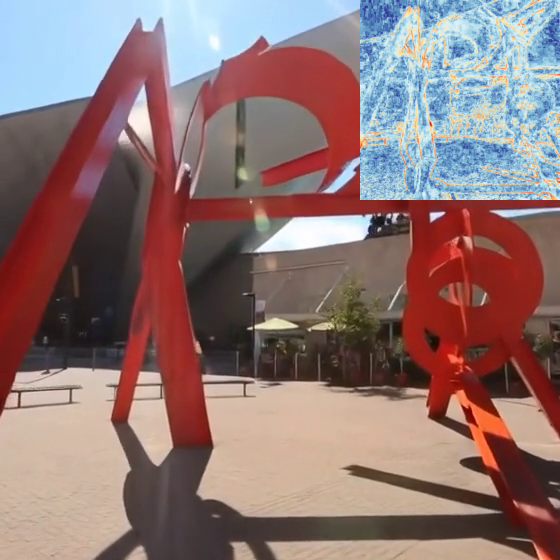} &
        \includegraphics[width=1.17\linewidth]{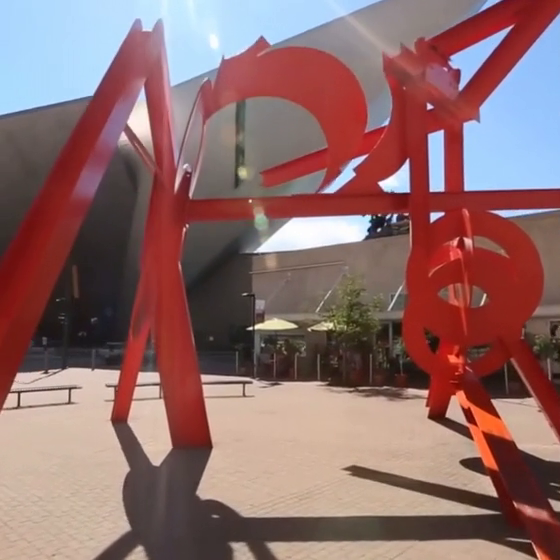}
        \\ [-2.5pt]

        \begin{minipage}[c]{0.1\textwidth}
            \includegraphics[height=0.49\linewidth]{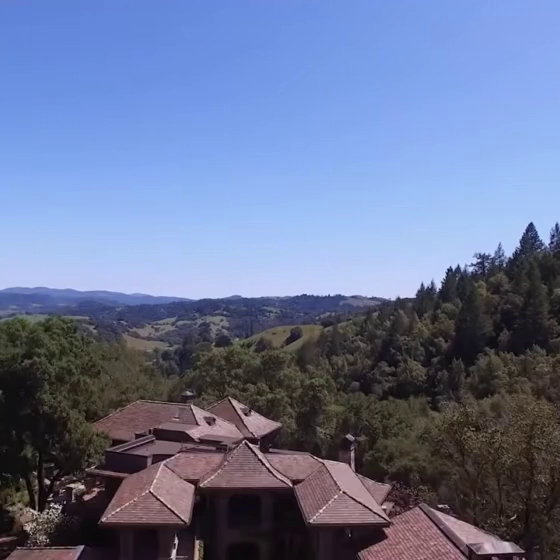} \\
            \includegraphics[height=0.49\linewidth]{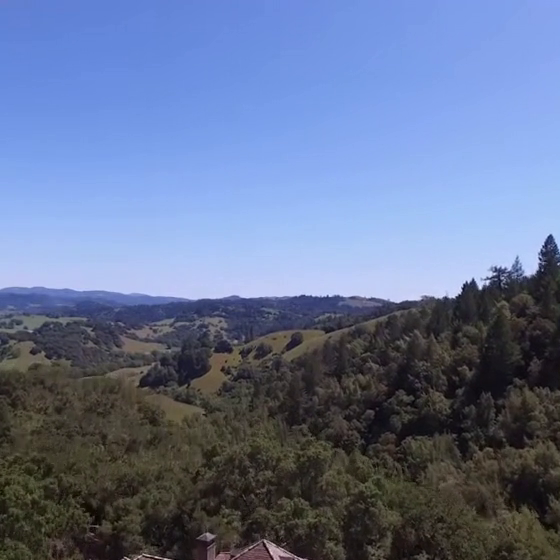}
        \end{minipage}
        &
        \includegraphics[width=1.17\linewidth]{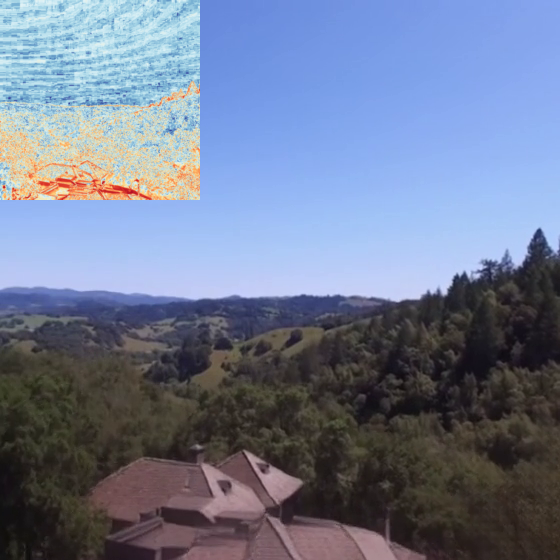} &
        \includegraphics[width=1.17\linewidth]{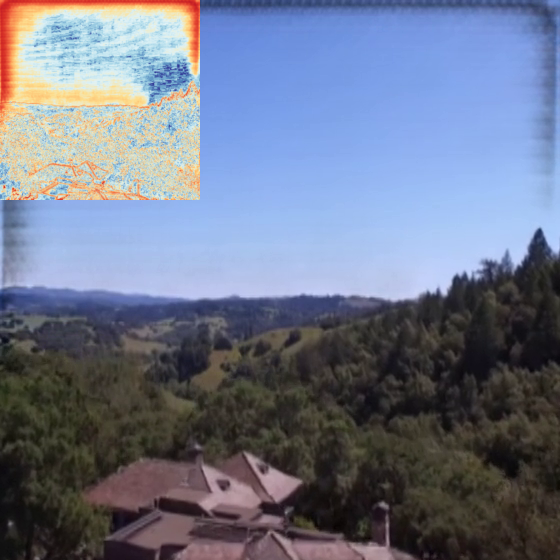} &
        \includegraphics[width=1.17\linewidth]{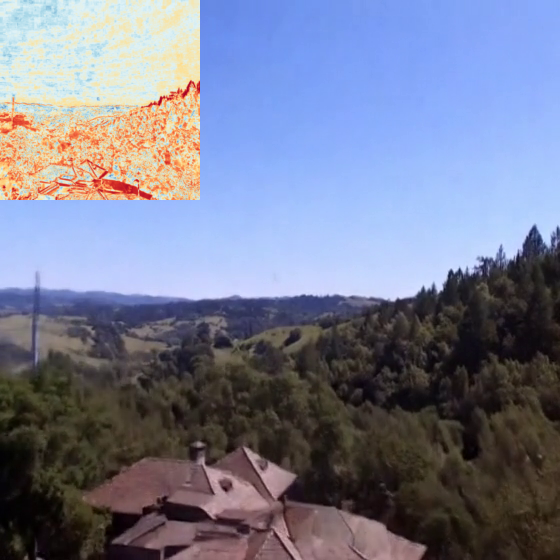} &
        \includegraphics[width=1.17\linewidth]{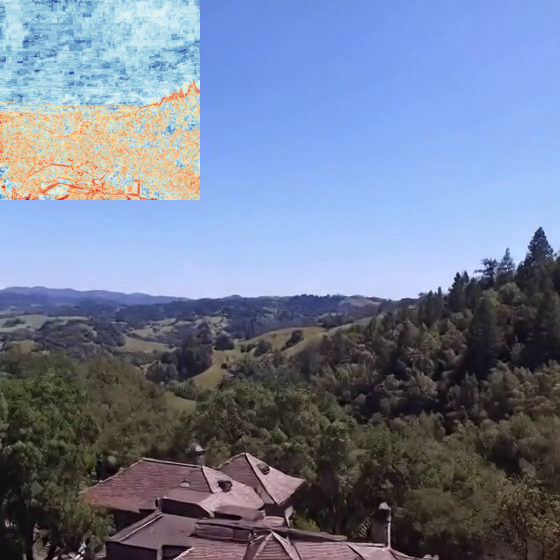} &
        \includegraphics[width=1.17\linewidth]{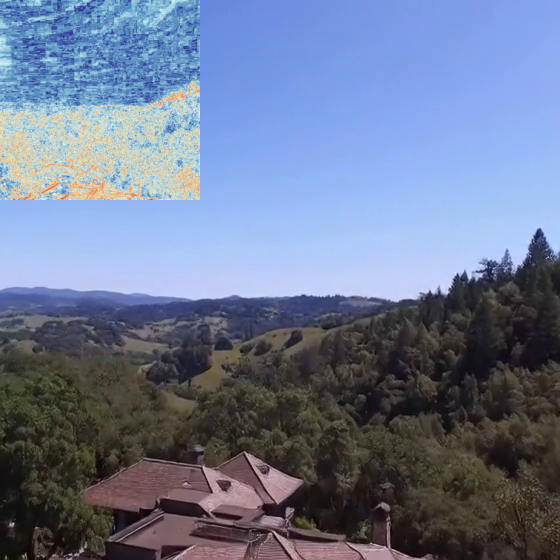} & 
        \includegraphics[width=1.17\linewidth]{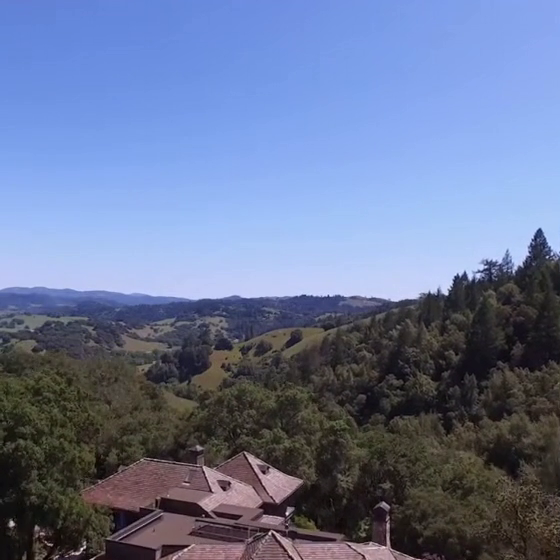}
        \\ [-2.5pt]

        \begin{minipage}[c]{0.1\textwidth}
            \includegraphics[height=0.49\linewidth]{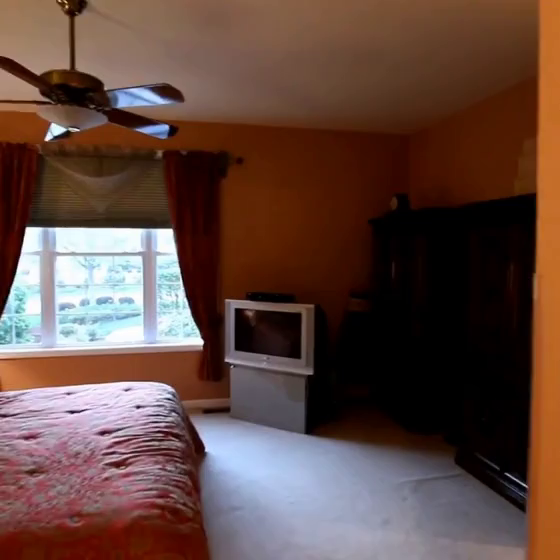} \\
            \includegraphics[height=0.49\linewidth]{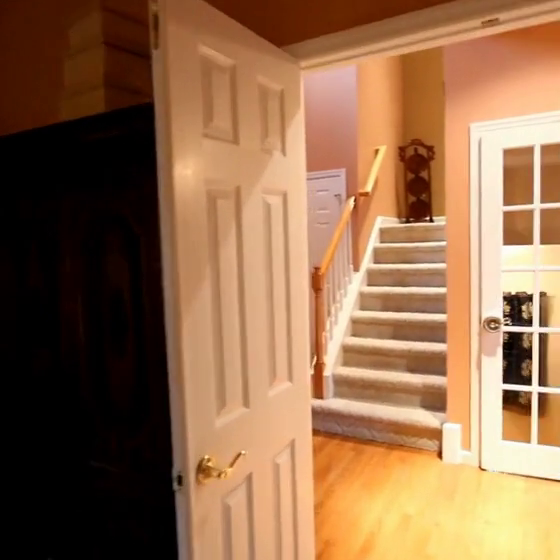}
        \end{minipage}
        &
        \includegraphics[width=1.17\linewidth]{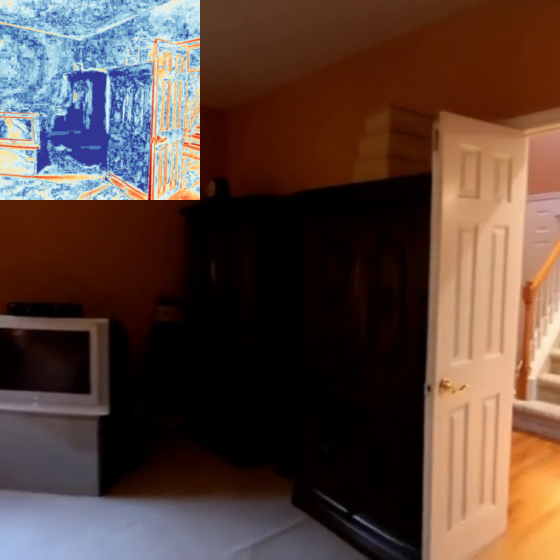} &
        \includegraphics[width=1.17\linewidth]{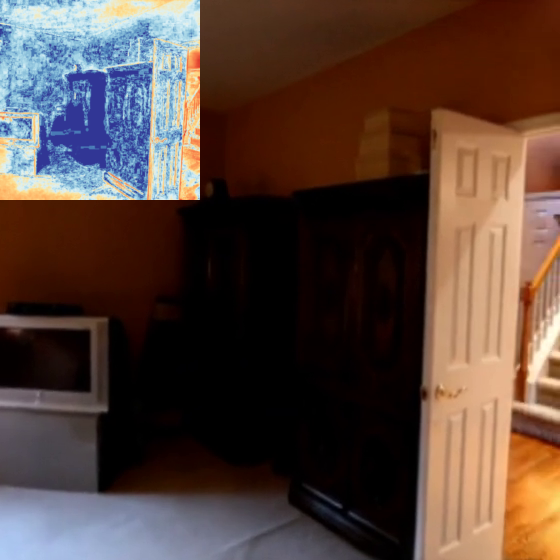} &
        \includegraphics[width=1.17\linewidth]{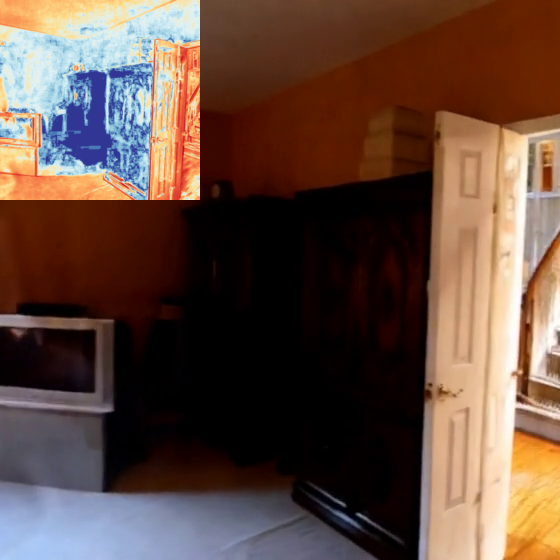} &
        \includegraphics[width=1.17\linewidth]{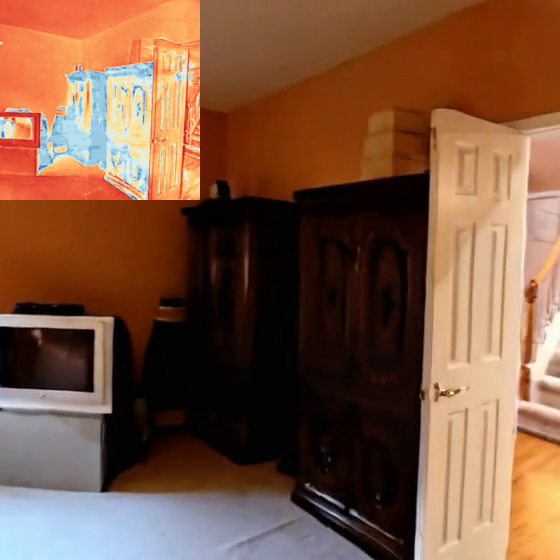} &
        \includegraphics[width=1.17\linewidth]{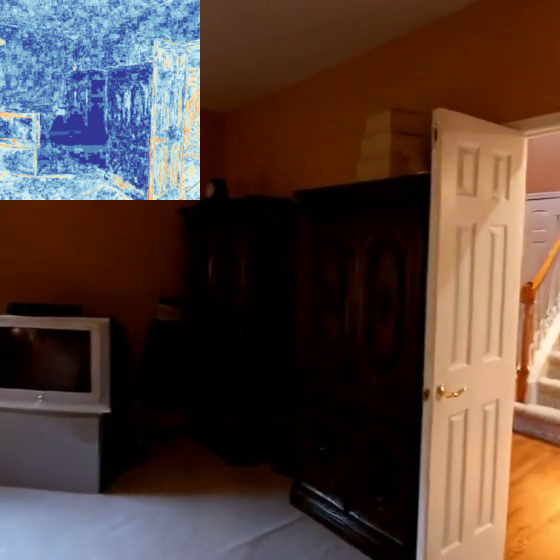} & 
        \includegraphics[width=1.17\linewidth]{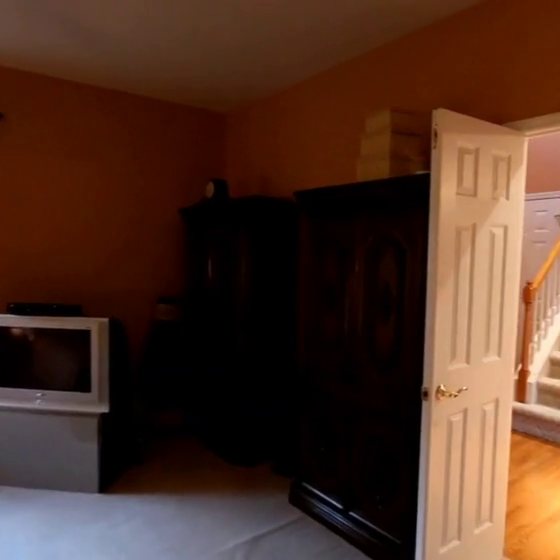}
        \\ [-2.5pt]

        \footnotesize Inputs &
        \footnotesize LVSM~\cite{lvsm} &
        \footnotesize DepthSplat~\cite{depthsplat} &
        \footnotesize Gen3C~\cite{gen3c} &
        \footnotesize WVD~\cite{wvd} & 
        \footnotesize Ours &
        \footnotesize Ground Truth
    \end{tabular}
    \vspace{-0.25cm}
    \caption{\textbf{Qualitative Comparison of Novel View Synthesis} with 2-view conditions. The input images are shown on the left, and error maps are displayed overlaid on the results. Bluer colors indicate smaller errors, while redder colors indicate larger errors.}
    \label{fig: first_last_frame_to_3d}
    \vspace{-0.4cm}
\end{figure*}

\section{Experiments}
\label{sec: experiments}

In this section, we compare our method with state-of-the-art approaches across various conditions. We first describe the training details in~\cref{sec: training_detail}, followed by both quantitative and qualitative evaluations on 3D generation~(\cref{sec: 3d_generation}) and reconstruction~(\cref{sec: 3d_reconstruction}). Finally, we present ablation studies to further validate the effectiveness of our approach in~\cref{sec: ablation}. We highlight the~\colorbox{lightred}{best},~\colorbox{lightorange}{second-best}, and~\colorbox{lightyellow}{third-best} scores achieved on any metrics.

\begin{table*}[htbp]
    \centering
    \footnotesize\relsize{0.5}
    \begin{adjustbox}{max width=\linewidth}
    \begin{tabular}{lccccccccc}
    \toprule
    \multirow{2}{*}{Method} & \multicolumn{3}{c}{Co3Dv2} & \multicolumn{3}{c}{WildRGB-D} & \multicolumn{3}{c}{TartanAir} \\
    \cmidrule(lr){2-4} \cmidrule(lr){5-7} \cmidrule(lr){8-10}
    & Accuracy~$\downarrow$ & Completeness~$\downarrow$ & CD~$\downarrow$ & Accuracy~$\downarrow$ & Completeness~$\downarrow$ & CD~$\downarrow$ & Accuracy~$\downarrow$ & Completeness~$\downarrow$ & CD~$\downarrow$ \\
    \midrule
    VGGT~\cite{vggt} & \best{\textbf{0.9157}} & \sbest{1.0107} & \sbest{0.9632} & \best{\textbf{0.0925}} & \sbest{0.1405} & \best{\textbf{0.1165}} & \sbest{2.2929} & \best{\textbf{0.8985}} & \sbest{1.5957} \\
    WVD~(VAE only) & 1.0533 & 1.3627 & 1.2080 & 0.1273 & 0.1780 & 0.1526 & 3.5337 & 2.0396 & 2.7867\\
    Ours~(VAE only) & \sbest{0.9236} & \tbest{1.0735} & \tbest{0.9986} & \sbest{0.0929} & \best{\textbf{0.1400}} & \best{\textbf{0.1165}} & \tbest{2.2972} & \sbest{1.0063} & \tbest{1.6518} \\
    Aether~\cite{aether} & 1.7755 & 1.2280 & 1.5018 & 0.3033 & 0.1665 & 0.2349 & 3.0287 & 2.3684 & 2.6985 \\
    WVD~\cite{wvd} & 1.7997 & 1.4609 & 1.6303 & 0.2758 & 0.1542 & 0.2150 & 3.8018 & 2.0820 & 2.9419 \\
    Ours & \tbest{0.9270} & \best{\textbf{0.9980}} & \best{\textbf{0.9625}} & \tbest{0.1058} & \tbest{0.1463} & \tbest{0.1260} & \best{\textbf{1.9243}} & \tbest{1.0959} & \best{\textbf{1.5101}} \\
    \bottomrule
    \end{tabular}
    \end{adjustbox}
    \vspace{-0.25cm}
    \caption{\textbf{Quantitative Comparison of Geometry Reconstruction.} WVD~(VAE only) uses pretrained RGB VAE to encode and reconstruct point clouds, while Ours~(VAE only) projects and reconstruct VGGT tokens to decode scene geometry.}
    \label{tab: feed_forward_geo}
    \vspace{-0.45cm}
\end{table*}

\subsection{Training Details}
\label{sec: training_detail}

\boldparagraph{Datasets} We train our model on a diverse collection of 3D datasets with camera calibrations, including: RealEstate10K~\cite{re10k}, DL3DV-10K~\cite{dl3dv}, ACID~\cite{acid}, TartanAir~\cite{tartanair}, KITTI-360~\cite{kitti360}, Waymo~\cite{waymo}, Co3Dv2~\cite{co3d}, MVImgNet~\cite{mvimgnet}, Virtual KITTI 2~\cite{vkitti2} and WildRGB-D~\cite{wildrgbd}. 
Together, these datasets provide over \(300k\) multi-view consistent 3D scenes, spanning a wide range of domains, including object-centric, indoor, outdoor, driving, and synthetic scenarios. For RealEstate10K, we follow the official train-test split, while for the other datasets, we randomly sample around \(90\%\) of the scenes for training and use the rest for testing. Text prompts for each scene are generated using a multi-modal large language model~\cite{internlm}. Notably, our method does not require explicit 3D GT representations for training, which are unavailable for many datasets.

\boldparagraph{Implementation Details} For the geometry adapter, we adopt a causal autoencoder architecture similar to~\cite{wan}, but with different input, output, and hidden dimensions. The adapter is trained on the mixed dataset described above. To ensure stability, the model is initially trained with \(25\) frames at a resolution of \(560 \times 560\) for \(15k\) iterations, using a batch size of \(2\) and gradient accumulation steps of \(4\) on \(24\) H20 GPUs, resulting a total batch size of \(192\). It is then fine-tuned with \(49\) frames for another \(6k\) iterations, using a batch size of \(1\) and gradient accumulation steps of \(8\) on the same hardware. The adapter weights are randomly initialized.

For the video diffusion model, we fine-tune a pretrained image-camera conditioned Wan2.1~\cite{wan}. Similar to the geometry adapter, during training we randomly sample \(49\) consecutive frames from each video clip, which are then resized and center-cropped to \(560 \times 560\). The model is trained for \(8k\) iterations with a batch size of \(4\) on \(24\) H20 GPUs. To enhance capability in handling diverse conditioning inputs, each training step has \(\frac{1}{3}\) probability of using~(i) 1-view condition,~(ii) 2-view~(first-last frame) conditions, or~(iii) the entire frame sequence as input. Additionally, the text prompt is dropped with a \(20\%\) probability for CFG~\cite{cfg}, and the camera condition is omitted with a \(50\%\) probability.

\subsection{3D Generation}
\label{sec: 3d_generation}

\boldparagraph{Datasets and Metrics}
We evaluate 3D generation on RealEstate10K~\cite{re10k}, DL3DV-10K~\cite{dl3dv}, Co3Dv2~\cite{co3d}, WildRGB-D~\cite{wildrgbd} and TartanAir~\cite{tartanair} datasets. For each task, we assess both appearance~(RGB) and geometry~(point clouds) metrics. Appearance metrics are computed across all these datasets, while geometry metrics are evaluated only on Co3Dv2, WildRGB-D, and TartanAir, as the other datasets do not provide ground truth geometry. Note that we report appearance metrics only on RealEstate10K and DL3DV-10K, and geometry metrics only on Co3Dv2, WildRGB-D, and TartanAir in the main text. Please refer to Supp. for complete results on all datasets.

For appearance evaluation, we randomly sample \(200\) sequences with camera conditions from each of the RealEstate10K and DL3DV-10K, and compute PSNR, SSIM~\cite{ssim}, and LPIPS~\cite{lpips} between the generated and ground-truth images. We additionally report the VBench Score~\cite{vbench, vbenchpp} to assess the models' generative capability, focusing on I2V Subject~(I2V Subj.), I2V Background~(I2V BG), and Imaging Quality~(I.Q.) given the presence of image-based conditioning.

For geometry evaluation, we randomly sample \(300\) sequences with camera conditions from each of the Co3Dv2 and WildRGB-D, along with an additional \(80\) sequences from TartanAir. We first use the Umeyama algorithm~\cite{umeyama} to align the generated point clouds to the ground truth, then sample \(20k\) points from both point clouds using Farthest Point Sampling~(FPS)~\cite{pointnetpp}, and finally compute Accuracy, Completeness, and Chamfer Distance~(CD)~\cite{cd}.

\begin{figure}[t]
    \centering
    \small
    \begin{tabular}{{@{}c@{\hspace{1pt}}c@{\hspace{1pt}}c@{\hspace{1pt}}c@{}}}
        \includegraphics[width=0.115\textwidth]{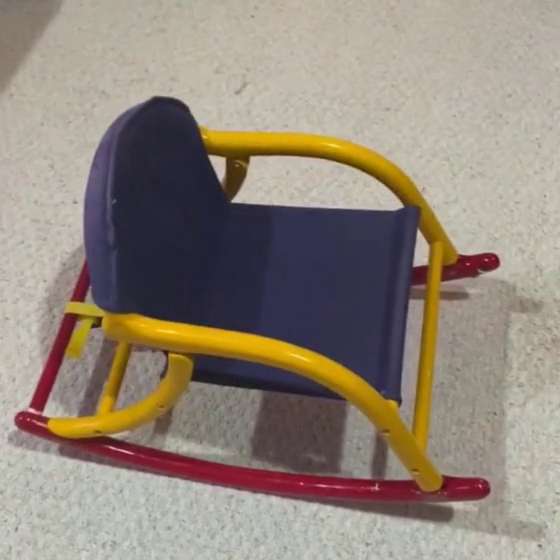} &
        \includegraphics[width=0.115\textwidth]{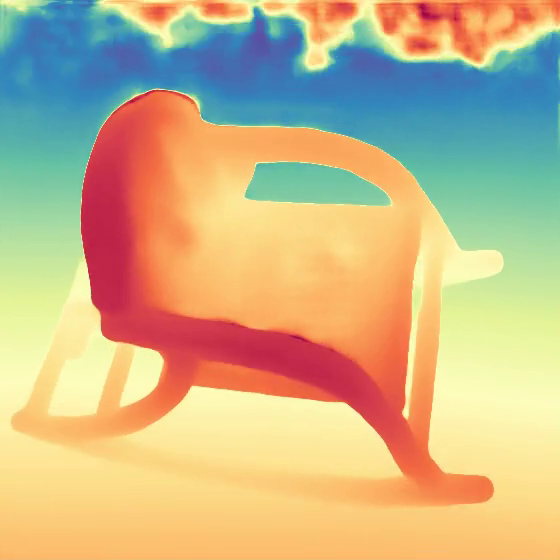} &
        \includegraphics[width=0.115\textwidth]{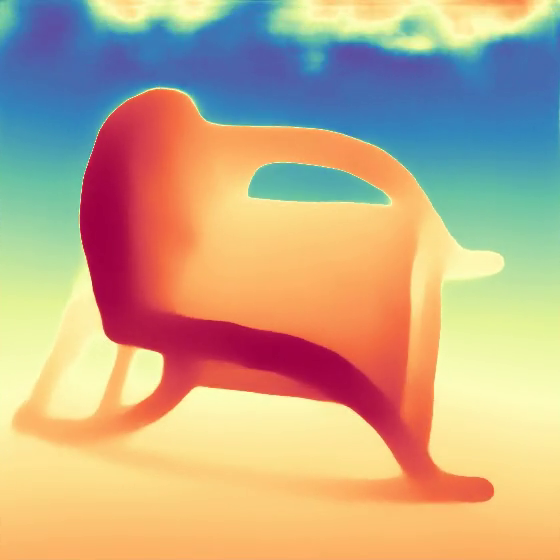} &
        \includegraphics[width=0.115\textwidth]{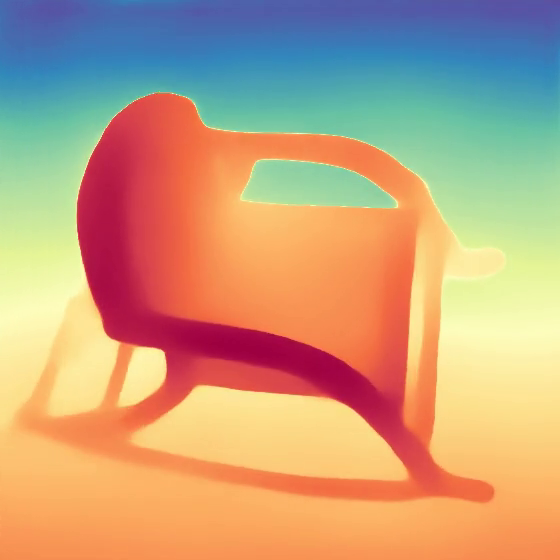} \\
        Input & VGGT & VAE~(Ours) & Ours \\
    \end{tabular}
    \vspace{-0.2cm}
    \caption{\textbf{Quali. Comparison of Geometry Reconstruction.}}
    \label{fig: reconstruction_vae}
    \vspace{-0.5cm}
\end{figure}

\boldparagraph{Comparison Baselines}
We compare our method with several state-of-the-art approaches that use image and camera conditions, including 
\textit{1) Reconstruction-based method:} DepthSplat~\cite{depthsplat}; 
\textit{2) 2D generation methods:} LVSM~\cite{lvsm}, Gen3C~\cite{gen3c}, and Geometry Forcing (GF)~\cite{geometryforcing}; and 
\textit{3) Explicit 3D generation methods: } Aether~\cite{aether} and WVD~\cite{wvd}. We use the official implementations for all of these methods except for WVD, as it is not open-sourced; we re-implement it following the same training strategy as ours. Note that Aether does not output point maps, so we back-project its generated depths using predicted camera parameters to obtain point clouds. 

\boldparagraph{Comparison on Appearance Generation}~\cref{tab: 3d_generation_rgb} presents quantitative results for 1-view-based and 2-view-based~(first-last frames) appearance generation.~\cref{fig: first_last_frame_to_3d} provides the corresponding qualitative comparisons for the 2-view setting (see Supp. for 1-view results). \method~outperforms or matches the baselines in most cases. 

\noindent\textit{1) Reconstruction-based methods:}
We evaluate DepthSplat only in the 2-view setting, as it requires multi-view inputs to construct cost volumes. Although it achieves competitive results, it leaves holes in occluded regions~(see~\cref{fig: first_last_frame_to_3d}). In contrast, our method can plausibly complete these regions using diffusion-based generation.

\noindent\textit{2) 2D generation methods:}
LVSM performs best in the 2-view case, as it is a non-generative model well suited for interpolation. However, it often produces blurred results in over-exposed scenes~(\cref{fig: first_last_frame_to_3d}, 
1st row), and its performance degrades notably in the 1-view case.
Gen3C also achieves competitive results in 1-view generation by combining depth-based warping and inpainting, but its quality heavily depends on depth accuracy, leading to misaligned boundaries when the depth estimates are inaccurate. In addition, it sometimes exhibits color differences from the input image, as shown in \cref{fig: first_last_frame_to_3d}.
More relevant to our method, GF similarly attempts to bridge reconstruction and generation. Unlike ours, which aligns latent spaces \textit{prior} to diffusion training, GF aligns intermediate diffusion features to the reconstruction model \textit{during} training, which is less effective in practice.
Finally, all of the above baselines operate purely in 2D and do not produce any 3D outputs.

\noindent\textit{3) Explicit 3D generation methods:}
Our method clearly surpasses the most relevant generative baselines, Aether and WVD, both of which jointly generate RGB images and scene geometry. As shown in~\cref{fig: first_last_frame_to_3d}, our approach yields higher-quality results and better camera alignment than WVD. This highlights the advantage of bridging reconstruction and generation models in the latent space, rather than compressing the reconstruction outputs for generation.

\begin{table}[t]
    \centering
    \begin{adjustbox}{max width=\linewidth}
    \begin{tabular}{lccccccccc}
    \toprule
    \multicolumn{1}{l}{\textbf{Cond.}} & \multicolumn{3}{c}{RealEstate10K} & \multicolumn{3}{c}{DL3DV-10K} & Co3Dv2 & WildRGB-D & TartanAir \\
    \cmidrule(lr){1-1} \cmidrule(lr){2-4} \cmidrule(lr){5-7} \cmidrule(lr){8-8} \cmidrule(lr){9-9} \cmidrule(lr){10-10}
    Method & PSNR~$\uparrow$ & SSIM~$\uparrow$ & LPIPS~$\downarrow$ & PSNR~$\uparrow$ & SSIM~$\uparrow$ & LPIPS~$\downarrow$ & CD~$\downarrow$ & CD~$\downarrow$ & CD~$\downarrow$ \\
    \midrule
    \multicolumn{10}{l}{\textbf{1-view}} \\
    \midrule
    2-Stage & \sbest{17.38} & \sbest{0.6617} & \sbest{0.3412} & \sbest{14.37} & \sbest{0.5085} & \sbest{0.5014} & \sbest{1.6223} & \sbest{0.2330} & \sbest{3.7029} \\
    w/o \(\mathcal{L}_\mathrm{KL}\) & \tbest{16.31} & \tbest{0.6476} & \tbest{0.3941} & \tbest{13.68} & \tbest{0.4797} & \tbest{0.5094} & \tbest{1.9620} & \tbest{0.3280} & \tbest{4.0395} \\
    Ours & \best{\textbf{20.51}} & \best{\textbf{0.7388}} & \best{\textbf{0.2281}} & \best{\textbf{16.38}} & \best{\textbf{0.5821}} & \best{\textbf{0.4234}} & \best{\textbf{1.1047}} & \best{\textbf{0.1992}} & \best{\textbf{2.7809}} \\
    \midrule
    \multicolumn{10}{l}{\textbf{2-view}} \\
    \midrule
    2-Stage & \sbest{23.56} & \sbest{0.7883} & \sbest{0.1931} & \sbest{15.92} & \sbest{0.5413} & \sbest{0.4427} & \sbest{1.3615} & \sbest{0.1623} & \sbest{2.6579} \\
    w/o \(\mathcal{L}_\mathrm{KL}\) & \tbest{21.62} & \tbest{0.7592} & \tbest{0.2185} & \tbest{15.41} & \tbest{0.5222} & \tbest{0.4527} & \tbest{1.7144} & \tbest{0.2898} & \tbest{3.5508} \\
    Ours & \best{\textbf{27.05}} & \best{\textbf{0.8732}} & \best{\textbf{0.1352}} & \best{\textbf{18.59}} & \best{\textbf{0.6149}} & \best{\textbf{0.3416}} & \best{\textbf{0.9767}} & \best{\textbf{0.1426}} & \best{\textbf{1.9734}}\\
    \bottomrule
    \end{tabular}
    \end{adjustbox}
    \vspace{-0.25cm}
    \caption{\textbf{Ablation Study} on appearance and geometry generation.}
    \label{tab: ablation_baselines}
    \vspace{-0.5cm}
\end{table}

\boldparagraph{Comparison on Geometry Generation} We further compare the generated point clouds with 3D-based methods in~\cref{tab: 3d_generation_geo}, and~\cref{fig: single_to_3d_rgb_transposed} visualizes point clouds generated from a single input view.
Our method clearly outperforms Aether and WVD in CD across both generation settings. The qualitative results in~\cref{fig: single_to_3d_rgb_transposed} are consistent with the quantitative findings: Aether and WVD exhibit poor global consistency, whereas our method produces more complete objects and scenes from single-view observations, with plausible geometry in unseen regions.
We also include VGGT in~\cref{tab: 3d_generation_geo} as a reference. It performs pure reconstruction from one or two input views without generation. Although VGGT achieves better accuracy, it suffers from lower completeness since it does not generate geometry for novel views, leading to a worse CD than \method.

\begin{figure}[t]
    \centering
    \small
    \begin{tabular}{{@{}c@{\hspace{1pt}}c@{\hspace{1pt}}c@{\hspace{1pt}}c@{\hspace{1pt}}c@{}}}
        \includegraphics[width=0.09\textwidth]{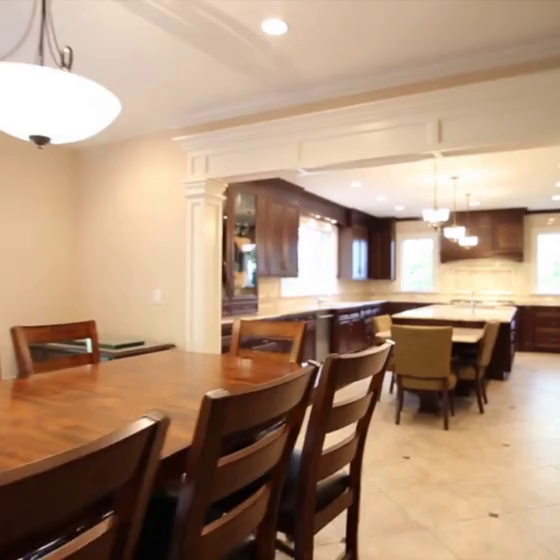} &
        \includegraphics[width=0.09\textwidth]{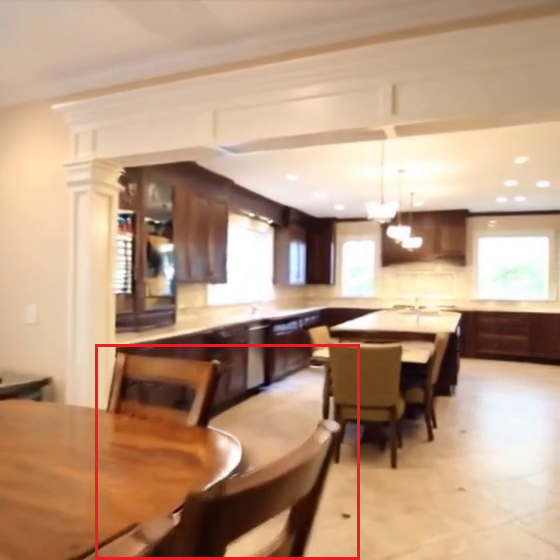} &
        \includegraphics[width=0.09\textwidth]{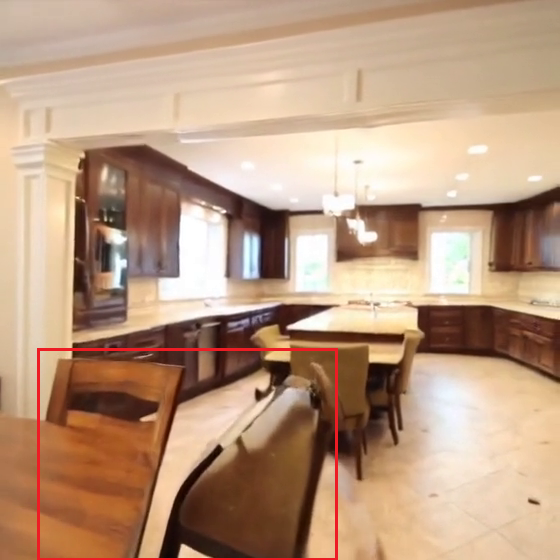} &
        \includegraphics[width=0.09\textwidth]{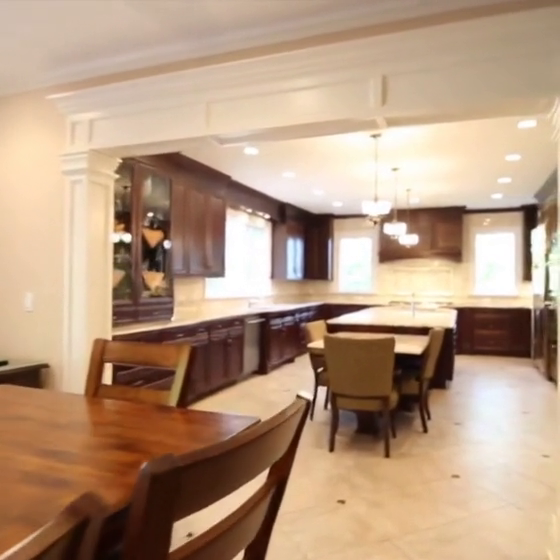} &
        \includegraphics[width=0.09\textwidth]{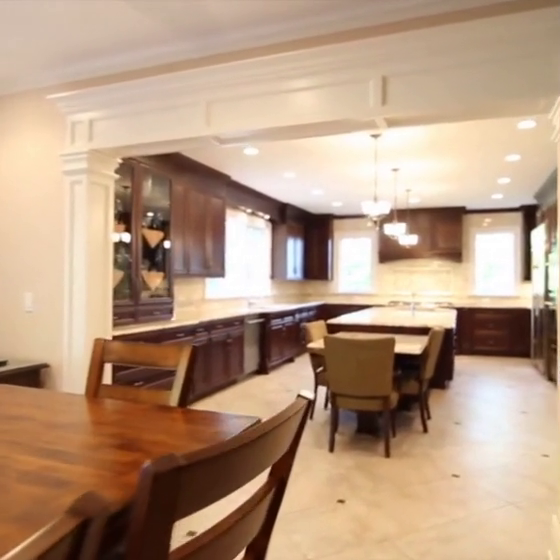} \\ [-2.5pt]
        \includegraphics[width=0.045\textwidth]{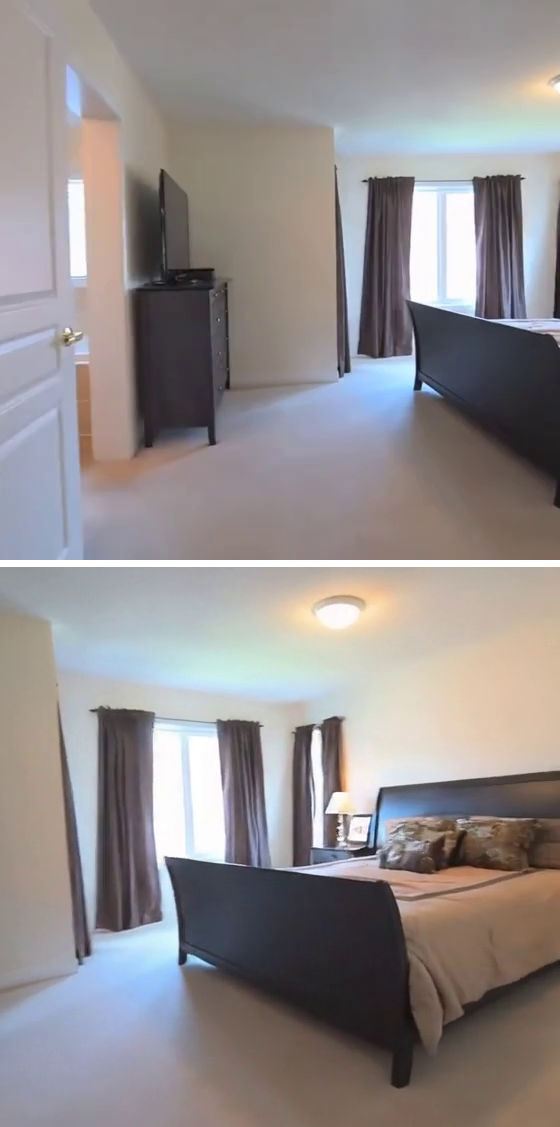} & 
        \includegraphics[width=0.09\textwidth]{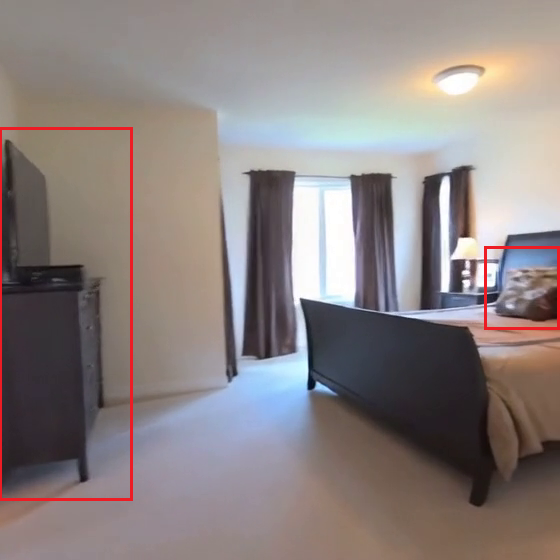} &
        \includegraphics[width=0.09\textwidth]{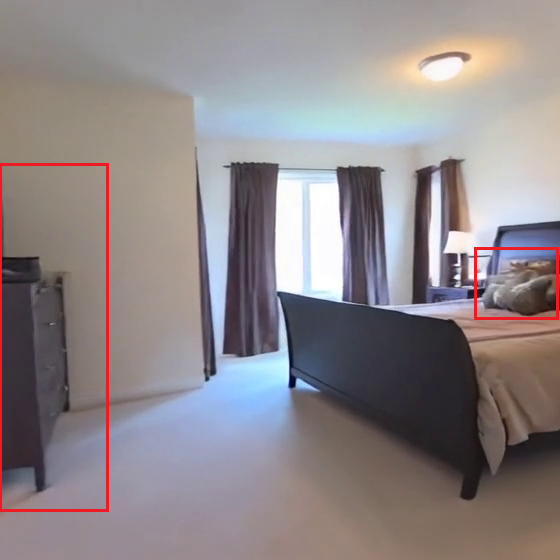} &
        \includegraphics[width=0.09\textwidth]{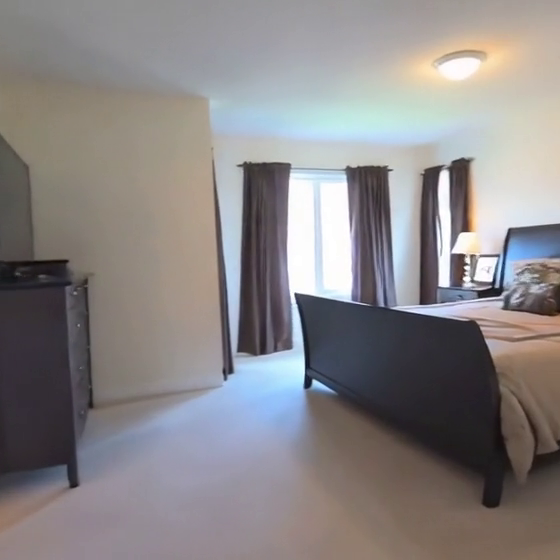} &
        \includegraphics[width=0.09\textwidth]{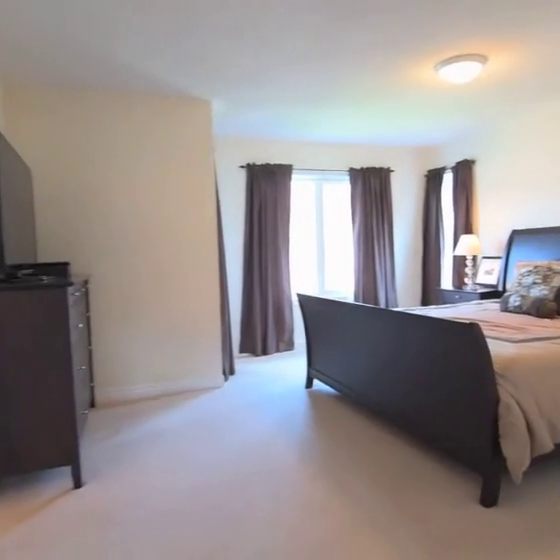} \\
        Inputs & 2-Stage & w/o \(\mathcal{L}_\mathrm{KL}\) & Ours & GT \\ [-2.5pt]
    \end{tabular}
    \vspace{-0.2cm}
    \caption{\textbf{Appearance Comparison} with ablation baselines. We highlight the artifacts of the baselines directly in the figures.}
    \label{fig: ablation_generation}
\end{figure}

\subsection{Feed-forward 3D Reconstruction}
\label{sec: 3d_reconstruction}

\boldparagraph{Dataset and Metrics}
Similar to 3D generation, we evaluate feed-forward 3D reconstruction on the same sequences sample from Co3Dv2, WildRGB-D and TartanAir datasets as before, but \textit{without} camera conditions. We assess both geometric quality and camera pose estimation. For geometry, we first align the predicted point clouds with GT, downsample both using FPS, and then compute Accuracy, Completeness, and Chamfer Distance~(CD). For camera pose estimation, we use the RealEstate10K and WildRGB-D datasets and follow VGGT~\cite{vggt} in reporting AUC@30, which combines both Relative Rotation Accuracy~(RRA) and Relative Translation Accuracy~(RTA). Please refer to Supplementary for camera pose estimation results.

\boldparagraph{Comparison Baselines}
We compare our method with \textit{1) feed-forward 3D reconstruction approach}, VGGT, as well as different variants of VAE for compressing VGGT. WVD~(VAE only) encodes and decodes VGGT's global point clouds using a pre-trained RGB VAE, while Ours~(VAE only)~encodes VGGT's geometry tokens through our adapter and decodes them back. We also compare against \textit{2) 3D generation methods} Aether and WVD. 

\begin{table}[t]
    \centering
    \footnotesize
    \begin{adjustbox}{max width=\linewidth}
    \begin{tabular}{l|c|cc|c|cc}
    \toprule
    \multirow{2}{*}{Method} & \multirow{2}{*}{\raisebox{-0.2\totalheight}{\rotatebox[origin=c]{90}{\textbf{Cond.}}}} &
    \multicolumn{1}{c}{RealEstate10K} &
    \multicolumn{1}{c}{WildRGB-D} &
    \multirow{2}{*}{\raisebox{-0.2\totalheight}{\rotatebox[origin=c]{90}{\textbf{Cond.}}}} &
    \multicolumn{1}{c}{RealEstate10K} &
    \multicolumn{1}{c}{WildRGB-D} \\
    \cmidrule(lr){3-3} \cmidrule(lr){4-4} \cmidrule(lr){6-6} \cmidrule(lr){7-7}
    & & AUC@30~$\uparrow$ & AUC@30~$\uparrow$ & & AUC@30~$\uparrow$ & AUC@30~$\uparrow$ \\
    \midrule
    Aether~\cite{aether} & \multirow{5}{*}{\rotatebox[origin=c]{90}{\textbf{1-view}}} & 0.6398 & 0.5375 & \multirow{5}{*}{\rotatebox[origin=c]{90}{\textbf{2-view}}} & 0.6220 & 0.5068 \\
    WVD~\cite{wvd} & & \tbest{0.6727} & \tbest{0.6780} & & \sbest{0.7249} & \tbest{0.7113} \\
    2-Stage & & \sbest{0.6832} & \sbest{0.6798} & & \tbest{0.7188} & \sbest{0.7211} \\
    w/o \(\mathcal{L}_\mathrm{KL}\) & & 0.4100 & 0.4759 & & 0.4683 & 0.4947 \\
    Ours & & \best{\textbf{0.7443}} & \best{\textbf{0.8004}} & & \best{\textbf{0.7732}} & \best{\textbf{0.8098}} \\
    \bottomrule
    \end{tabular}
    \end{adjustbox}
    \vspace{-0.25cm}
    \caption{\textbf{Quantitative Comparison of Camera Controllability} on RealEstate10K and WildRGB-D datasets.}
    \label{tab: generation_cam}
    \vspace{-0.5cm}
\end{table}

\boldparagraph{Comparison on Geometry Reconstruction}
We present quantitative results in~\cref{tab: feed_forward_geo}. \textit{1) Feed-forward 3D reconstruction:} Our VAE maintains the competitive performance of VGGT, whereas WVD's VAE produces subpar results when encoding and reconstructing explicit point clouds, as it is originally designed for RGB image reconstruction. Moreover, our generative version even enhances the reconstruction performance. This improvement arises because our method jointly models the appearance and geometry distribution, enabling mutual interactions between the two modalities and thereby refining noisy geometric predictions. As shown in~\cref{fig: reconstruction_vae}, VGGT occasionally exhibits floaters in its predicted geometry, and our adapted VAE inherits these artifacts. However, our generative model corrects the errors and produces cleaner depth. \textit{2) 3D generation methods}: Our method significantly outperforms existing generative models, Aether and WVD, on the reconstruction task, despite that our re-implemented WVD also leverages the prior of VGGT.

\begin{figure}[t]
    \centering
    \small
    \begin{tabular}{{@{}c@{\hspace{1pt}}c@{\hspace{1pt}}c@{\hspace{1pt}}c@{}}}
        \includegraphics[width=0.115\textwidth]{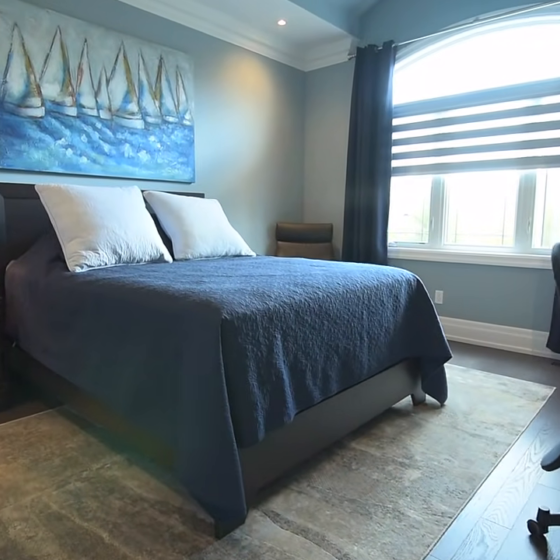} &
        \includegraphics[width=0.115\textwidth]{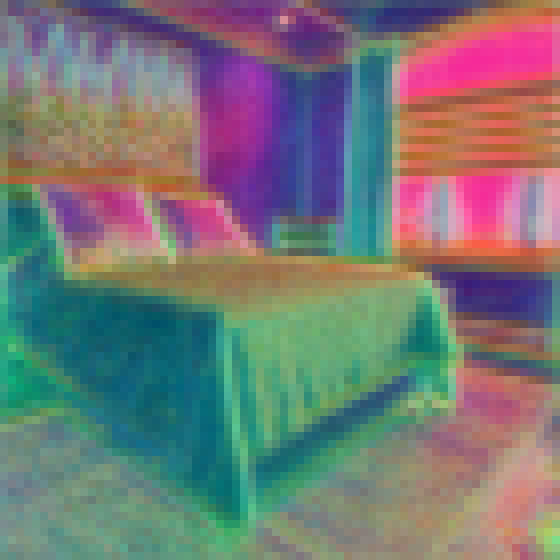} &
        \includegraphics[width=0.115\textwidth]{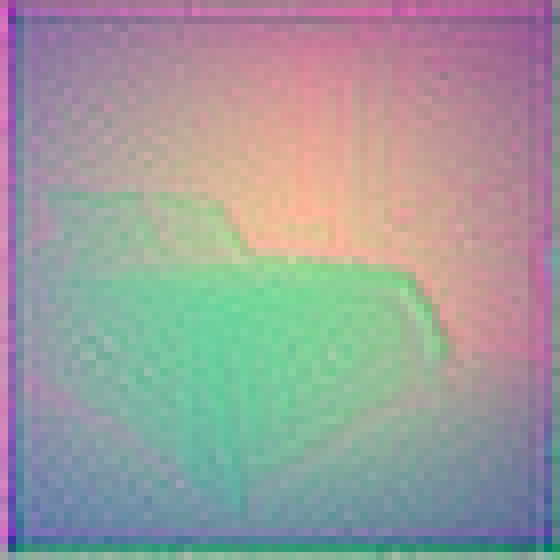} &
        \includegraphics[width=0.115\textwidth]{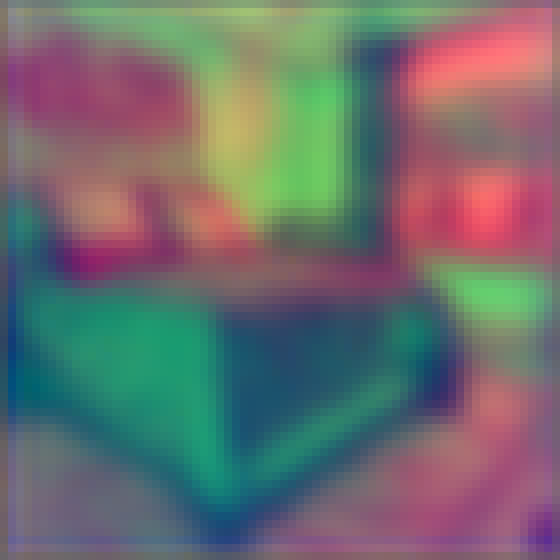} \\
        Input & RGB Latents & w/o \(\mathcal{L}_\mathrm{KL}\) & Ours \\
    \end{tabular}
    \vspace{-0.25cm}
    \caption{\textbf{Visualization of Latent Spaces} from different VAEs.}
    \label{fig: ablation_kl_loss}
    \vspace{-0.6cm}
\end{figure}

\subsection{Ablation Study}
\label{sec: ablation}

\boldparagraph{Effect of Joint Generation}
We investigate the impact of jointly generating RGB and geometry. To this end, we design a \textit{2-Stage} baseline: a video diffusion model finetuned on our training set generates only RGB under camera control, while geometry is predicted separately using VGGT from the generated images. Results in~\cref{tab: ablation_baselines} show that our joint generation approach outperforms the 2-Stage pipeline in both appearance and geometry. This is because the 2-Stage approach naively connects 2D generation with 3D reconstruction, leading to accumulated errors. Besides, ~\cref{tab: generation_cam} and~\cref{fig: ablation_generation} further show that our method outperforms this 2-Stage alternative in terms of camera control accuracy.

\boldparagraph{Effect of the Distribution Alignment Loss}
We further evaluate the impact of our distribution alignment loss~\(\mathcal{L}_\mathrm{KL}\) by training a variant of the adapter without it and visualizing the resulting latents in~\cref{fig: ablation_kl_loss}. Without this constraint, the geometry latents clearly deviate from the appearance latents. Results in~\cref{tab: ablation_baselines},~\cref{tab: generation_cam}, and~\cref{fig: ablation_generation} show that this misalignment hinders convergence and significantly degrades both camera controllability and generation quality.

\section{Conclusion}
\label{sec: conclusion}

We introduced \method, a unified framework that couples feed-forward reconstruction with video diffusion for high-fidelity 3D scene synthesis. By reformulating VGGT as an asymmetric geometry VAE with the geometry adapter and aligning its latents with a video diffusion model, \method{} jointly generates RGB videos and globally consistent 3D geometry.
Extensive experiments show that \method{} outperforms existing 2D and 3D based generative methods in both appearance and geometry, while also delivering superior camera controllability. Furthermore, \method{} improves the robustness of feed-forward reconstruction, highlighting the mutual benefits of combining generative priors with strong geometric foundations. We believe \method{} offers a promising direction toward controllable and high-fidelity 3D scene generation, and opens new possibilities for bridging reconstruction and generative modeling at scale.

\section*{Acknowledgements}
\label{sec: acknowledgement}

This work is supported by NSFC under grant 62441223.

{
    \small
    \bibliographystyle{ieeenat_fullname}
    \bibliography{main}
}
\clearpage
\setcounter{page}{1}
\maketitlesupplementary

\section{Implementation Details}
\label{sec: implementation_details}

\subsection{Processing Input Conditions}
\label{sec: processing_input_conditions}

We employ multiple conditions into the diffusion process, including a text prompt \(\mathbf{y}\), 
a condition image sequence \(\mathcal{I}_{cond}\) with a flexible number of available frames~(missing images are set to zero), corresponding binary masks \(\mathcal{M}\) and optional per-view camera conditions \(\mathcal{T}_{cond}\). 
The condition images \(\mathcal{I}_{cond}\) are encoded into appearance latents \(\mathcal{A}_{cond}\) by pretrained RGB VAE \(\mathcal{E}_\mathcal{W}\):
\begin{equation}
    \vspace{-0.1cm}
    \mathcal{E}_\mathcal{W}: \mathcal{I}_{cond} \rightarrow \mathcal{A}_{cond}  \in \mathbb{R}^{n \times h \times w \times c},
\end{equation}
while the masks \(\mathcal{M}\) are downsampled to \(\mathcal{M}_{a} \in \mathbb{R}^{n \times h \times w \times 4}\) to match the latent resolution. 
To ensure dimensional consistency with the noised latents, we initialize the geometry branch's condition latents \(\mathcal{G}_{cond} \in \mathbb{R}^{n \times h \times w \times c}\) and corresponding masks \(\mathcal{M}_{g} \in \mathbb{R}^{n \times h \times w \times 4}\) as \textbf{zeros}. 

Finally, the appearance and geometry latents are fused with their respective latent masks along the channel dimension, and the two modalities are further concatenated in the width dimension to construct the unified condition latent: 
\begin{equation}
    \vspace{-0.1cm}
    \mathcal{Z}_{cond} = [\mathcal{A}_{cond} \oplus \mathcal{M}_{a}; \mathcal{G}_{cond} \oplus \mathcal{M}_{g}] \in \mathbb{R}^{n \times h \times 2w \times c'},
\end{equation}
where \((\cdot \oplus \cdot)\) means concatenation along channel dimension, and \(c'=c+4\).
The input to the diffusion model is then constructed by concatenating the noised latents \(\mathcal{Z}_t\) with the condition latents \(\mathcal{Z}_{cond}\) along the channel dimension:
\begin{equation}
    \vspace{-0.1cm}
    \mathcal{Z}_{in} = \mathcal{Z}_t \oplus \mathcal{Z}_{cond},
    \vspace{-0.1cm}
\end{equation}
\begin{equation}
    \vspace{-0.1cm}
    G_\theta: \mathcal{Z}_{in} \rightarrow \hat{\mathcal{Z}}_{t-1}.
\end{equation}

\subsection{Model Architectures}
\label{sec: model_architectures}

\boldparagraph{Geometry Adapter}
We obtain our adapter \((\mathcal{E}_{\mathrm{adp}}, \mathcal{D}_{\mathrm{adp}})\) by modifying Wan's causal VAE~\cite{wan}. The adapter projects VGGT~\cite{vggt} geometry tokens \(\mathcal{V} \in \mathbb{R}^{N \times L \times h_v \times w_v \times C}\) into the video diffusion model's latent space and maps them back:
\begin{align}
    \vspace{-0.1cm}
    \mathcal{E}_\mathrm{adp}&: \mathcal{V} \rightarrow \mathcal{G} \in \mathbb{R}^{n \times h \times w \times c}, \\
    \mathcal{D}_\mathrm{adp}&: \mathcal{G} \rightarrow \mathcal{V} \in \mathbb{R}^{N \times L \times h_v \times w_v \times C},
    \vspace{-0.1cm}
\end{align}
where \(L = 5\), since we broadcast VGGT's camera tokens of each frame to the spatial resolution \(h_v \times w_v\)~(\(h_v = w_v = 40\)), and concatenate it with the other \(4\) tokens along the \(L\) dimension.

To match the VAE input format, we first reshape \(\mathcal{V}\) into \(\mathcal{V}' \in \mathbb{R}^{N \times h_v \times w_v \times (L \times C)}\). Accordingly, we set the adapter input dimension to \(L \times C = 10240\) and use hidden dimensions \([512, 256, 128, 128]\). We then re-sample the input tokens \(\mathcal{V}'\) to a spatial resolution of \(h \times w = 70 \times 70\) using nearest-exact interpolation, and apply a 2D convolution to project the channels to \(1024\). The resulting features are processed by causal convolution layers, where we keep the spatial resolution unchanged, yielding geometry latents \(\mathcal{G} \in \mathbb{R}^{n \times h \times w \times c}\). Similarly, the decoder \(\mathcal{D}_{\mathrm{adp}}\) mirrors the encoder architecture in reverse, reconstructing the geometry tokens \(\mathcal{V}\) from the latents \(\mathcal{G}\).

\begin{figure}[t]
    \centering
    \newcommand{\imgwidth}{0.132\textwidth}
    \def\mywidth{2.4cm}
    
    \setlength{\tabcolsep}{3pt}
    \setlength\dashlinedash{1.5pt}
    \setlength\dashlinegap{1pt}
    \setlength\arrayrulewidth{1.5pt}

    \begin{tabular}{P{0.4cm}P{\mywidth}P{\mywidth}P{\mywidth}P{\mywidth}P{\mywidth}P{\mywidth}P{\mywidth}}
        \rotatebox{90}{\footnotesize Inputs} &
        \begin{minipage}[c]{0.15\textwidth}
            \includegraphics[height=0.42\linewidth]{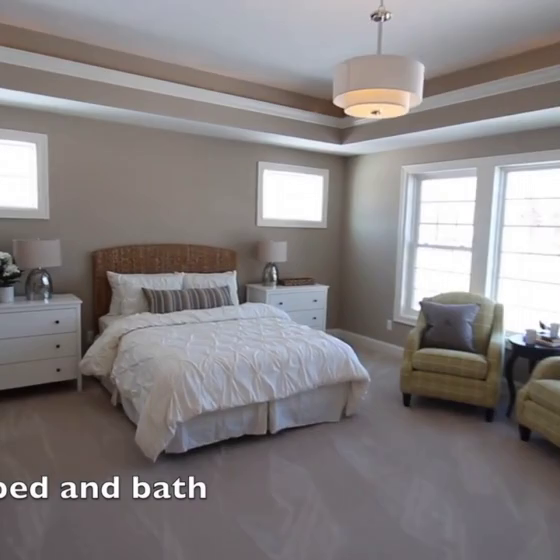}
            \includegraphics[height=0.42\linewidth]{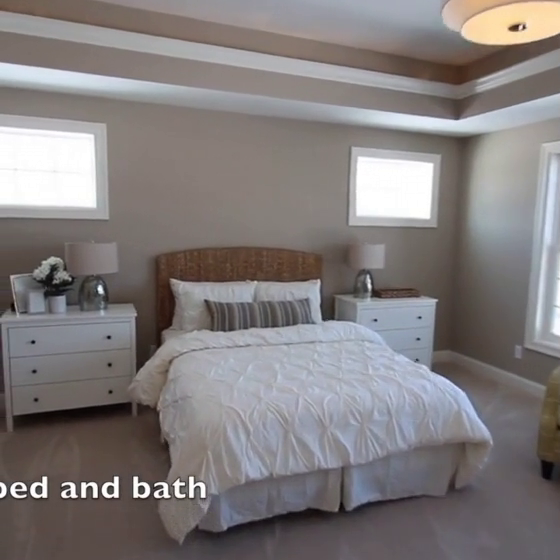}
        \end{minipage} &
        \begin{minipage}[c]{0.15\textwidth}
            \includegraphics[height=0.42\linewidth]{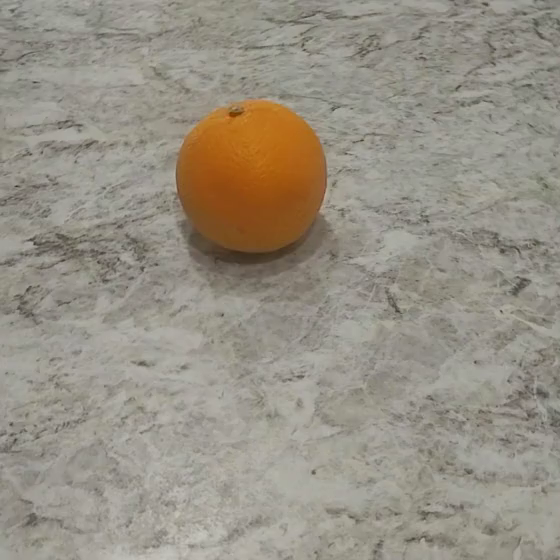}
            \includegraphics[height=0.42\linewidth]{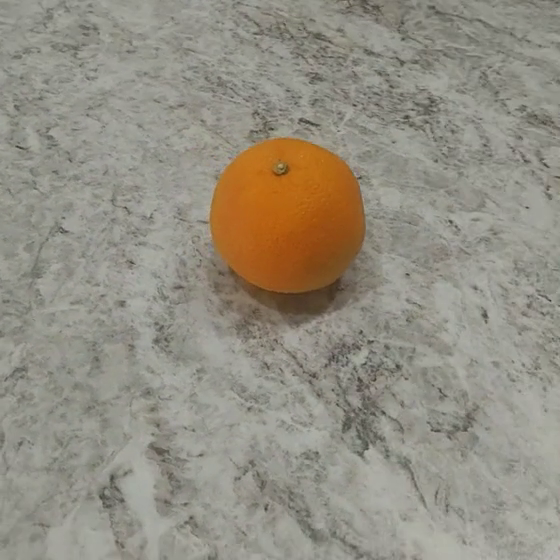}
        \end{minipage} &
        \begin{minipage}[c]{0.15\textwidth}
            \includegraphics[height=0.42\linewidth]{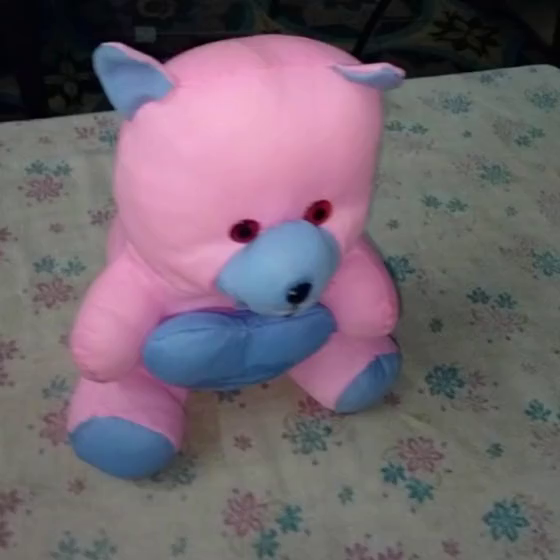}
            \includegraphics[height=0.42\linewidth]{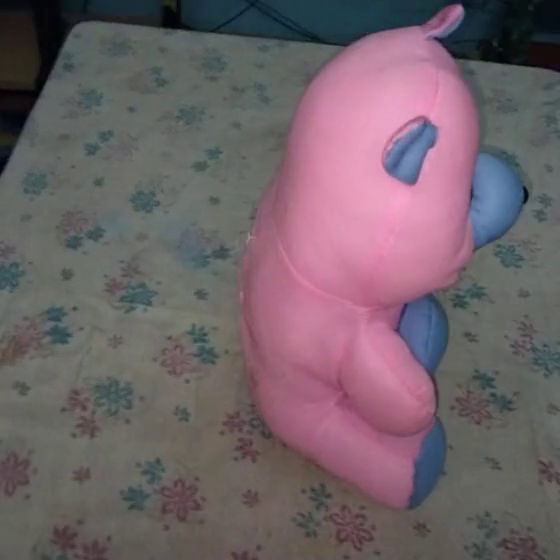}
        \end{minipage} \\
        \rotatebox{90}{\footnotesize Aether~\cite{aether}} &
        \raisebox{0.2\height}{\includegraphics[width=\imgwidth]{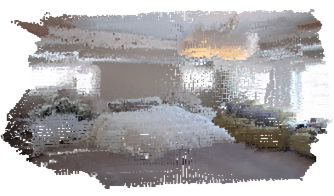}} &
        \raisebox{0.2\height}{\includegraphics[width=\imgwidth]{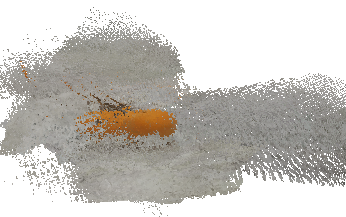}} &
        \raisebox{0.15\height}{\includegraphics[width=\imgwidth]{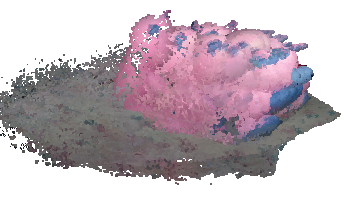}} \\ [-13pt]
        \rotatebox{90}{\footnotesize WVD~\cite{wvd}} &
        \raisebox{0.1\height}{\includegraphics[width=\imgwidth]{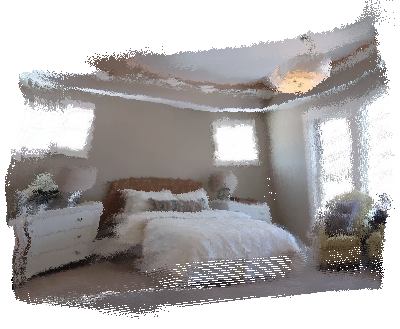}} &
        \raisebox{0.1\height}{\includegraphics[width=\imgwidth]{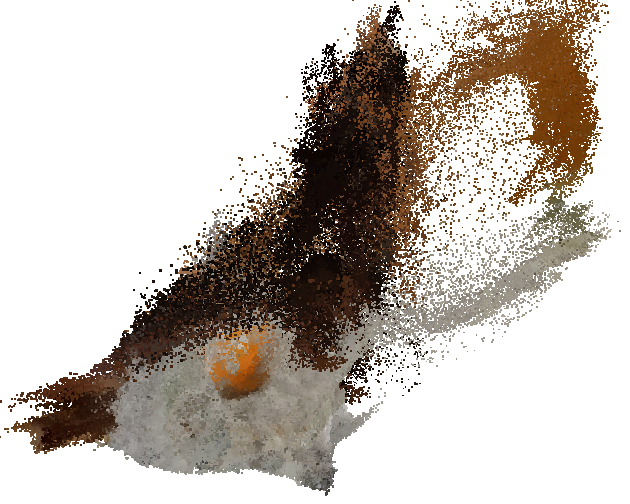}} &
        \raisebox{0.\height}{\includegraphics[width=\imgwidth]{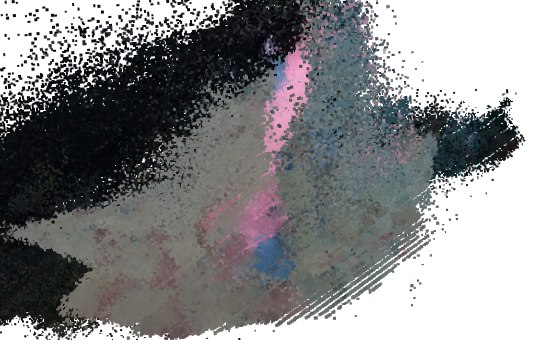}} \\ [-8pt]
        \rotatebox{90}{\footnotesize Ours} &
        \raisebox{0.1\height}{\includegraphics[width=\imgwidth]{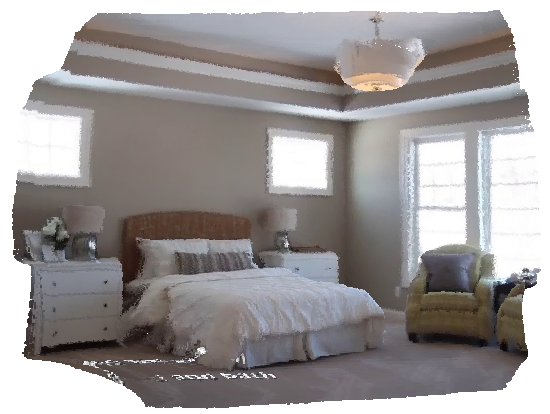}} &
        \raisebox{0.1\height}{\includegraphics[width=\imgwidth]{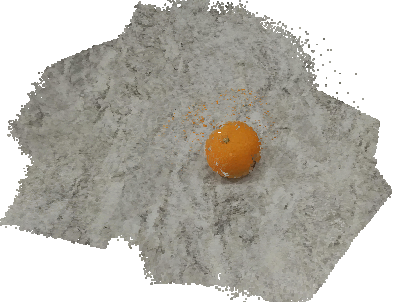}} &
        \raisebox{0.05\height}{\includegraphics[width=\imgwidth]{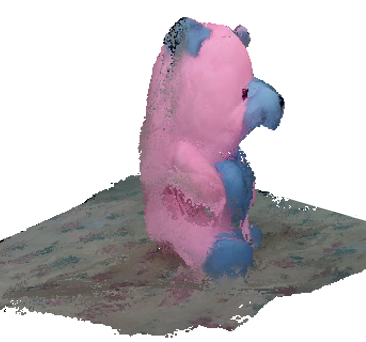}} \\
    \end{tabular}
    \vspace{-0.4cm}
    \caption{\textbf{Qualitative Comparison of Geometry Generation} in the 2-view based setting.}
    \label{fig: supp_first_last_frame_to_3d}
\end{figure}

\begin{figure}[t]
    \centering
    \small
    \begin{tabular}{{@{}c@{\hspace{3pt}}c@{\hspace{1pt}}c@{}}}
        \includegraphics[width=0.12\textwidth]{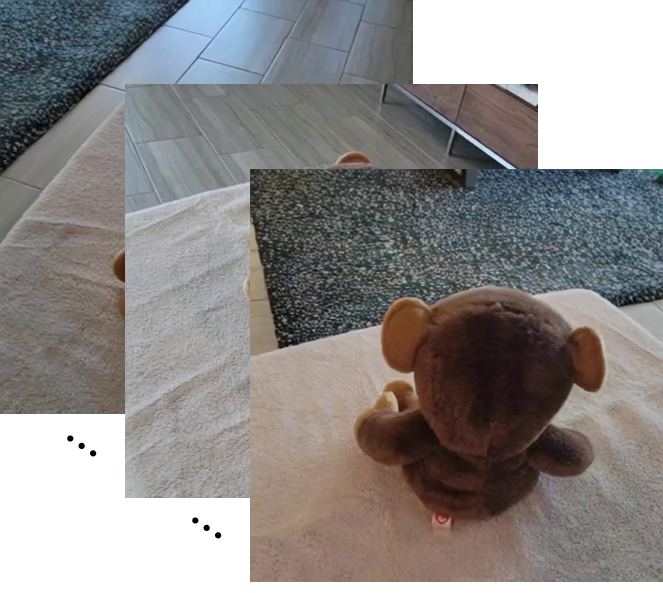} &
        \includegraphics[width=0.15\textwidth]{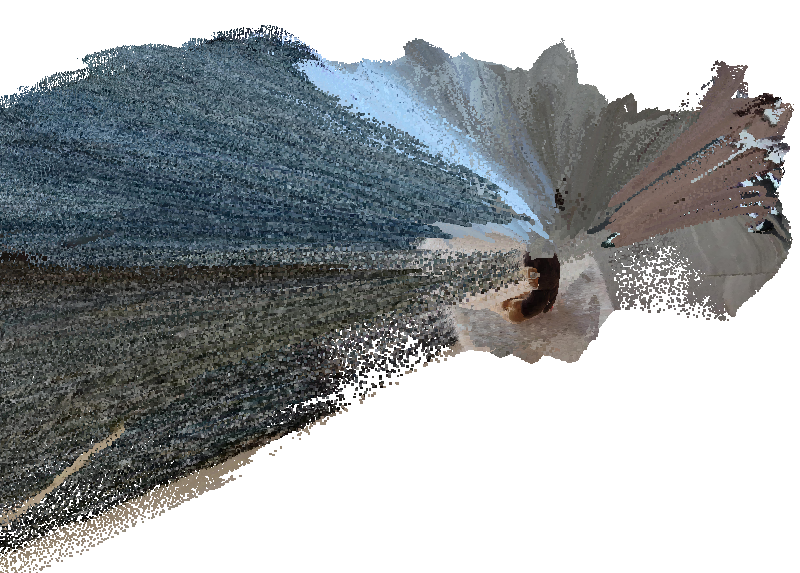} &
        \raisebox{0.05\height}{\includegraphics[width=0.16\textwidth]{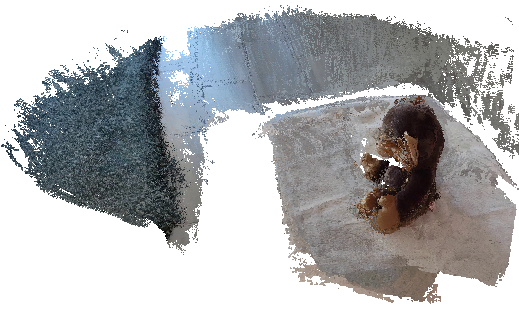}} \\ [-2pt]
        \includegraphics[width=0.12\textwidth]{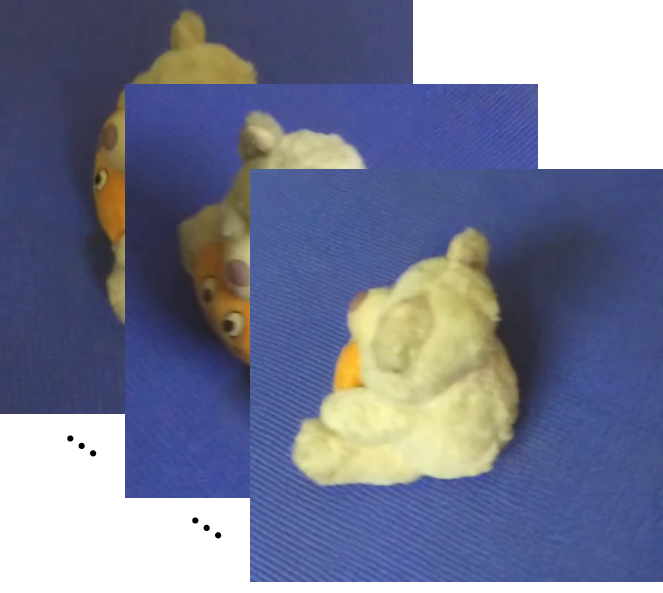} &
        \raisebox{-0.1\height}{\includegraphics[width=0.15\textwidth]{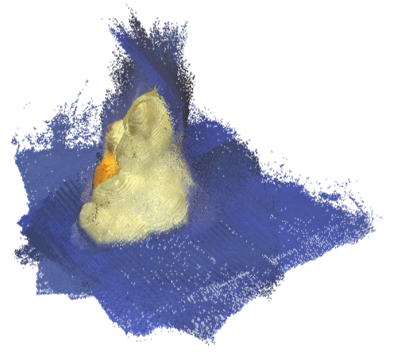}} &
        \raisebox{-0.12\height}{\includegraphics[width=0.15\textwidth]{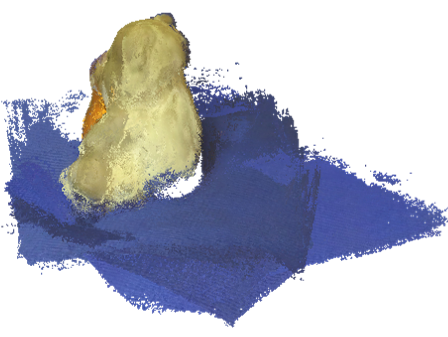}} \\ [4pt]
        Input & VGGT & Ours \\
    \end{tabular}
    \vspace{-0.2cm}
    \caption{\textbf{Qualitative Comparison of Geometry Reconstruction.}}
    \label{fig: supp_reconstruction}
    \vspace{-0.45cm}
\end{figure}

\boldparagraph{Diffusion Transformer}
We adapt the DiT architecture from VideoX-Fun's Wan2.1~\cite{wan} to accommodate our joint appearance-geometry latents. Specifically, we set the input channel dimension to \(c + c' = 36\). To support width-wise concatenation of appearance and geometry latents, we modify the positional embeddings so that corresponding pixels in the left and right halves of the latents share identical RoPE embeddings.

\begin{table*}[t]
    \centering
    \begin{adjustbox}{max width=\linewidth}
    \begin{tabular}{lcccccccccccccccccc}
    \toprule
    \multicolumn{1}{l}{\textbf{Cond.}} & \multicolumn{6}{c}{Co3Dv2} & \multicolumn{6}{c}{WildRGB-D} & \multicolumn{6}{c}{TartanAir} \\
    \cmidrule(lr){1-1} \cmidrule(lr){2-7} \cmidrule(lr){8-13} \cmidrule(lr){14-19}
    Method & PSNR~$\uparrow$ & SSIM~$\uparrow$ & LPIPS~$\downarrow$ & I2V Subj.~$\uparrow$ & I2V BG~$\uparrow$ & I.Q.~$\uparrow$ &  PSNR~$\uparrow$ & SSIM~$\uparrow$ & LPIPS~$\downarrow$ & I2V Subj.~$\uparrow$ & I2V BG~$\uparrow$ & I.Q.~$\uparrow$ & PSNR~$\uparrow$ & SSIM~$\uparrow$ & LPIPS~$\downarrow$ & I2V Subj.~$\uparrow$ & I2V BG~$\uparrow$ & I.Q.~$\uparrow$ \\
    \midrule
    \multicolumn{19}{l}{\textbf{1-view}} \\
    \midrule
    LVSM~\cite{lvsm} & \tbest{14.08} & \tbest{0.5623} & \tbest{0.5698} & \sbest{0.9482} & \sbest{0.9581} & 0.3579 & \tbest{13.9483} & \tbest{0.5239} & \tbest{0.5195} & \sbest{0.9692} & \sbest{0.9713} & 0.4004 & \sbest{14.44} & \sbest{0.5044} & \sbest{0.5210} & \sbest{0.9325} & \sbest{0.9540} & 0.3542 \\
    Gen3C~\cite{gen3c} & \sbest{15.82} & \sbest{0.5666} & \sbest{0.5095} & 0.9134 & 0.9355 & 0.4335 & \sbest{14.60} & \sbest{0.5463} & \sbest{0.4513} & 0.9622 & 0.9646 & 0.4629 & \tbest{13.95} & \tbest{0.4731} & \tbest{0.5385} & 0.9142 & 0.9403 & 0.3713 \\
    GF~\cite{geometryforcing} & 10.25 & 0.3150 & 0.6761 & 0.7933 & 0.8193 & \tbest{0.5320} & 11.8944 & 0.4147 & 0.5940 & 0.9215 & 0.9214 & \tbest{0.5310} & 10.21 & 0.3249 & 0.6249 & 0.7447 & 0.7864 & \tbest{0.4379} \\
    Aether~\cite{aether} & 12.78 & 0.5106 & 0.6052 & 0.9229 & 0.9395 & 0.4411 & 11.87 & 0.4289 & 0.5973 & 0.9595 & 0.9614 & 0.4786 & 12.88 & 0.4585 & 0.5645 & \tbest{0.9295} & \tbest{0.9480} & 0.4303 \\
    WVD~\cite{wvd} & 13.35 & 0.4733 & 0.5765 & \tbest{0.9339} & \tbest{0.9484} & \sbest{0.5355} & 12.95 & 0.4522 & 0.5362 & \tbest{0.9669} & \tbest{0.9671} & \sbest{0.5513} & 12.77 & 0.4513 & 0.5652 & 0.9271 & 0.9473 & \sbest{0.4571} \\
    Ours & \best{\textbf{16.09}} & \best{\textbf{0.5754}} & \best{\textbf{0.4997}} & \best{\textbf{0.9535}} & \best{\textbf{0.9588}} & \best{\textbf{0.5383}} & \best{\textbf{14.73}} & \best{\textbf{0.5501}} & \best{\textbf{0.4398}} & \best{\textbf{0.9715}} & \best{\textbf{0.9716}} & \best{\textbf{0.5609}} & \best{\textbf{15.04}} & \best{\textbf{0.5069}} & \best{\textbf{0.5073}} & \best{\textbf{0.9350}} & \best{\textbf{0.9546}} & \best{\textbf{0.4620}} \\
    \midrule
    \multicolumn{19}{l}{\textbf{2-view}} \\
    \midrule
    DepthSplat~\cite{depthsplat} & 10.45 & 0.3262 & 0.6167 & 0.8314 & 0.8585 & 0.2992 & 16.22 & 0.5382 & 0.4518 & 0.9012 & 0.9067 & 0.3779 & 13.87 & 0.4585 & 0.5195 & 0.8073 & 0.8474 & 0.3301 \\
    LVSM~\cite{lvsm} & \sbest{17.87} & \sbest{0.5986} & \sbest{0.4534} & \sbest{0.9467} & \sbest{0.9519} & 0.4064 & \best{\textbf{19.13}} & \best{\textbf{0.6789}} & \best{\textbf{0.3134}} & \best{\textbf{0.9747}} & \sbest{0.9730} & 0.4555 & \best{\textbf{17.79}} & \best{\textbf{0.5685}} & \best{\textbf{0.4265}} & \best{\textbf{0.9415}} & \best{\textbf{0.9569}} & 0.3628 \\
    Gen3C~\cite{gen3c} & \tbest{17.16} & \tbest{0.5927} & \tbest{0.4776} & 0.9149 & 0.9361 & 0.4263 & \tbest{17.81} & \tbest{0.6307} & \tbest{0.3882} & \tbest{0.9636} & \tbest{0.9651} & 0.4634 & \tbest{15.24} & \tbest{0.5055} & 0.5318 & 0.9119 & 0.9376 & 0.3668 \\
    GF~\cite{geometryforcing} & 12.67 & 0.3855 & 0.5998 & 0.7645 & 0.7925 & \tbest{0.4969} & 13.51 & 0.3991 & 0.4609 & 0.8785 & 0.8852 & \tbest{0.5374} & 12.06 & 0.3670 & 0.5666 & 0.7447 & 0.7946 & \tbest{0.4589} \\
    Aether~\cite{aether} & 14.28 & 0.5405 & 0.5498 & \tbest{0.9322} & \tbest{0.9426} & 0.4647 & 13.79 & 0.4884 & 0.5161 & 0.9491 & 0.9512 & 0.4685 & 14.53 & 0.4989 & \tbest{0.5153} & \tbest{0.9294} & \tbest{0.9496} & 0.4267 \\
    WVD~\cite{wvd} & 14.66 & 0.5101 & 0.5334 & 0.9246 & 0.9409 & \sbest{0.5306} & 16.27 & 0.5421 & 0.4098 & 0.9631 & 0.9646 & \sbest{0.5627} & 14.22 & 0.4605 & 0.5266 & 0.9116 & 0.9371 & \sbest{0.4680} \\
    Ours & \best{\textbf{18.01}} & \best{\textbf{0.6085}} & \best{\textbf{0.4371}} & \best{\textbf{0.9547}} & \best{\textbf{0.9597}} & \best{\textbf{0.5405}} & \sbest{18.88} & \sbest{0.6448} & \sbest{0.3256} & \sbest{0.9746} & \best{\textbf{0.9755}} & \best{\textbf{0.5685}} & \sbest{17.34} & \sbest{0.5581} & \sbest{0.4416} & \sbest{0.9385} & \sbest{0.9559} & \best{\textbf{0.4748}} \\
    \bottomrule
    \end{tabular}
    \end{adjustbox}
    \vspace{-0.25cm}
    \caption{\textbf{Quantitative Comparison of Appearance Generation.} We compare both 1-view and 2-view settings with camera conditions.}
    \label{tab: supp_3d_generation_rgb}
\end{table*}

\begin{table*}[htbp]
    \centering
    \footnotesize
    \begin{adjustbox}{max width=\linewidth}
    \begin{tabular}{c|lcccccccccc}
    \toprule
    \multirow{2}{*}{\rotatebox[origin=c]{90}{\scriptsize \textbf{Cond.}}} & \multirow{2}{*}{Method} &
    \multicolumn{5}{c}{RealEstate10K} &
    \multicolumn{5}{c}{DL3DV-10K} \\
    \cmidrule(lr){3-7} \cmidrule(lr){8-12}
    & & I2V Subj.~$\uparrow$ & I2V BG~$\uparrow$ & Aes.Q.~$\uparrow$ & I.Q.~$\uparrow$ & M.S.~$\uparrow$ & I2V Subj.~$\uparrow$ & I2V BG~$\uparrow$ & Aes.Q.~$\uparrow$ & I.Q.~$\uparrow$ & M.S.~$\uparrow$ \\
    \midrule
    \multirow{3}{*}{\rotatebox[origin=c]{90}{\textbf{1-view}}} &
    Aether~\cite{aether} & \tbest{0.9743} & \tbest{0.9770} & \tbest{0.5118} & \tbest{0.5060} & \tbest{0.9885} & \sbest{0.9377} & \sbest{0.9501} & \sbest{0.4704} & \tbest{0.4872} & \sbest{0.9600} \\ 
    & WVD~\cite{wvd} & \sbest{0.9815} & \sbest{0.9843} & \sbest{0.5125} & \sbest{0.5653} & \sbest{0.9895} & \tbest{0.9274} & \tbest{0.9412} & \tbest{0.4555} & \sbest{0.4916} & \tbest{0.9542} \\
    & Ours & \best{\textbf{0.9879}} & \best{\textbf{0.9890}} & \best{\textbf{0.5291}} & \best{\textbf{0.5761}} & \best{\textbf{0.9929}} & \best{\textbf{0.9461}} & \best{\textbf{0.9561}} & \best{\textbf{0.4727}} & \best{\textbf{0.5187}} & \best{\textbf{0.9701}} \\
    \midrule
    \multirow{3}{*}{\rotatebox[origin=c]{90}{\textbf{2-view}}} &
    Aether~\cite{aether} & \tbest{0.9852} & \tbest{0.9843} & \tbest{0.5278} & \tbest{0.5187} & \tbest{0.9923} & \sbest{0.9485} & \sbest{0.9521} & \sbest{0.4846} & \tbest{0.5026} & \sbest{0.9685} \\
    & WVD~\cite{wvd} & \sbest{0.9929} & \sbest{0.9923} & \sbest{0.5336} & \sbest{0.5973} & \sbest{0.9938} & \tbest{0.9403} & \tbest{0.9518} & \tbest{0.4760} & \sbest{0.5338} & \sbest{0.9685} \\
    & Ours & \best{\textbf{0.9949}} & \best{\textbf{0.9947}} & \best{\textbf{0.5369}} & \best{\textbf{0.6009}} & \best{\textbf{0.9947}} & \best{\textbf{0.9549}} & \best{\textbf{0.9576}} & \best{\textbf{0.4881}} & \best{\textbf{0.5357}} & \best{\textbf{0.9719}} \\
    \bottomrule
    \end{tabular}
    \end{adjustbox}
    \vspace{-0.25cm}
    \caption{\textbf{Quantitative Comparison of Appearance Generation} without camera conditions.}
    \label{tab: supp_wo_camera}
    \vspace{-0.3cm}
\end{table*}

\begin{table}[t]
    \centering
    \begin{adjustbox}{max width=\linewidth}
    \begin{tabular}{lcccccc}
    \toprule
    \multicolumn{1}{l}{\textbf{Cond.}} & \multicolumn{2}{c}{LLFF} & \multicolumn{2}{c}{Mip-NeRF 360} & \multicolumn{2}{c}{ScanNet++} \\
    \cmidrule(lr){1-1} \cmidrule(lr){2-3} \cmidrule(lr){4-5} \cmidrule(lr){6-7}
    Method & PSNR~$\uparrow$ & LPIPS~$\downarrow$ & PSNR~$\uparrow$ & LPIPS~$\downarrow$ & PSNR~$\uparrow$ & LPIPS~$\downarrow$ \\
    \midrule
    \multicolumn{7}{l}{\textbf{1-view}} \\
    \midrule
    LVSM~\cite{lvsm} & \sbest{12.39} & \sbest{0.5742} & \tbest{12.71} & \tbest{0.6328} & \sbest{15.25} & \sbest{0.4531} \\
    SEVA~\cite{seva} & \tbest{11.43} & 0.6562 & \sbest{12.77} & \sbest{0.5898} & \tbest{13.97} & \tbest{0.4688} \\
    Aether~\cite{aether} & 11.01 & \tbest{0.6133} & 11.06 & 0.6640 & 12.47 & 0.5351 \\
    Ours & \best{\textbf{13.19}} & \best{\textbf{0.5078}} &  \best{\textbf{13.17}} &  \best{\textbf{0.5703}} &  \best{\textbf{15.42}} & \best{\textbf{0.4420}} \\
    \midrule
    \multicolumn{7}{l}{\textbf{2-view}} \\
    \midrule
    LVSM~\cite{lvsm} & \best{\textbf{18.39}} &  \best{\textbf{0.3266}} & \best{\textbf{15.56}} & \sbest{0.5039} &  \best{\textbf{21.31}} &  \best{\textbf{0.2754}} \\
    SEVA~\cite{seva} & \tbest{16.86} & \tbest{0.3769} & \tbest{14.86} & \tbest{0.5080} & \tbest{18.07} & \tbest{0.3438} \\
    Aether~\cite{aether} & 13.66 & 0.4589 & 11.43 & 0.6289 & 17.14 & 0.4018 \\
    Ours & \sbest{17.66} & \sbest{0.3496} & \sbest{15.32} &  \best{\textbf{0.4941}} & \sbest{20.11} & \sbest{0.2964} \\
    \bottomrule
    \end{tabular}
    \end{adjustbox}
    \vspace{-0.25cm}
    \caption{\textbf{Quantitative Comparison of Appearance Generation on Out-of-Distribution Datasets.} We compare both 1-view and 2-view settings with camera conditions.}
    \label{tab: supp_ood_3d_generation_rgb}
\end{table}

\begin{table}[t]
    \centering
    \footnotesize
    \begin{adjustbox}{max width=\linewidth}
    \begin{tabular}{lcc}
    \toprule
    \multirow{2}{*}{Method} & \multicolumn{1}{c}{RealEstate10K} & \multicolumn{1}{c}{WildRGB-D} \\
    \cmidrule(lr){2-2} \cmidrule(lr){3-3}
    & AUC@30~$\uparrow$ & AUC@30~$\uparrow$ \\
    \midrule
    Aether & \tbest{0.7291} & \tbest{0.7303} \\
    VGGT & \best{\textbf{0.8387}} & \best{\textbf{0.8406}} \\
    Ours & \sbest{0.8265} & \sbest{0.8391} \\
    \bottomrule
    \end{tabular}
    \end{adjustbox}
    \vspace{-0.2cm}
    \caption{\textbf{Quantitative Comparison of Camera Pose Estimation} in feed-forward 3D reconstruction.}
    \label{tab: feed_forward_cam}
\end{table}

\begin{figure*}[t]
    \centering

    \def\mywidth{2.0cm}
    \begin{tabular}{P{1.4cm}P{\mywidth}P{\mywidth}P{\mywidth}P{\mywidth}P{\mywidth}P{\mywidth}}

        \begin{minipage}[c]{0.18\textwidth}
            \includegraphics[height=0.49\linewidth]{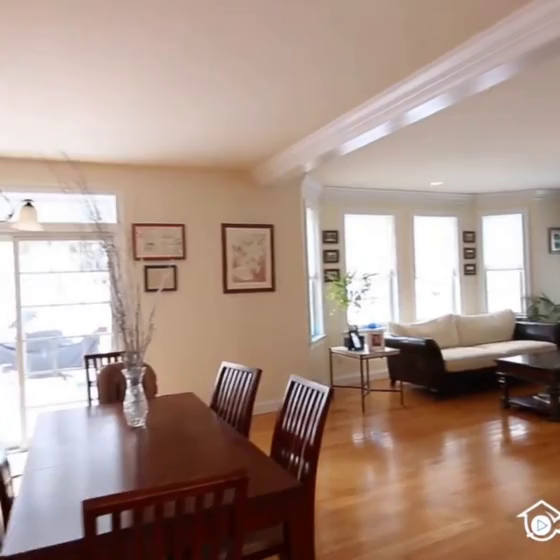}
        \end{minipage}
        &
        \includegraphics[width=1.13\linewidth]{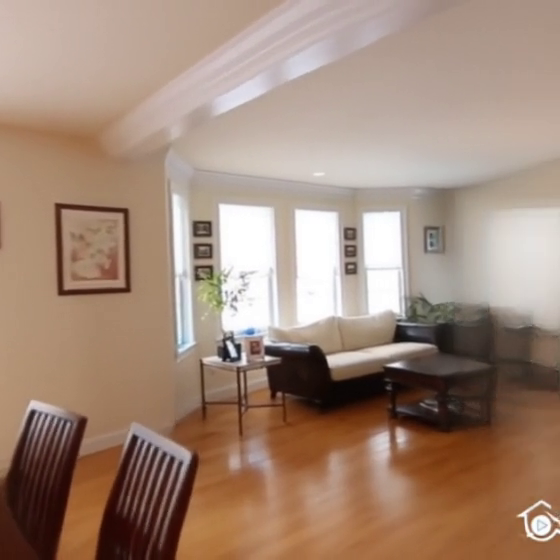} &
        \includegraphics[width=1.13\linewidth]{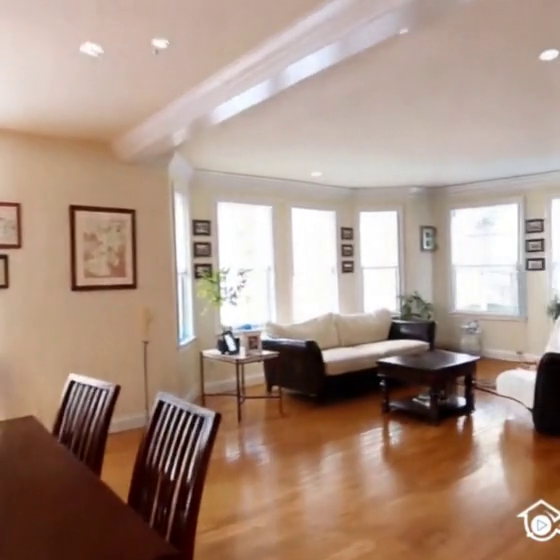} &
        \includegraphics[width=1.13\linewidth]{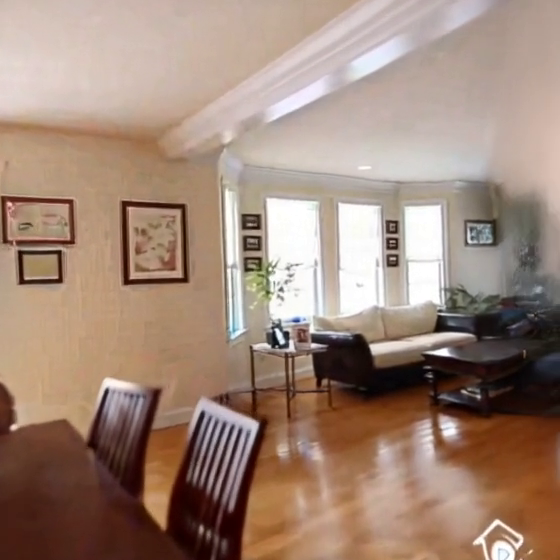} &
        \includegraphics[width=1.13\linewidth]{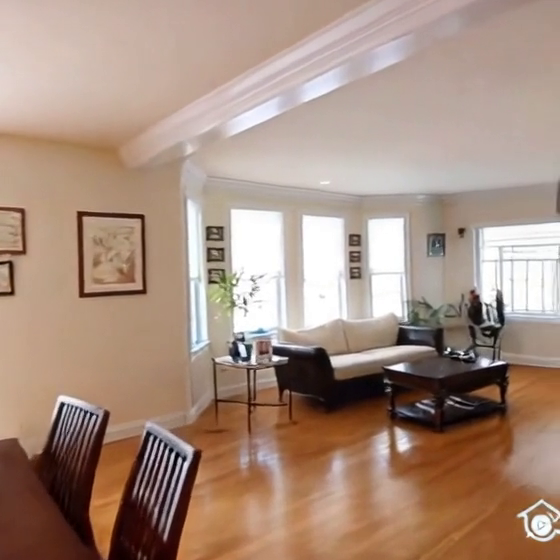} &
        \includegraphics[width=1.13\linewidth]{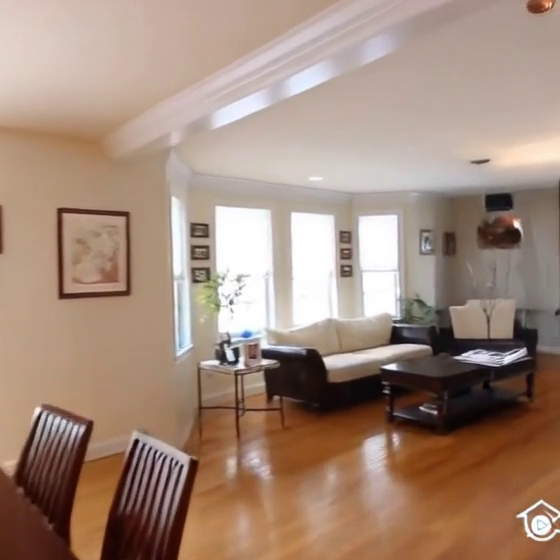} &
        \includegraphics[width=1.13\linewidth]{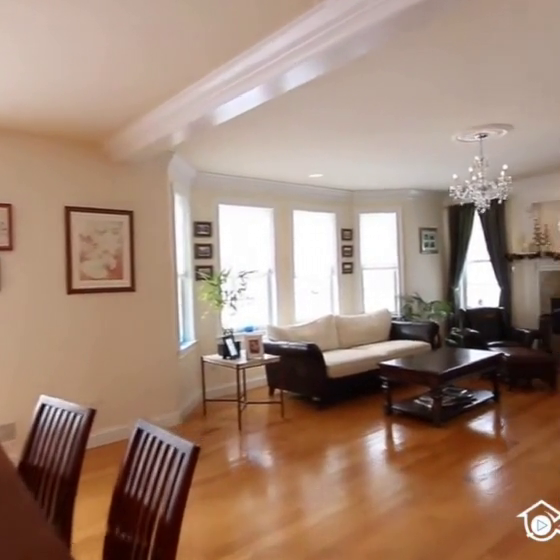}
        \\ [-2.5pt]

        \begin{minipage}[c]{0.18\textwidth}
            \includegraphics[height=0.49\linewidth]{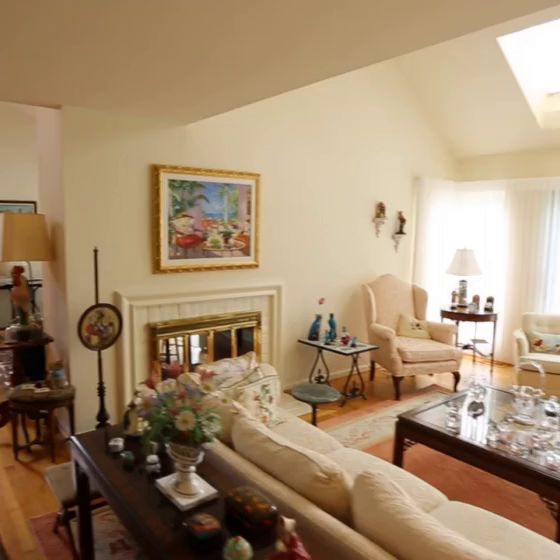}
        \end{minipage}
        &
        \includegraphics[width=1.13\linewidth]{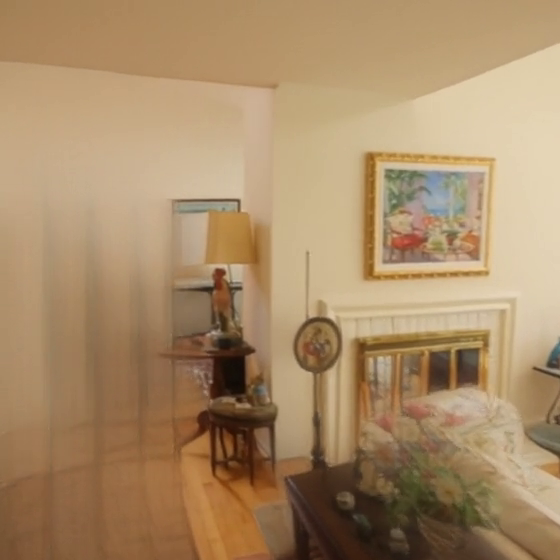} &
        \includegraphics[width=1.13\linewidth]{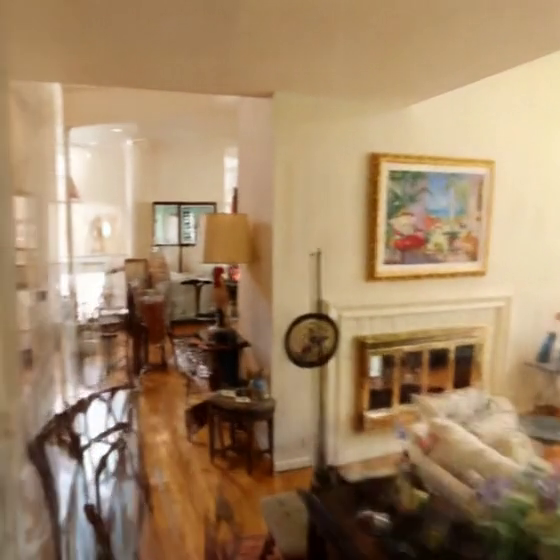} &
        \includegraphics[width=1.13\linewidth]{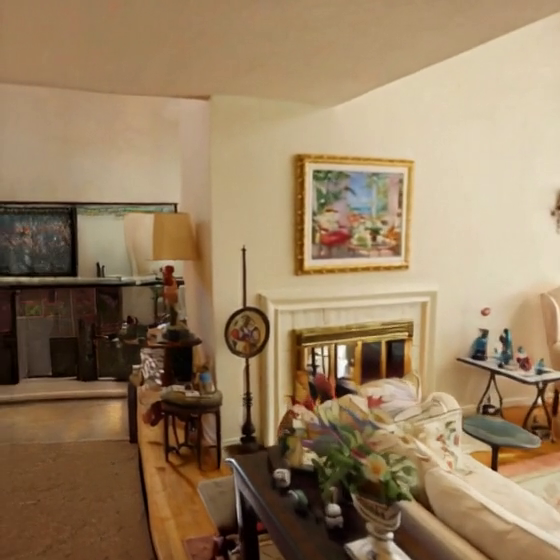} &
        \includegraphics[width=1.13\linewidth]{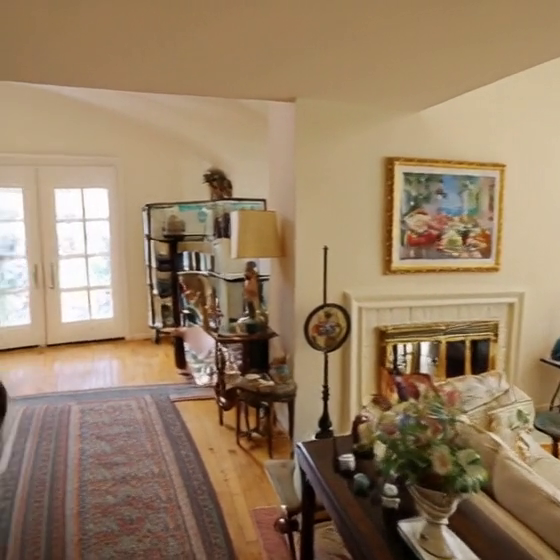} &
        \includegraphics[width=1.13\linewidth]{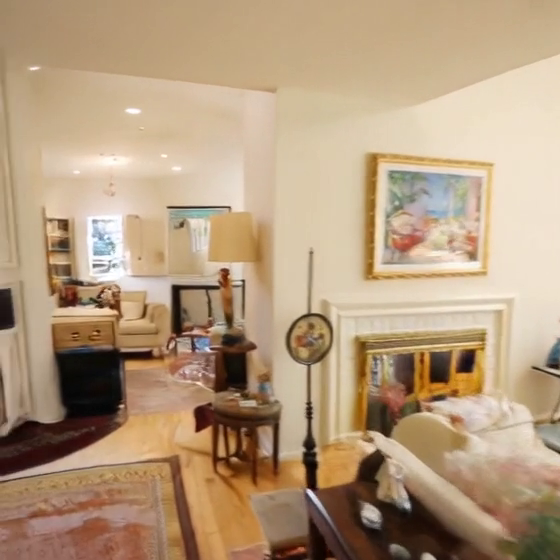} &
        \includegraphics[width=1.13\linewidth]{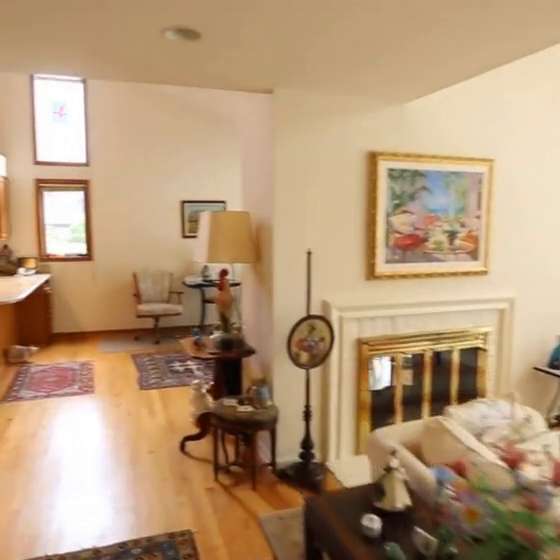}
        \\ [-2.5pt]

        \begin{minipage}[c]{0.18\textwidth}
            \includegraphics[height=0.49\linewidth]{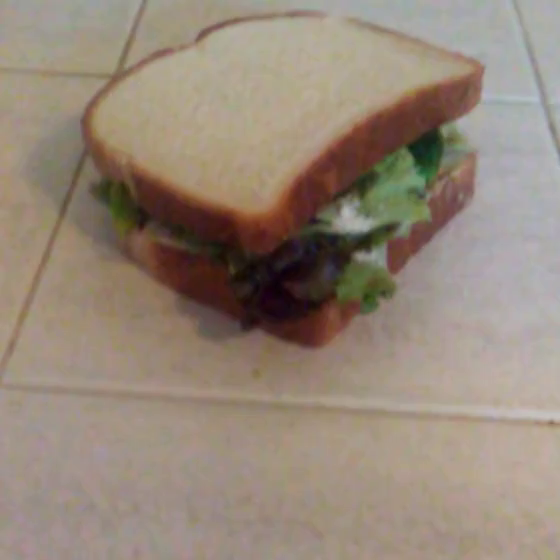}
        \end{minipage}
        &
        \includegraphics[width=1.13\linewidth]{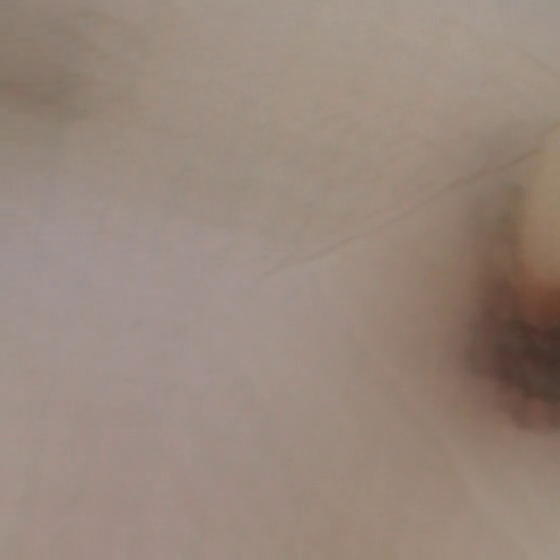} &
        \includegraphics[width=1.13\linewidth]{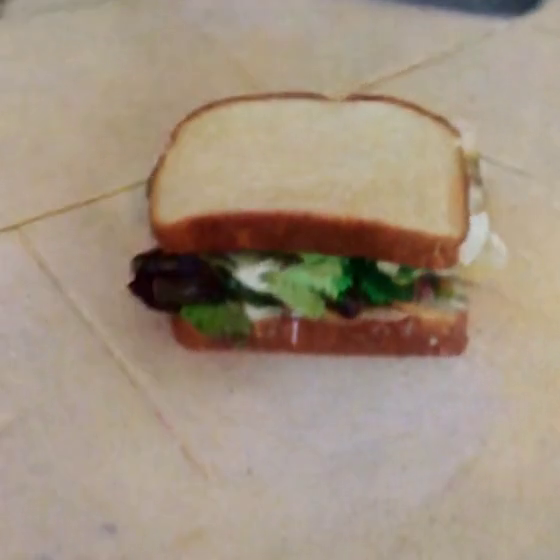} &
        \includegraphics[width=1.13\linewidth]{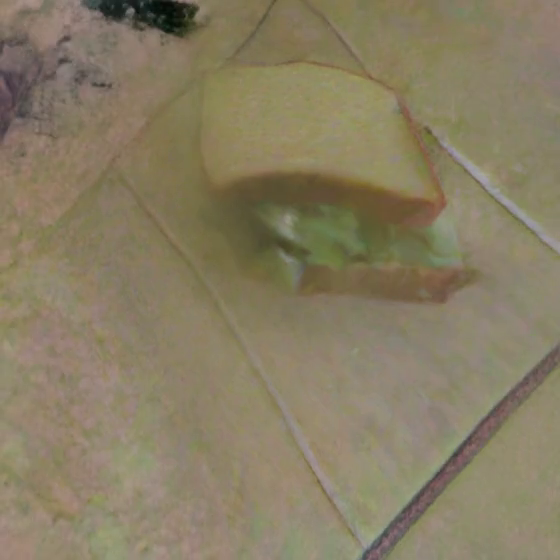} &
        \includegraphics[width=1.13\linewidth]{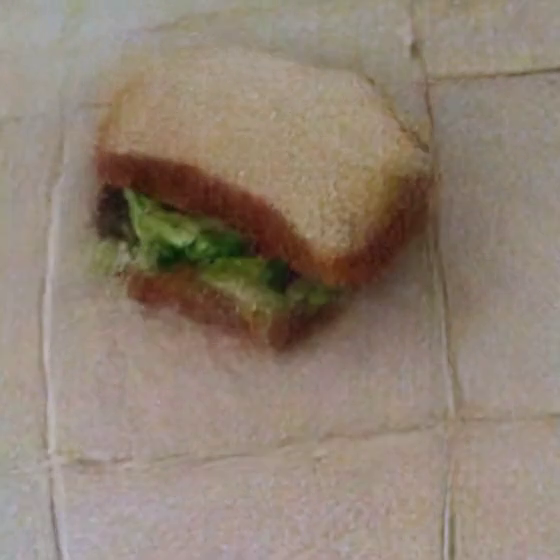} &
        \includegraphics[width=1.13\linewidth]{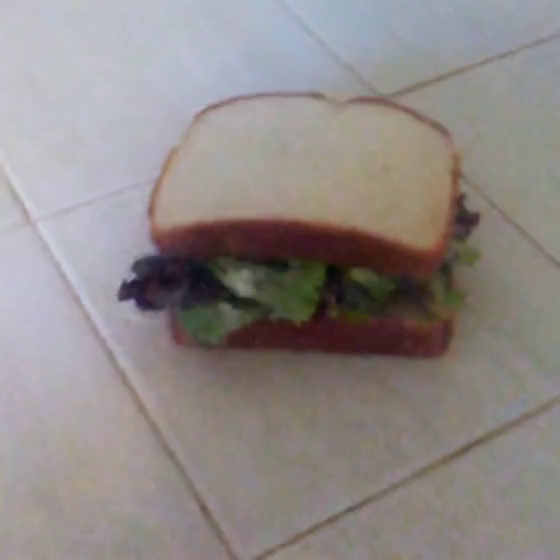} &
        \includegraphics[width=1.13\linewidth]{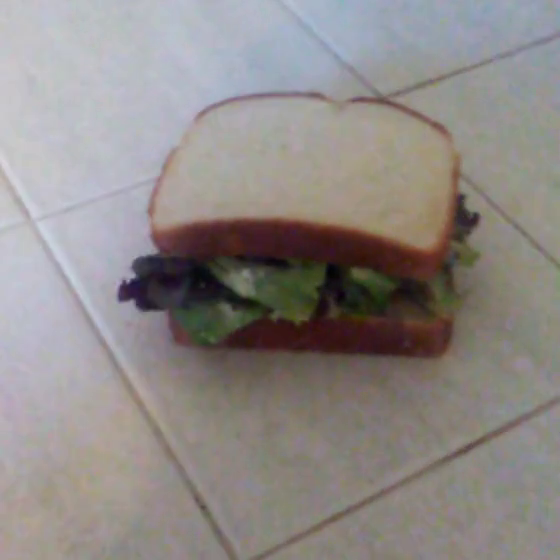}
        \\ [-2.5pt]

        \begin{minipage}[c]{0.18\textwidth}
            \includegraphics[height=0.49\linewidth]{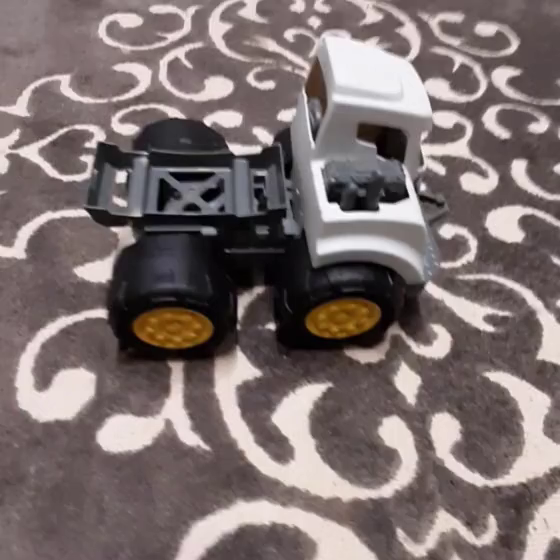}
        \end{minipage}
        &
        \includegraphics[width=1.13\linewidth]{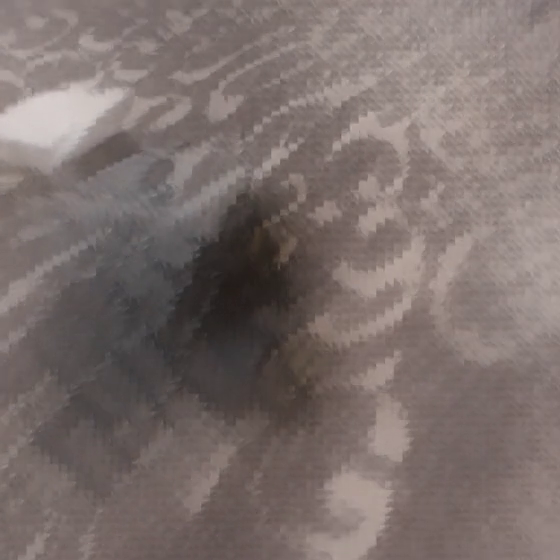} &
        \includegraphics[width=1.13\linewidth]{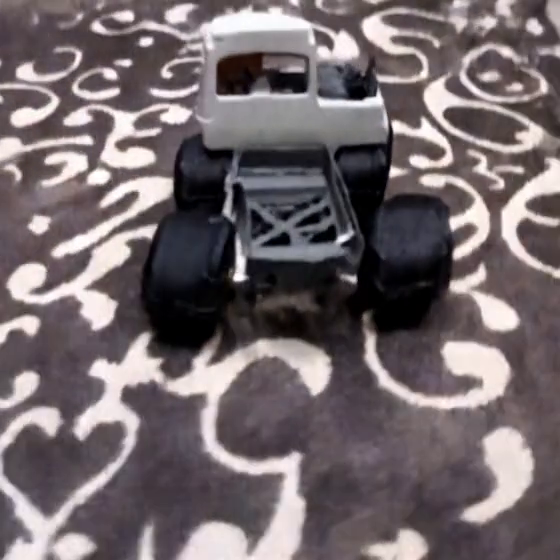} &
        \includegraphics[width=1.13\linewidth]{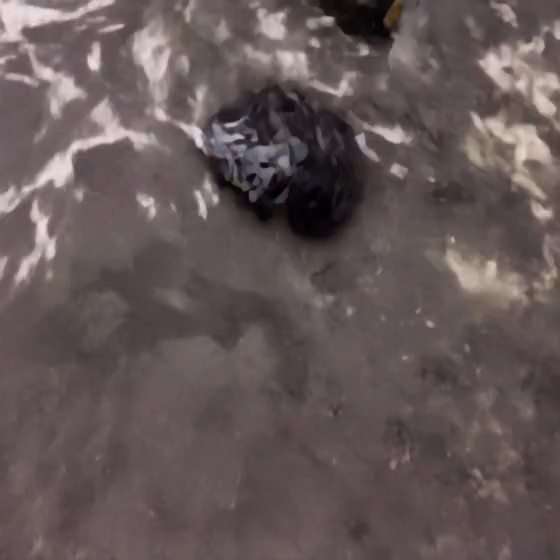} &
        \includegraphics[width=1.13\linewidth]{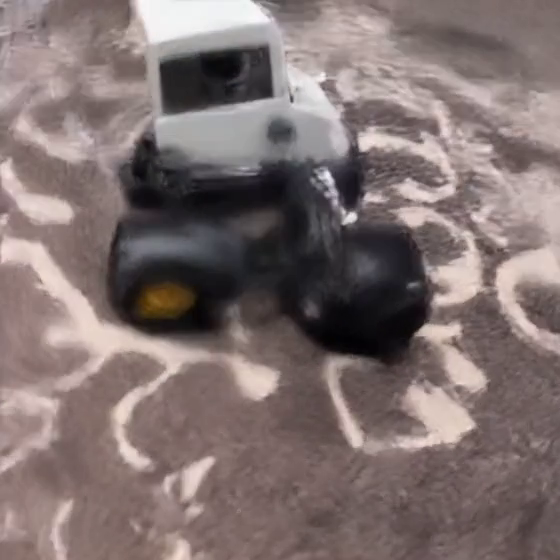} &
        \includegraphics[width=1.13\linewidth]{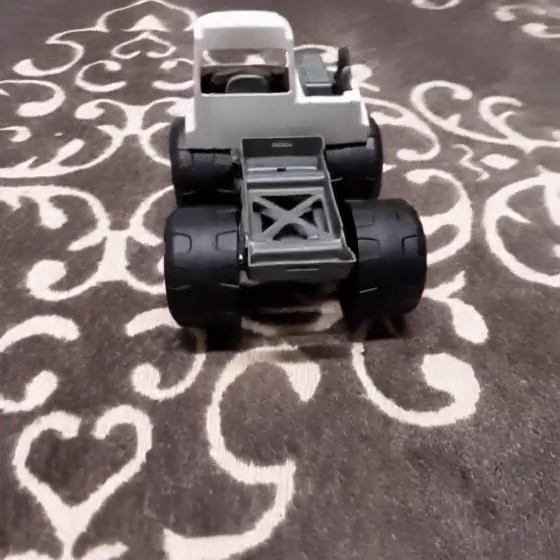} &
        \includegraphics[width=1.13\linewidth]{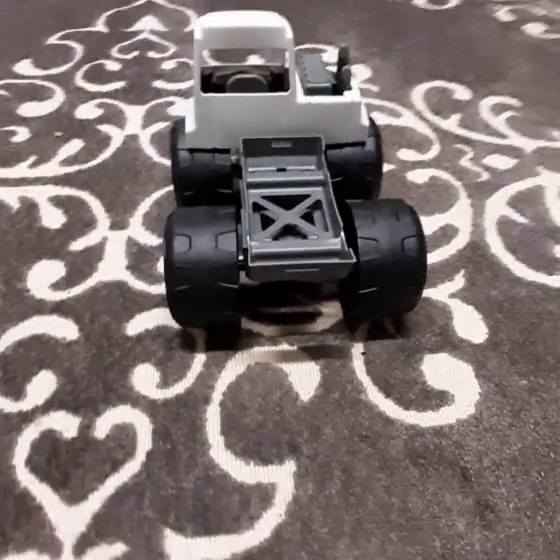}
        \\ [-2.5pt]

        \footnotesize Input &
        \footnotesize LVSM~\cite{lvsm} &
        \footnotesize Gen3C~\cite{gen3c} &
        \footnotesize Aether~\cite{aether} &
        \footnotesize WVD~\cite{wvd} & 
        \footnotesize Ours &
        \footnotesize Ground Truth
    \end{tabular}
    \vspace{-0.15cm}
    \caption{\textbf{Qualitative Comparison of Novel View Synthesis} in 1-view setting with camera conditions.}
    \label{fig: supp_single_to_3d}
    \vspace{-0.15cm}
\end{figure*}

\begin{figure*}[htbp]
    \centering

    \def\mywidth{2.0cm}
    \begin{tabular}{P{1.4cm}P{\mywidth}P{\mywidth}P{\mywidth}P{\mywidth}P{\mywidth}P{\mywidth}}

        \begin{minipage}[c]{0.18\textwidth}
            \includegraphics[height=0.49\linewidth]{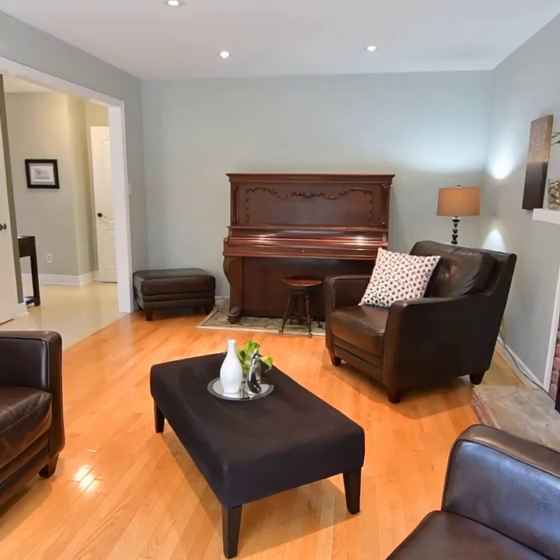}
        \end{minipage}
        &
        \includegraphics[width=1.13\linewidth]{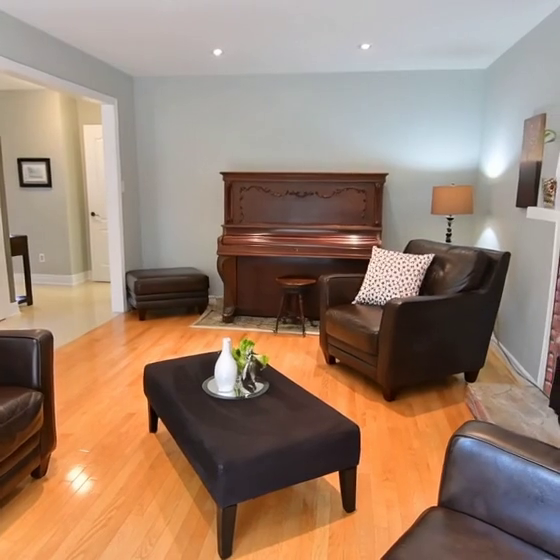} &
        \includegraphics[width=1.13\linewidth]{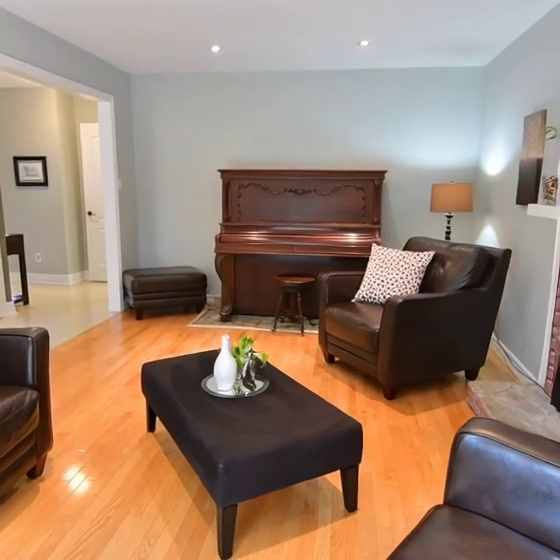} &
        \includegraphics[width=1.13\linewidth]{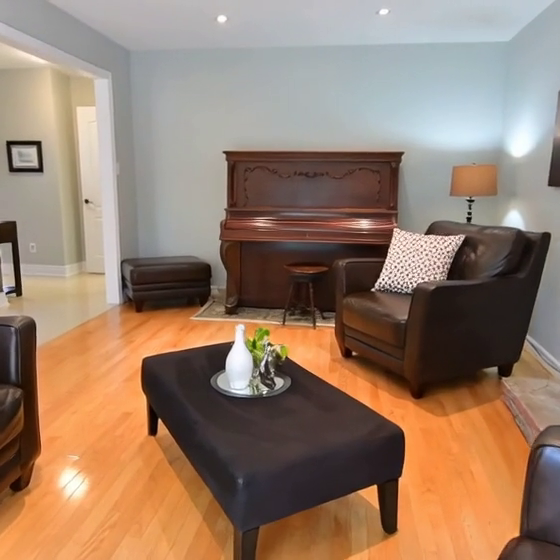} &
        \includegraphics[width=1.13\linewidth]{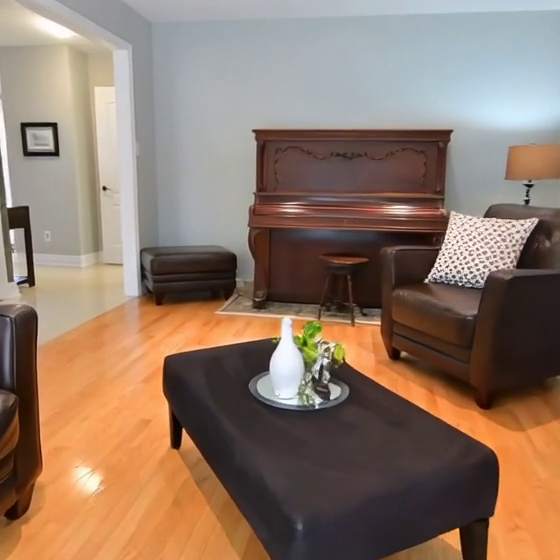} &
        \includegraphics[width=1.13\linewidth]{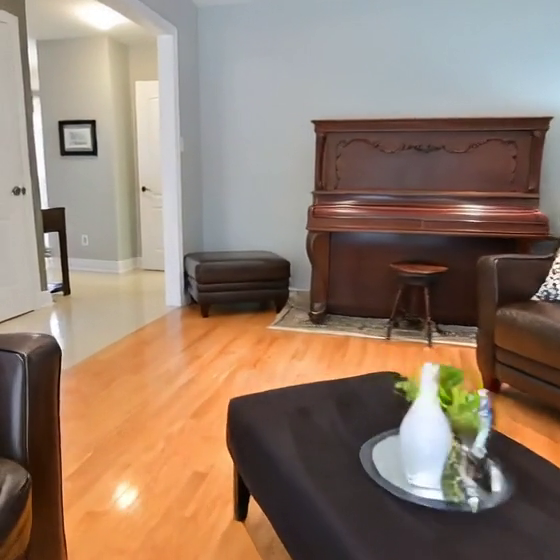} &
        \includegraphics[width=1.13\linewidth]{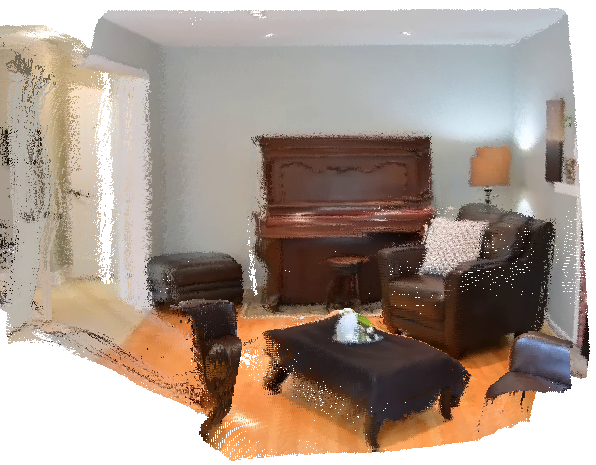}
        \\ [-2.5pt]

        \begin{minipage}[c]{0.18\textwidth}
            \includegraphics[height=0.49\linewidth]{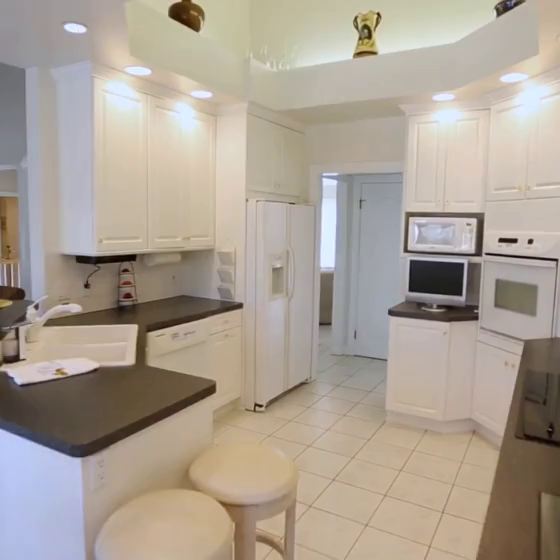}
        \end{minipage}
        &
        \includegraphics[width=1.13\linewidth]{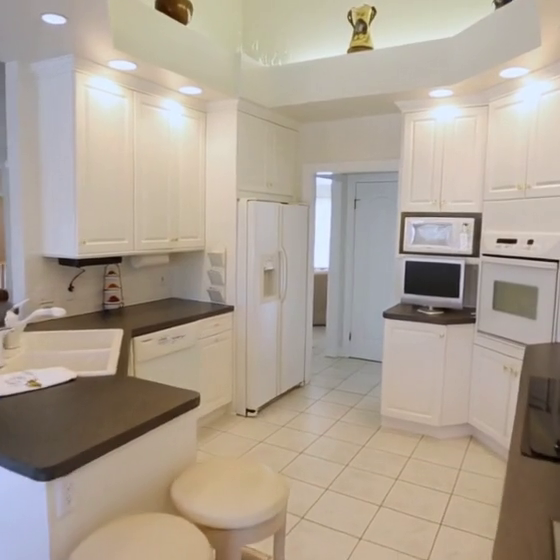} &
        \includegraphics[width=1.13\linewidth]{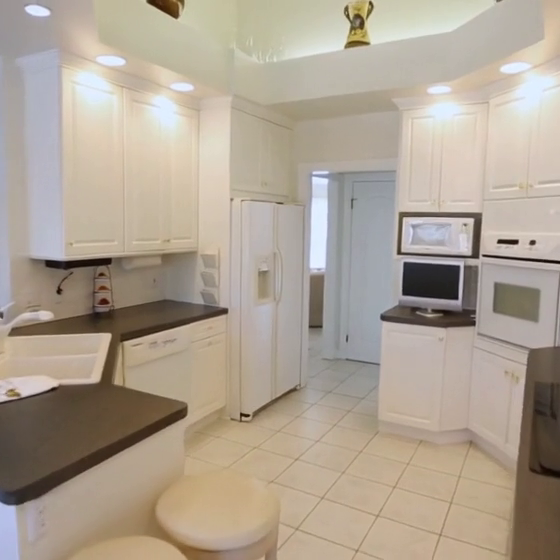} &
        \includegraphics[width=1.13\linewidth]{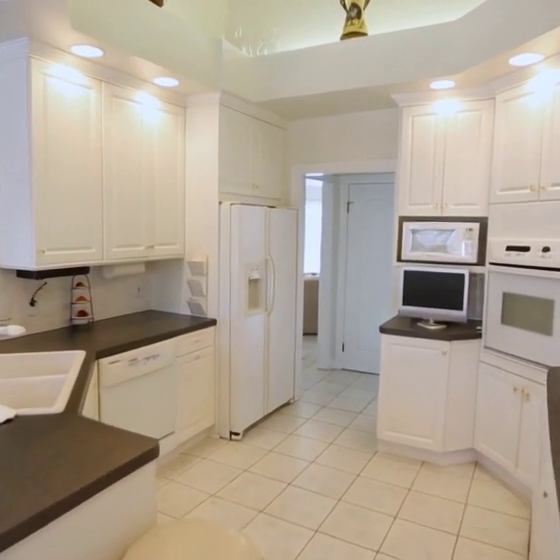} &
        \includegraphics[width=1.13\linewidth]{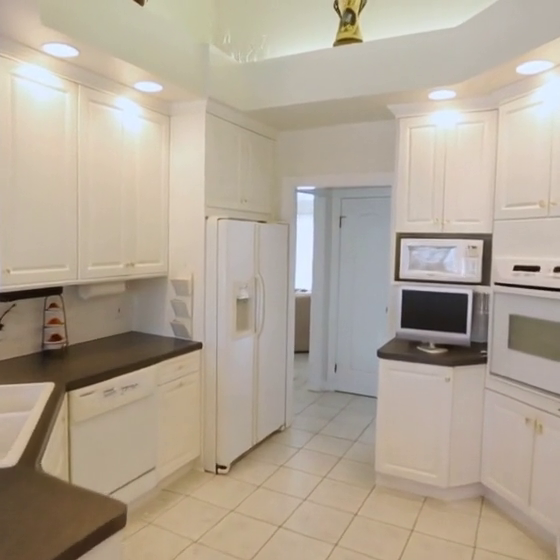} &
        \includegraphics[width=1.13\linewidth]{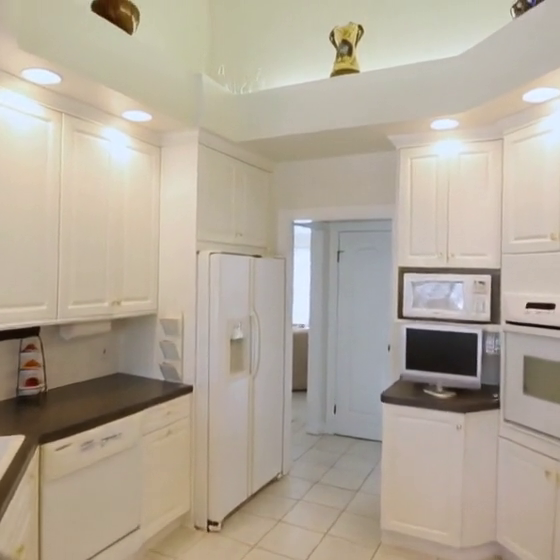} &
        \includegraphics[width=0.9\linewidth]{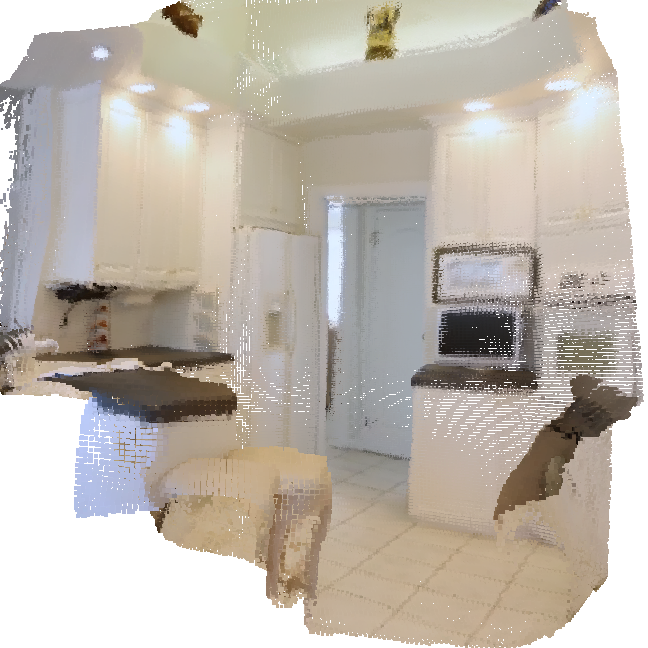}
        \\ [-2.5pt]

        \begin{minipage}[c]{0.18\textwidth}
            \includegraphics[height=0.49\linewidth]{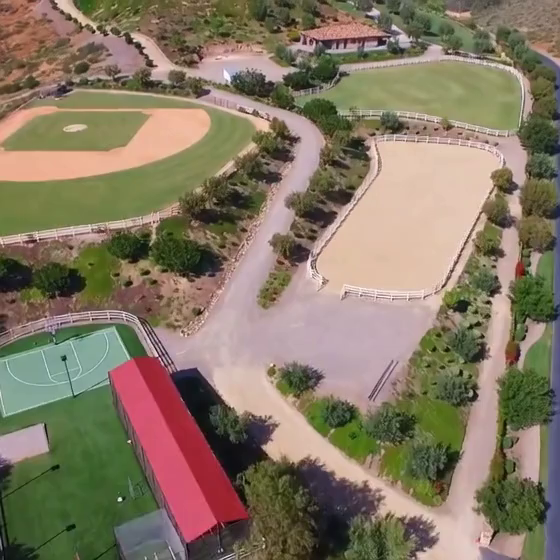}
        \end{minipage}
        &
        \includegraphics[width=1.13\linewidth]{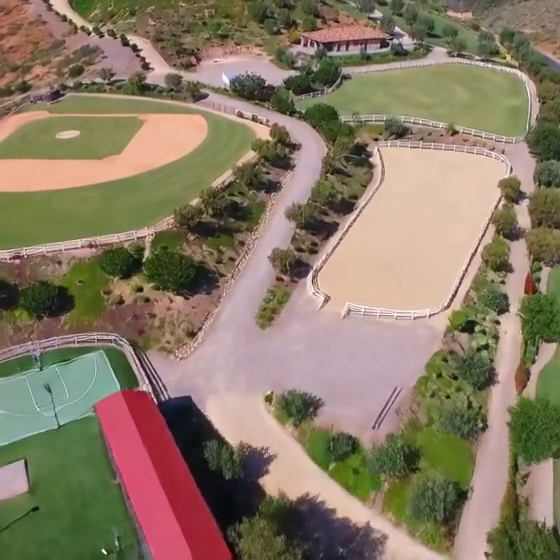} &
        \includegraphics[width=1.13\linewidth]{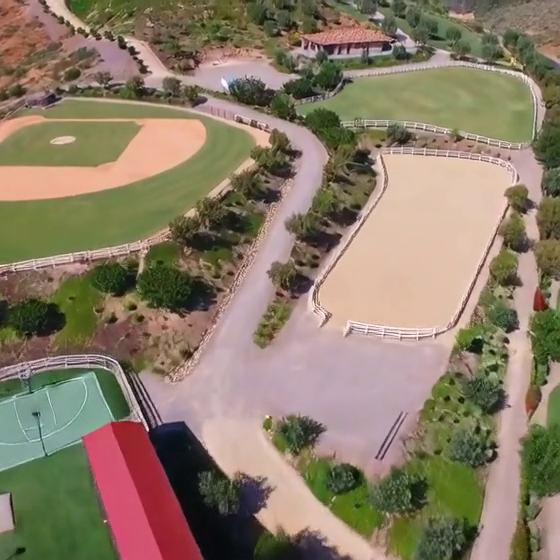} &
        \includegraphics[width=1.13\linewidth]{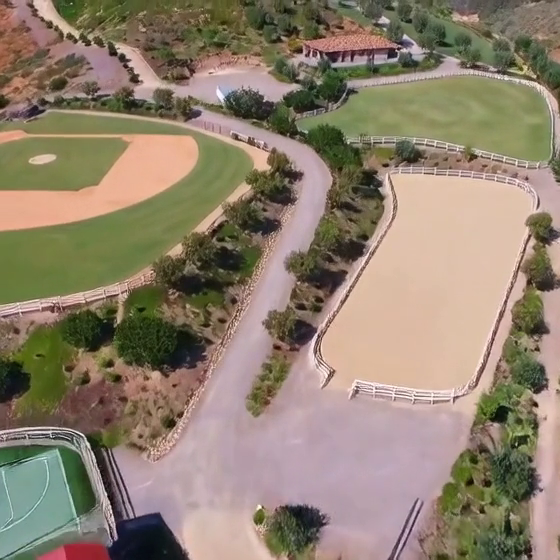} &
        \includegraphics[width=1.13\linewidth]{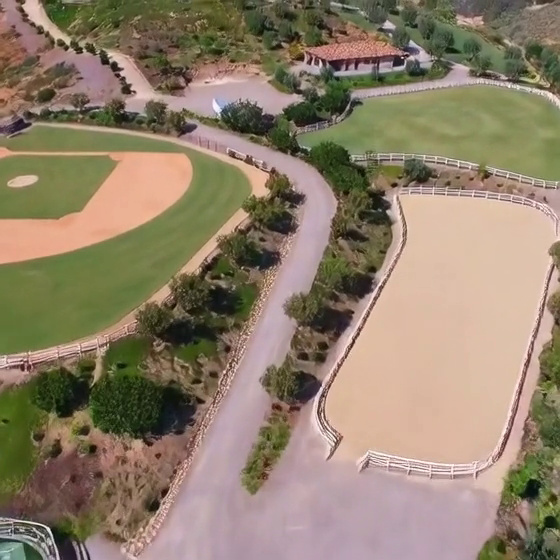} &
        \includegraphics[width=1.13\linewidth]{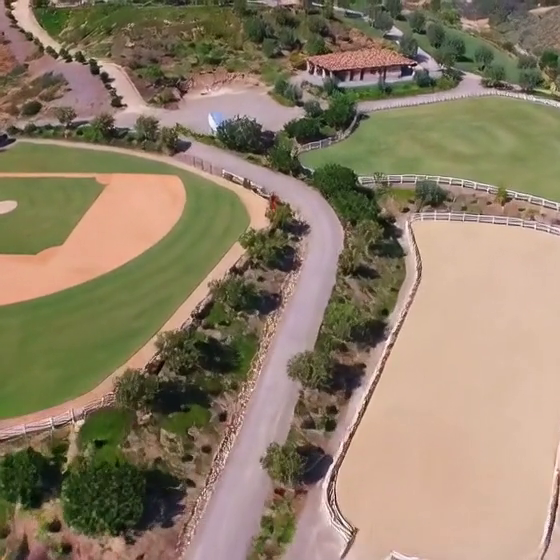} &
        \includegraphics[width=0.9\linewidth]{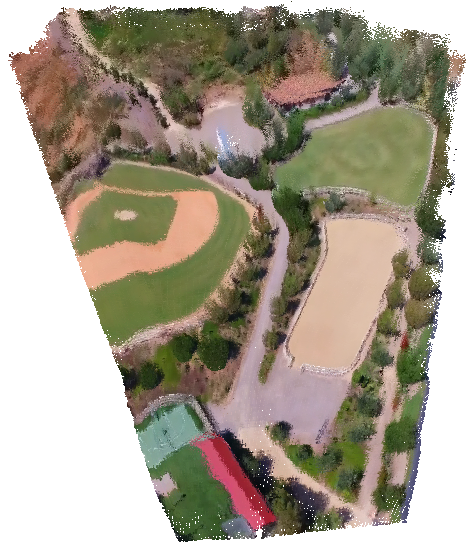}
        \\ [-2.5pt]

        \begin{minipage}[c]{0.18\textwidth}
            \includegraphics[height=0.49\linewidth]{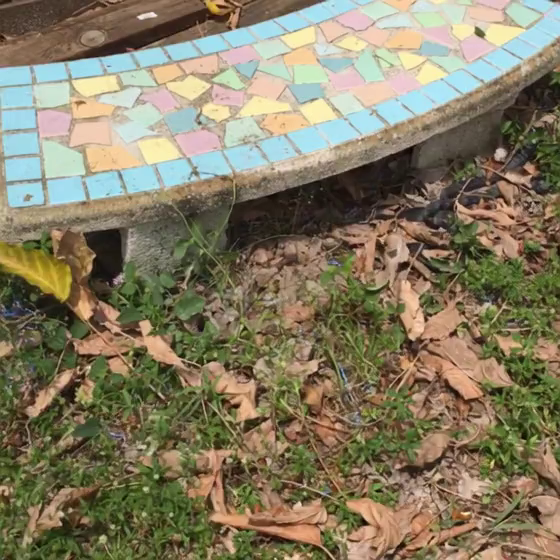}
        \end{minipage}
        &
        \includegraphics[width=1.13\linewidth]{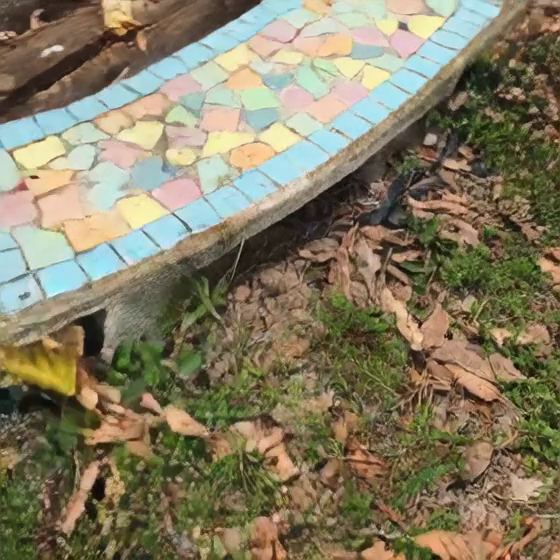} &
        \includegraphics[width=1.13\linewidth]{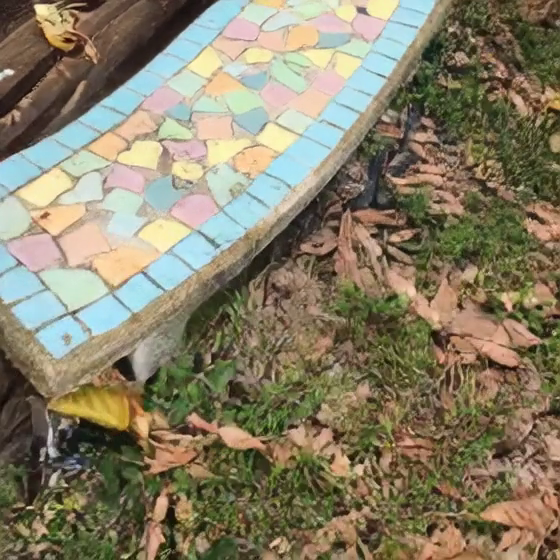} &
        \includegraphics[width=1.13\linewidth]{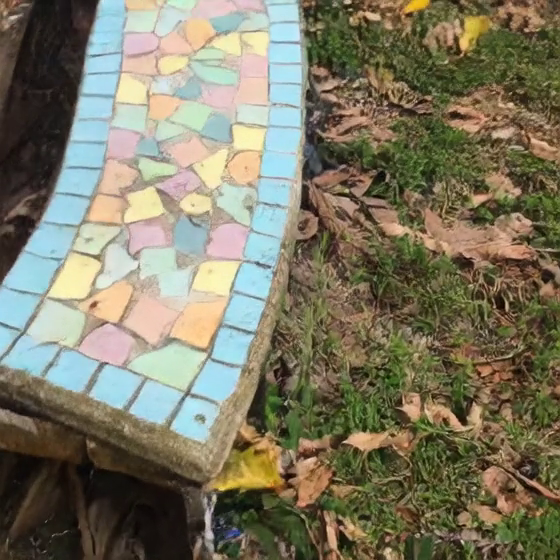} &
        \includegraphics[width=1.13\linewidth]{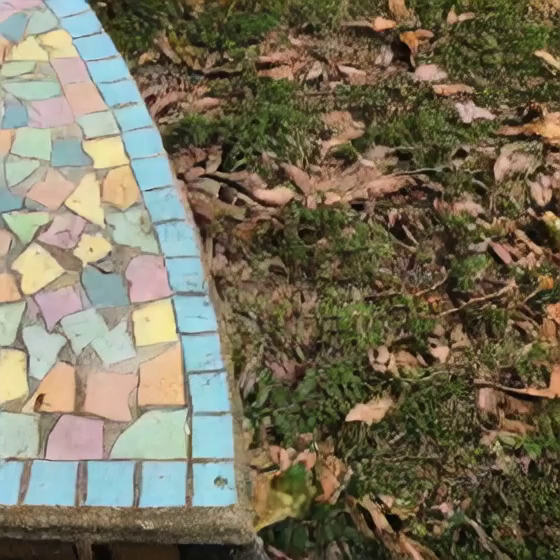} &
        \includegraphics[width=1.13\linewidth]{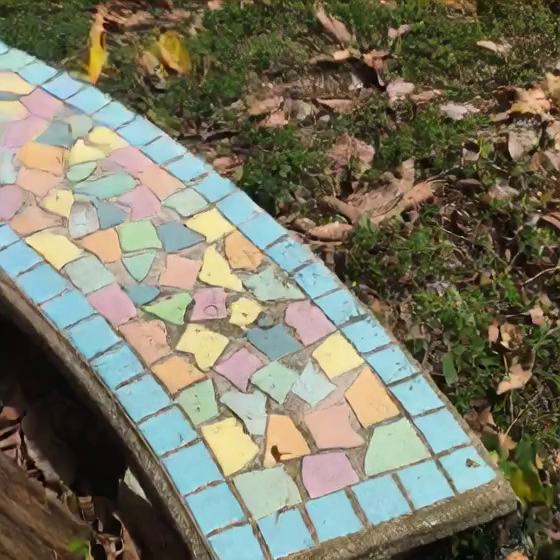} &
        \includegraphics[width=1.13\linewidth]{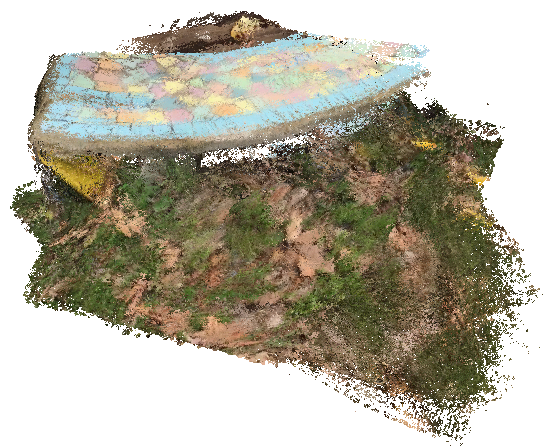}
        \\ [-2.5pt]

        \footnotesize Input &
        \multicolumn{5}{c}{\footnotesize Generated Frames} &
        \footnotesize Point Cloud
    \end{tabular}
    \vspace{-0.15cm}
    \caption{\textbf{More Qualitative Results} in 1-view setting with camera conditions.}
    \label{fig: more_1view_gen_w_camera}
    \vspace{-0.4cm}
\end{figure*}

\begin{figure*}[ht]
    \centering

    \def\mywidth{2.0cm}
    \begin{tabular}{P{0.7cm}P{\mywidth}P{\mywidth}P{\mywidth}P{\mywidth}P{\mywidth}P{\mywidth}}

        \begin{minipage}[c]{0.18\textwidth}
            \includegraphics[height=0.25\linewidth]{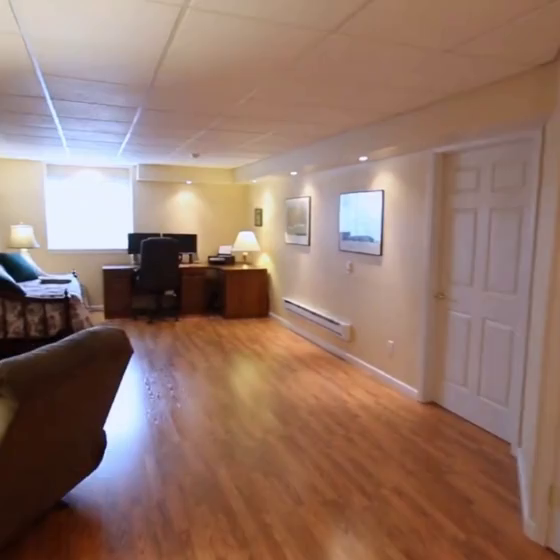} \\
            \includegraphics[height=0.25\linewidth]{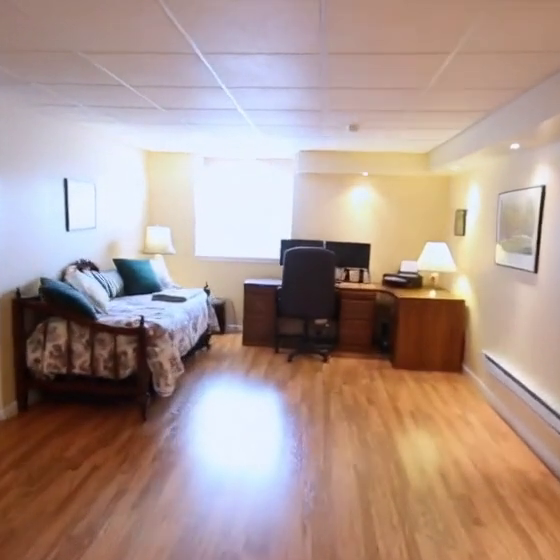}
        \end{minipage}
        &
        \includegraphics[width=1.13\linewidth]{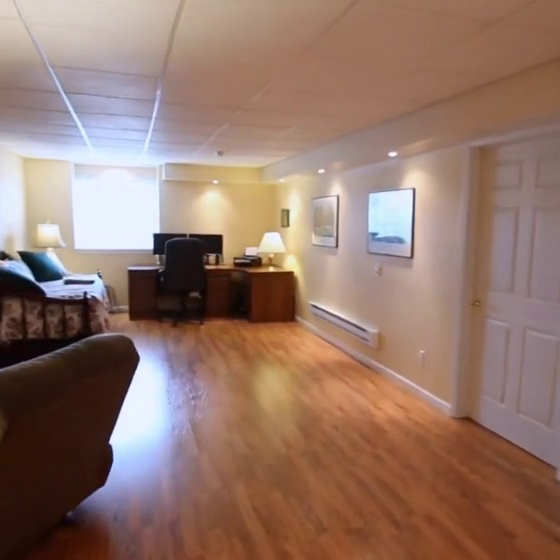} &
        \includegraphics[width=1.13\linewidth]{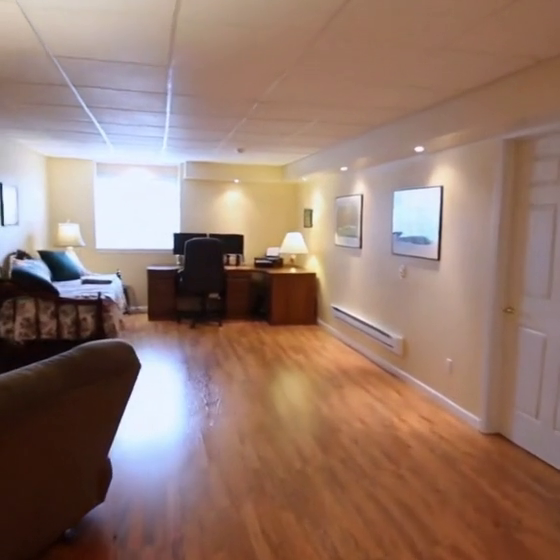} &
        \includegraphics[width=1.13\linewidth]{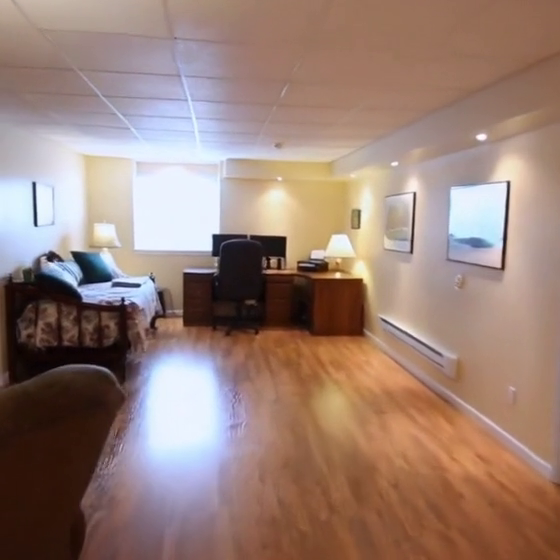} &
        \includegraphics[width=1.13\linewidth]{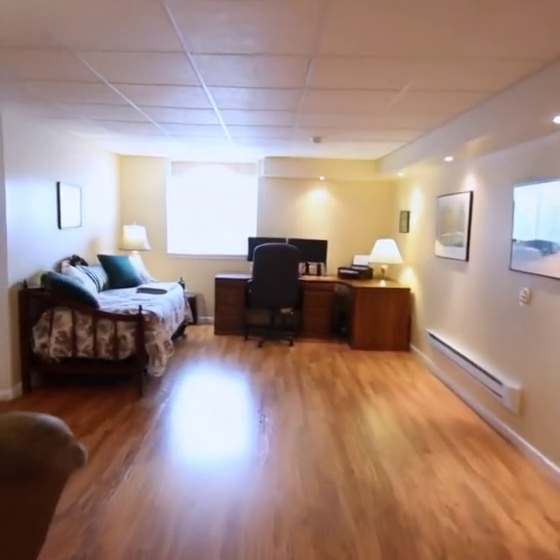} &
        \includegraphics[width=1.13\linewidth]{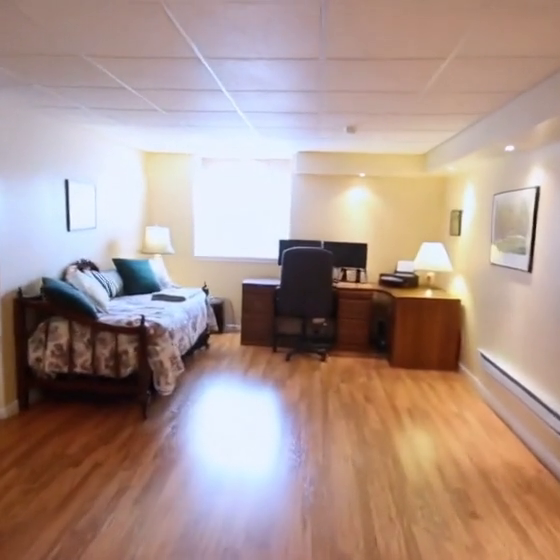} &
        \includegraphics[width=1.13\linewidth]{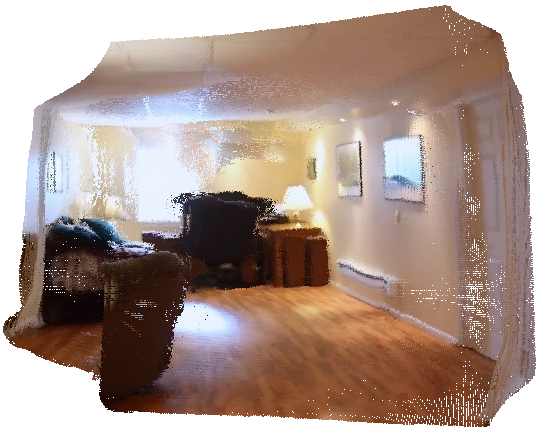}
        \\ [-2.5pt]

        \begin{minipage}[c]{0.18\textwidth}
            \includegraphics[height=0.25\linewidth]{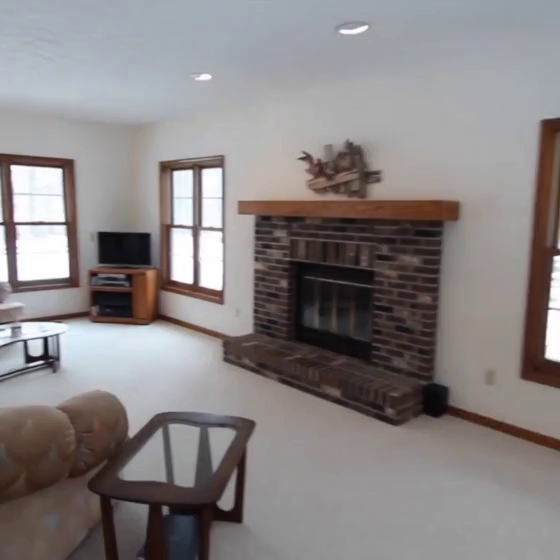} \\
            \includegraphics[height=0.25\linewidth]{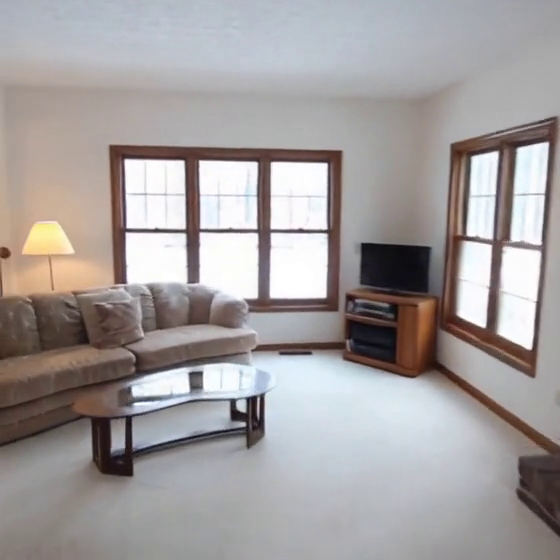}
        \end{minipage}
        &
        \includegraphics[width=1.13\linewidth]{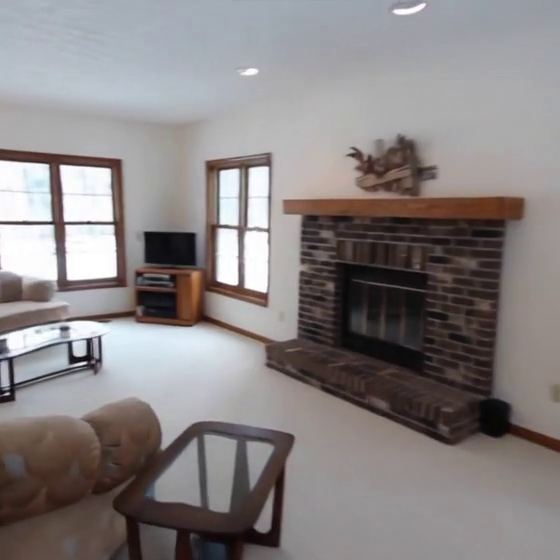} &
        \includegraphics[width=1.13\linewidth]{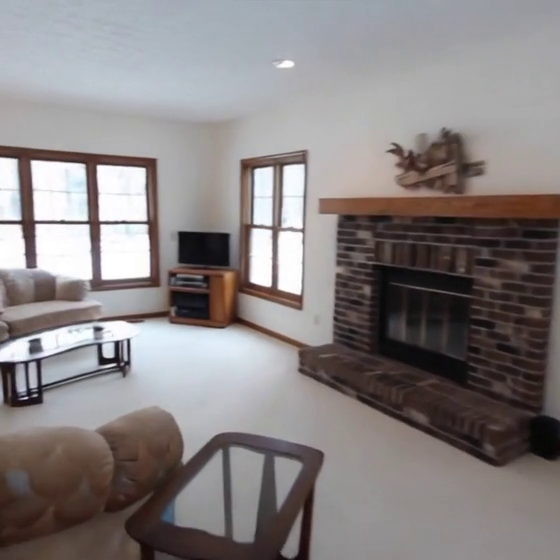} &
        \includegraphics[width=1.13\linewidth]{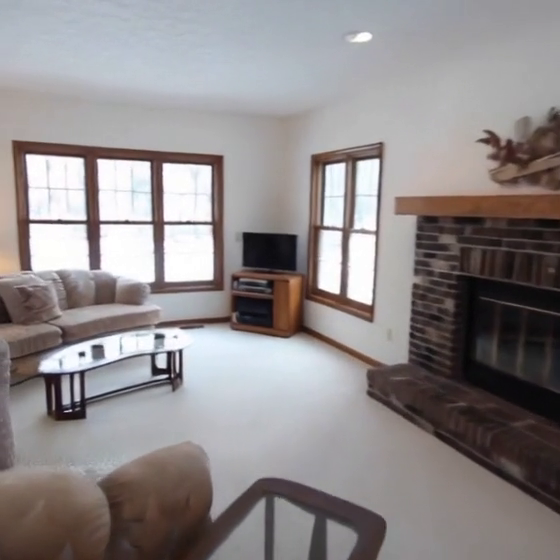} &
        \includegraphics[width=1.13\linewidth]{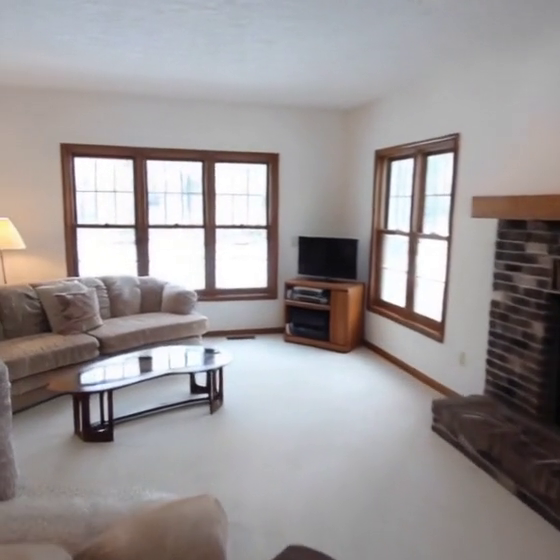} &
        \includegraphics[width=1.13\linewidth]{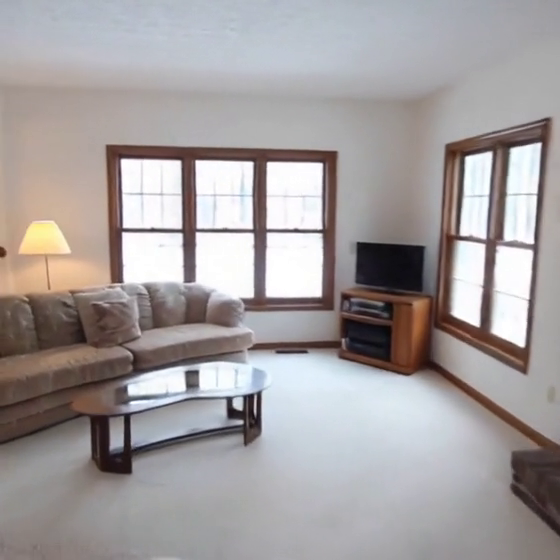} &
        \includegraphics[width=1.13\linewidth]{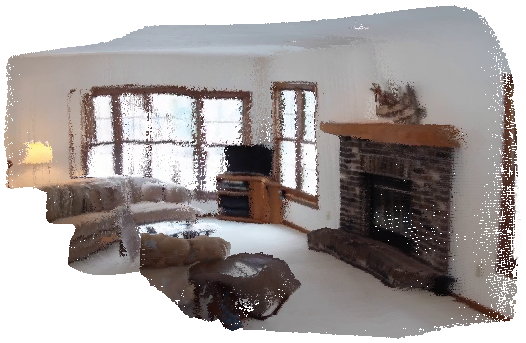}
        \\ [-2.5pt]

        \begin{minipage}[c]{0.18\textwidth}
            \includegraphics[height=0.25\linewidth]{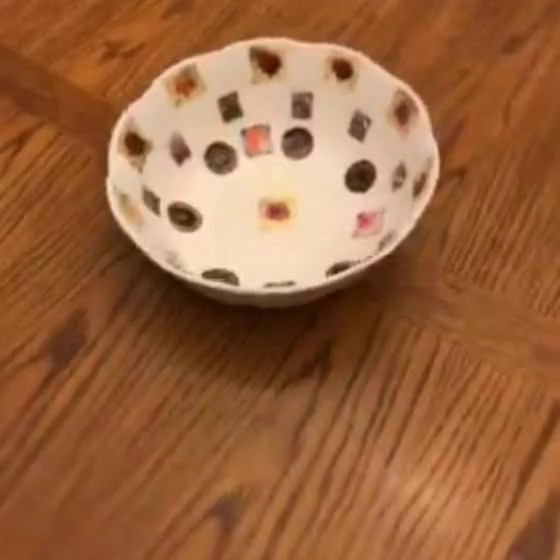} \\
            \includegraphics[height=0.25\linewidth]{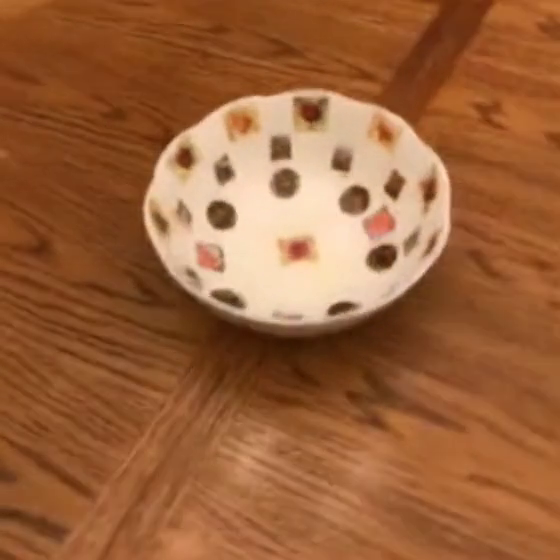}
        \end{minipage}
        &
        \includegraphics[width=1.13\linewidth]{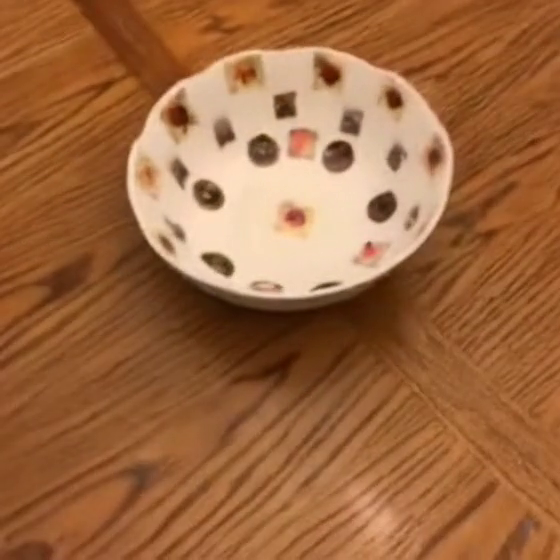} &
        \includegraphics[width=1.13\linewidth]{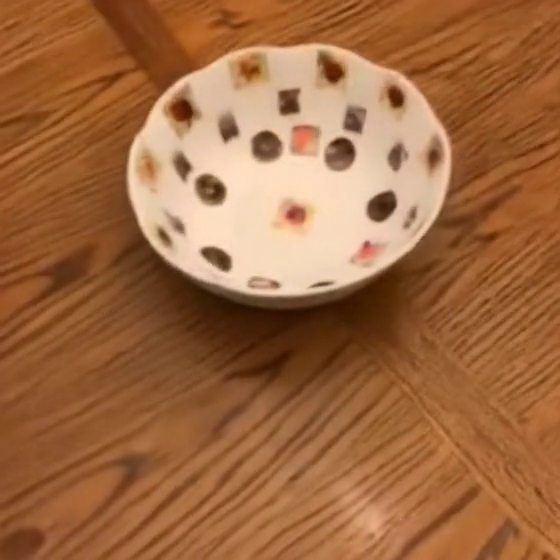} &
        \includegraphics[width=1.13\linewidth]{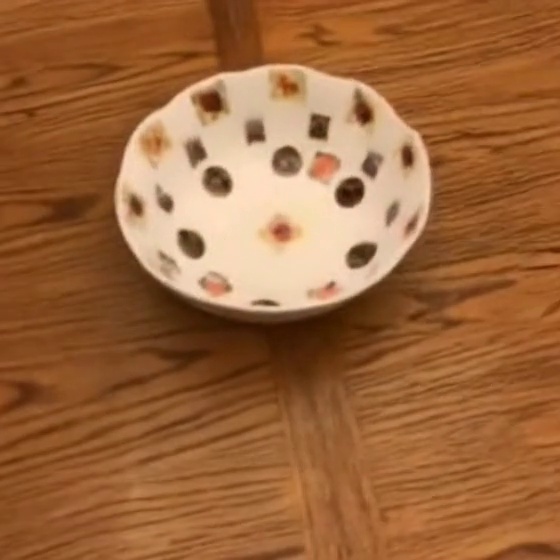} &
        \includegraphics[width=1.13\linewidth]{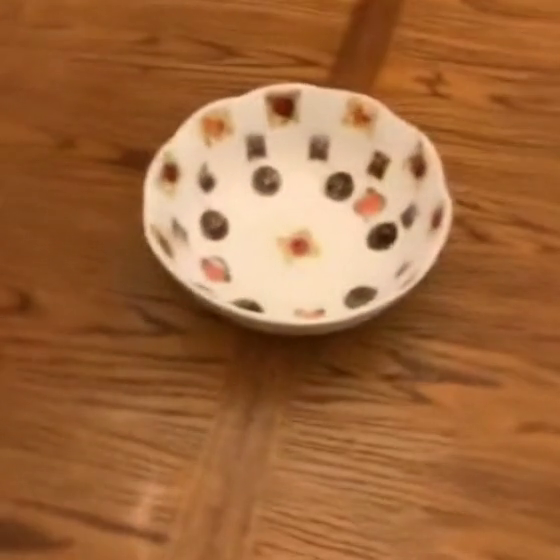} &
        \includegraphics[width=1.13\linewidth]{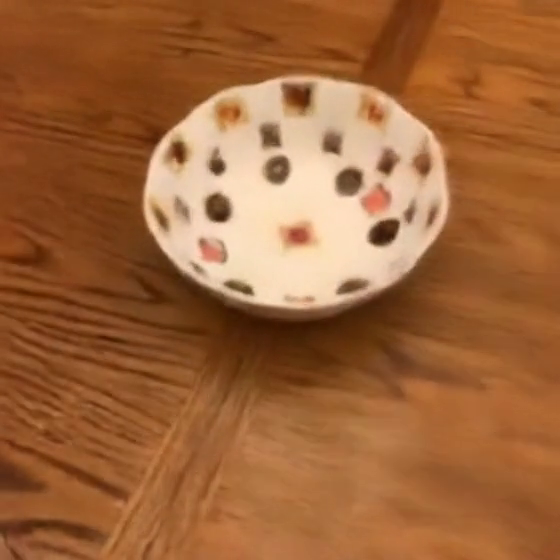} &
        \includegraphics[width=1.13\linewidth]{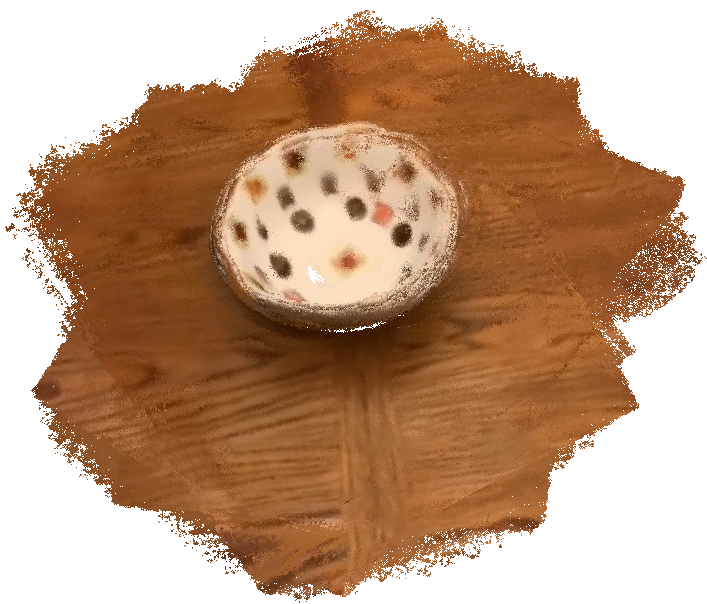}
        \\ [-2.5pt]

        \footnotesize Input &
        \multicolumn{5}{c}{\footnotesize Generated Frames} &
        \footnotesize Point Cloud
    \end{tabular}
    \vspace{-0.2cm}
    \caption{\textbf{More Qualitative Results} in 2-view setting with camera conditions.}
    \label{fig: more_2view_gen_w_camera}
\end{figure*}

\begin{figure*}[htbp]
    \centering

    \def\mywidth{2.0cm}
    \begin{tabular}{P{1.4cm}P{\mywidth}P{\mywidth}P{\mywidth}P{\mywidth}P{\mywidth}P{\mywidth}}

        \begin{minipage}[c]{0.18\textwidth}
            \includegraphics[height=0.49\linewidth]{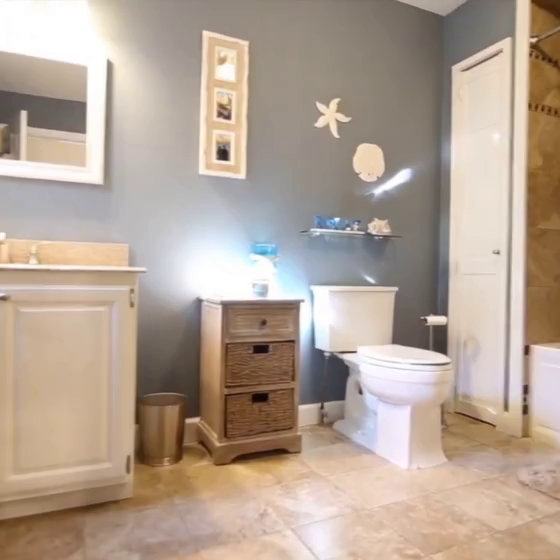}
        \end{minipage}
        &
        \includegraphics[width=1.13\linewidth]{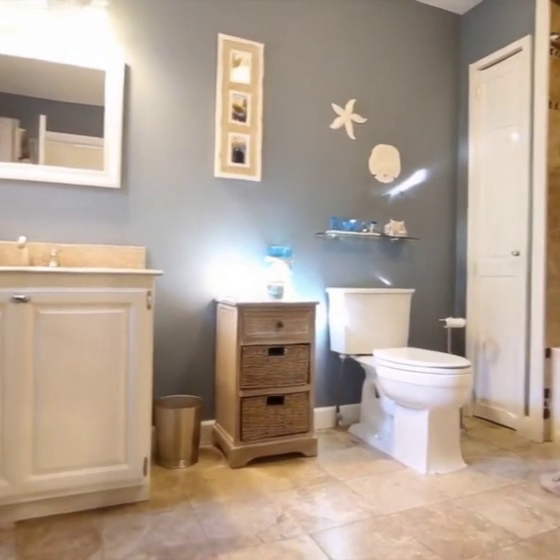} &
        \includegraphics[width=1.13\linewidth]{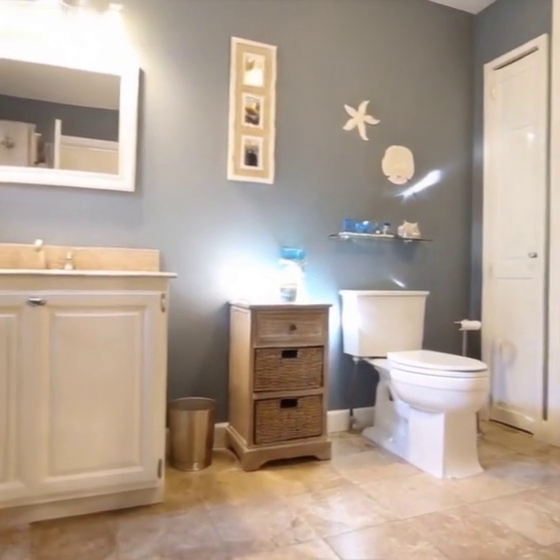} &
        \includegraphics[width=1.13\linewidth]{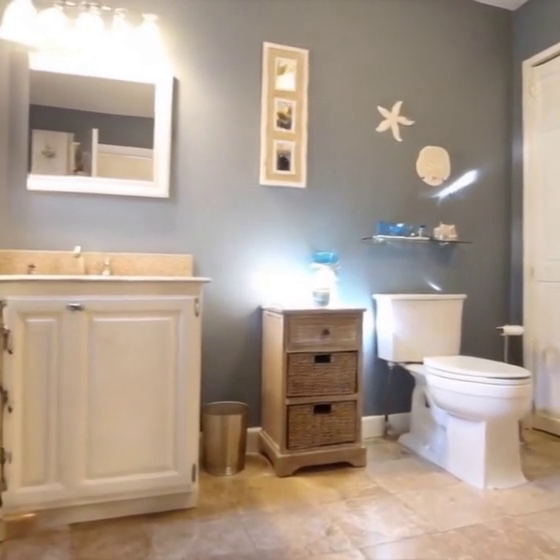} &
        \includegraphics[width=1.13\linewidth]{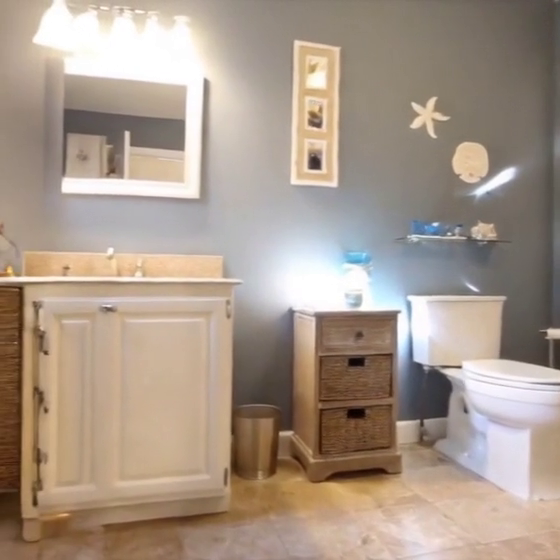} &
        \includegraphics[width=1.13\linewidth]{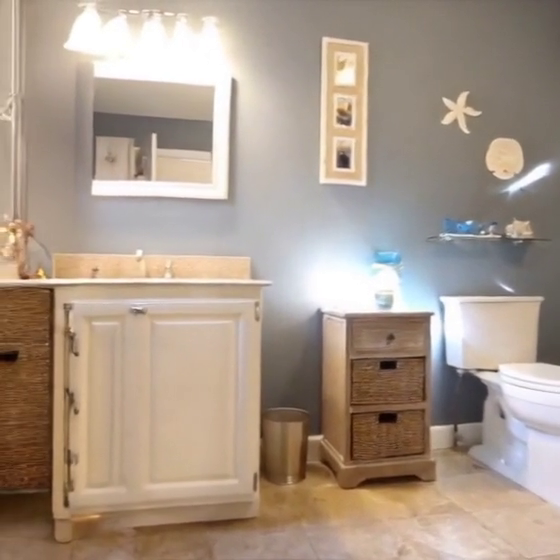} &
        \includegraphics[width=1.13\linewidth]{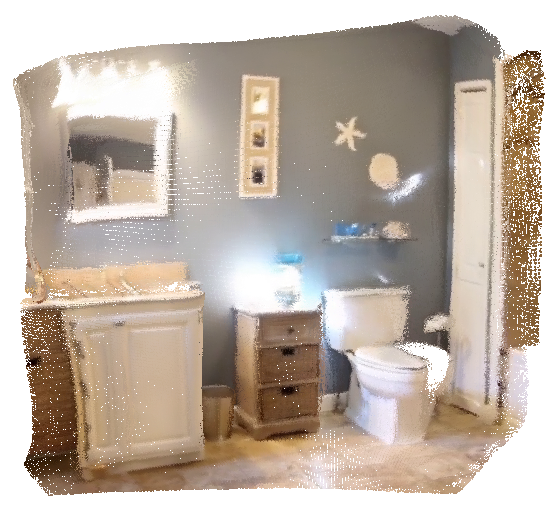}
        \\ [-2.5pt]

        \begin{minipage}[c]{0.18\textwidth}
            \includegraphics[height=0.49\linewidth]{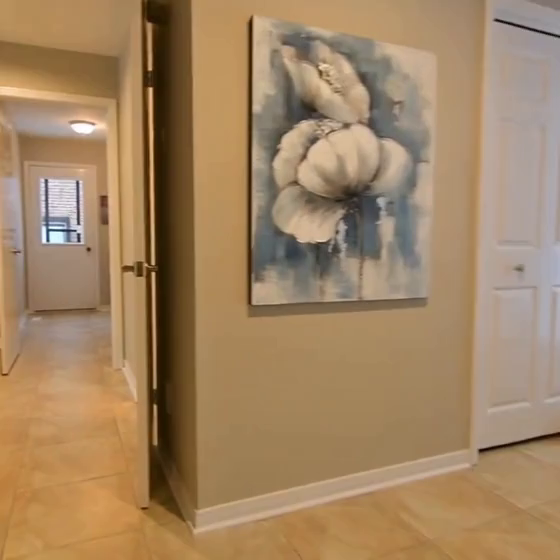}
        \end{minipage}
        &
        \includegraphics[width=1.13\linewidth]{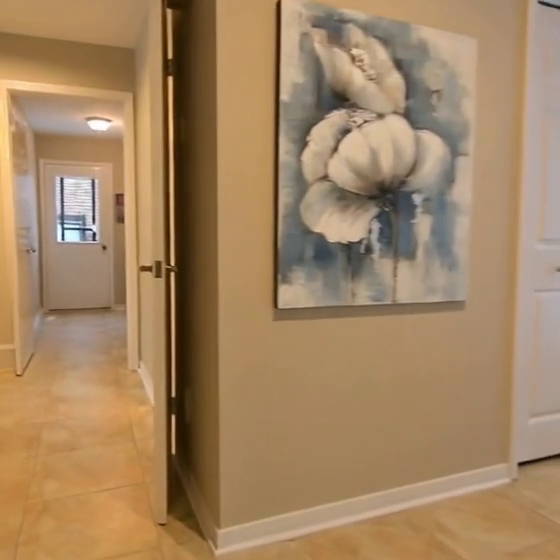} &
        \includegraphics[width=1.13\linewidth]{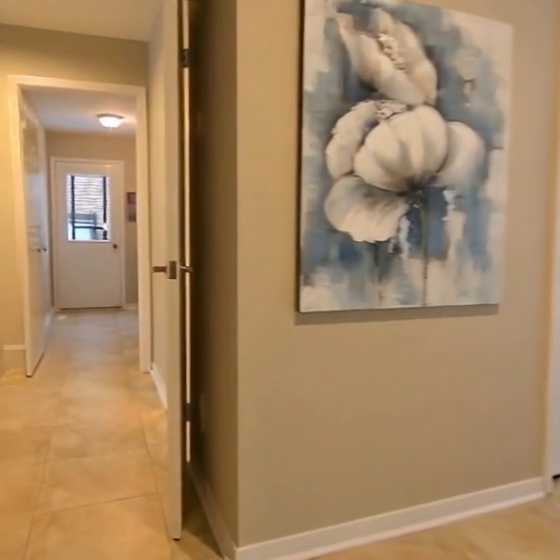} &
        \includegraphics[width=1.13\linewidth]{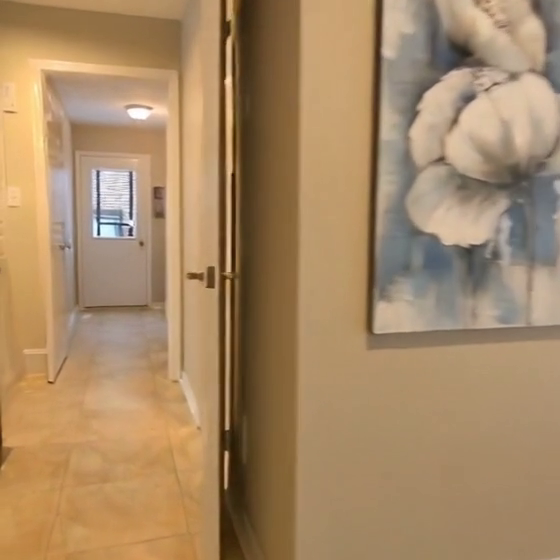} &
        \includegraphics[width=1.13\linewidth]{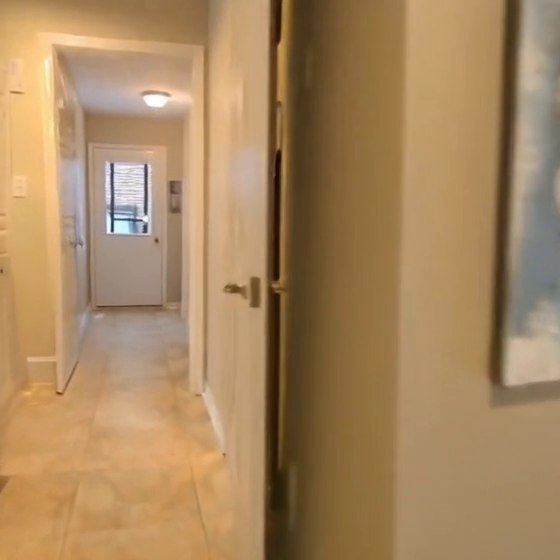} &
        \includegraphics[width=1.13\linewidth]{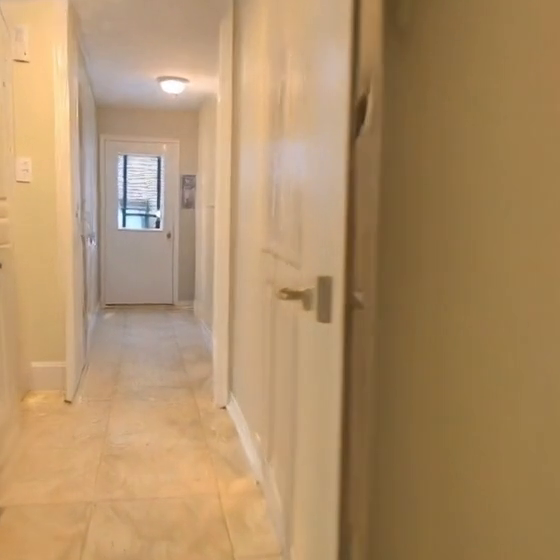} &
        \includegraphics[width=1.13\linewidth]{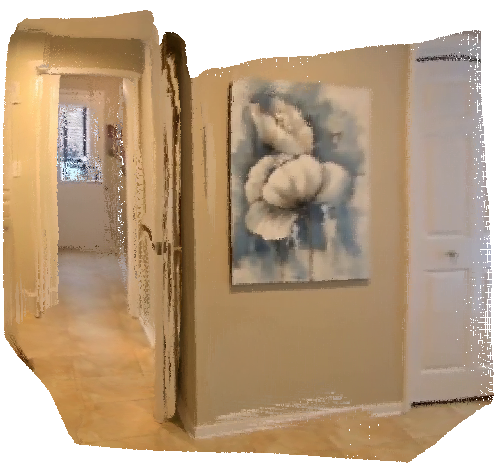}
        \\ [-2.5pt]

        \begin{minipage}[c]{0.18\textwidth}
            \includegraphics[height=0.25\linewidth]{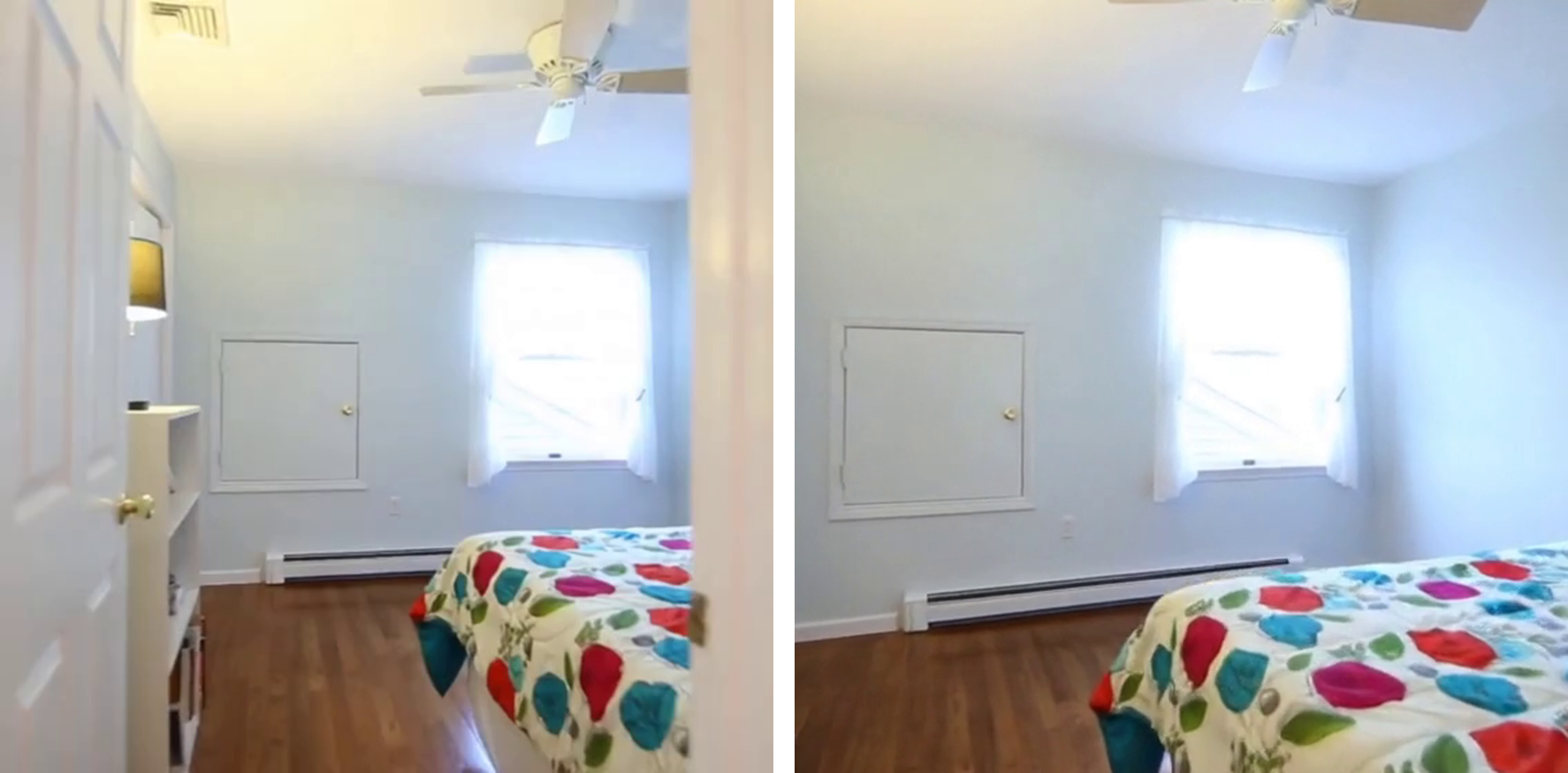}
        \end{minipage}
        &
        \includegraphics[width=1.13\linewidth]{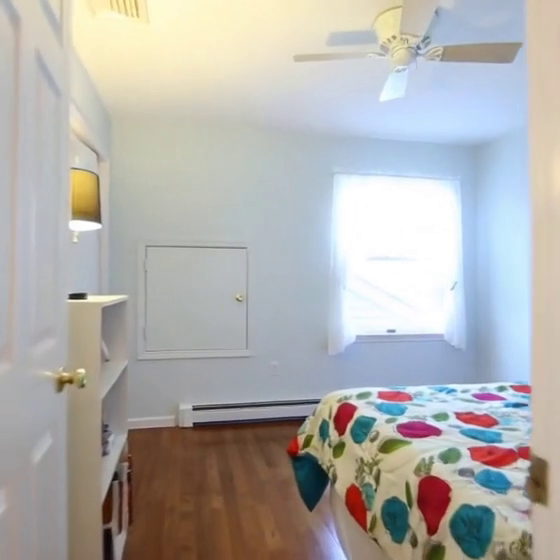} &
        \includegraphics[width=1.13\linewidth]{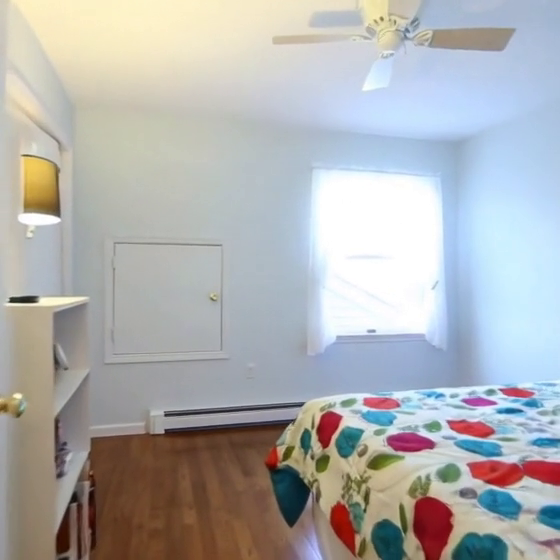} &
        \includegraphics[width=1.13\linewidth]{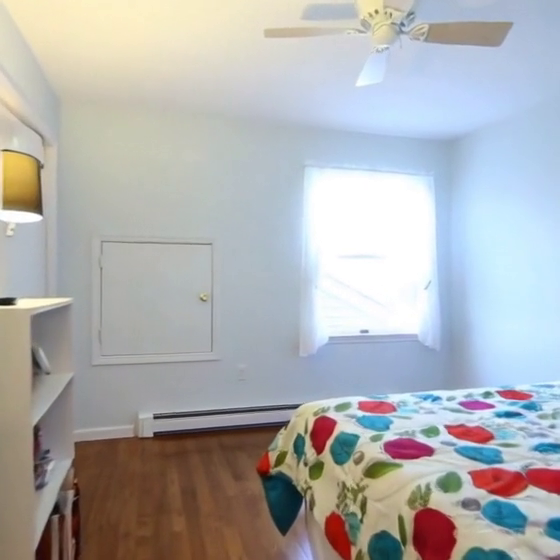} &
        \includegraphics[width=1.13\linewidth]{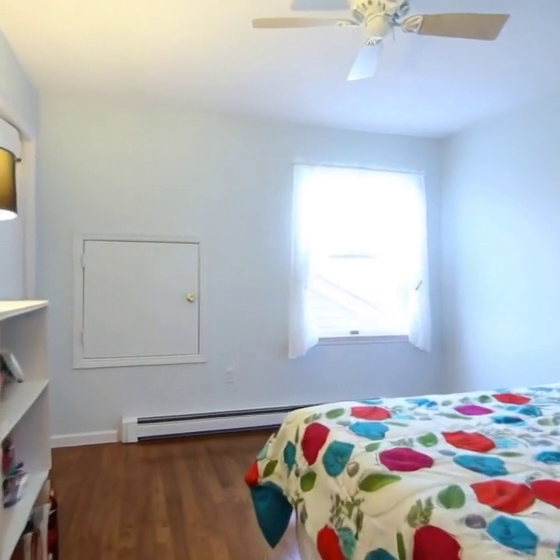} &
        \includegraphics[width=1.13\linewidth]{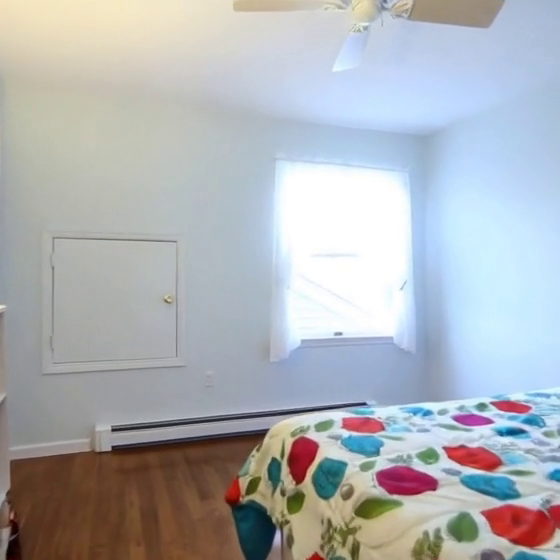} &
        \includegraphics[width=1.05\linewidth]{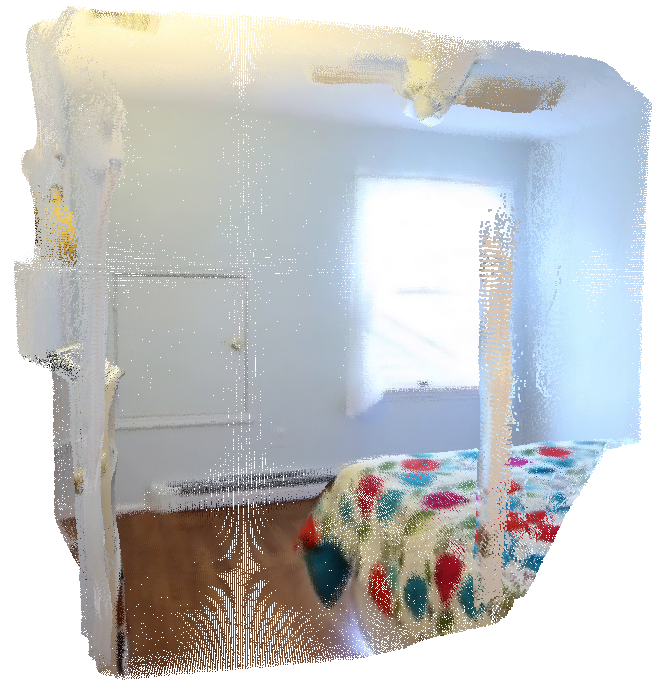}
        \\ [-2.5pt]

        \begin{minipage}[c]{0.18\textwidth}
            \includegraphics[height=0.25\linewidth]{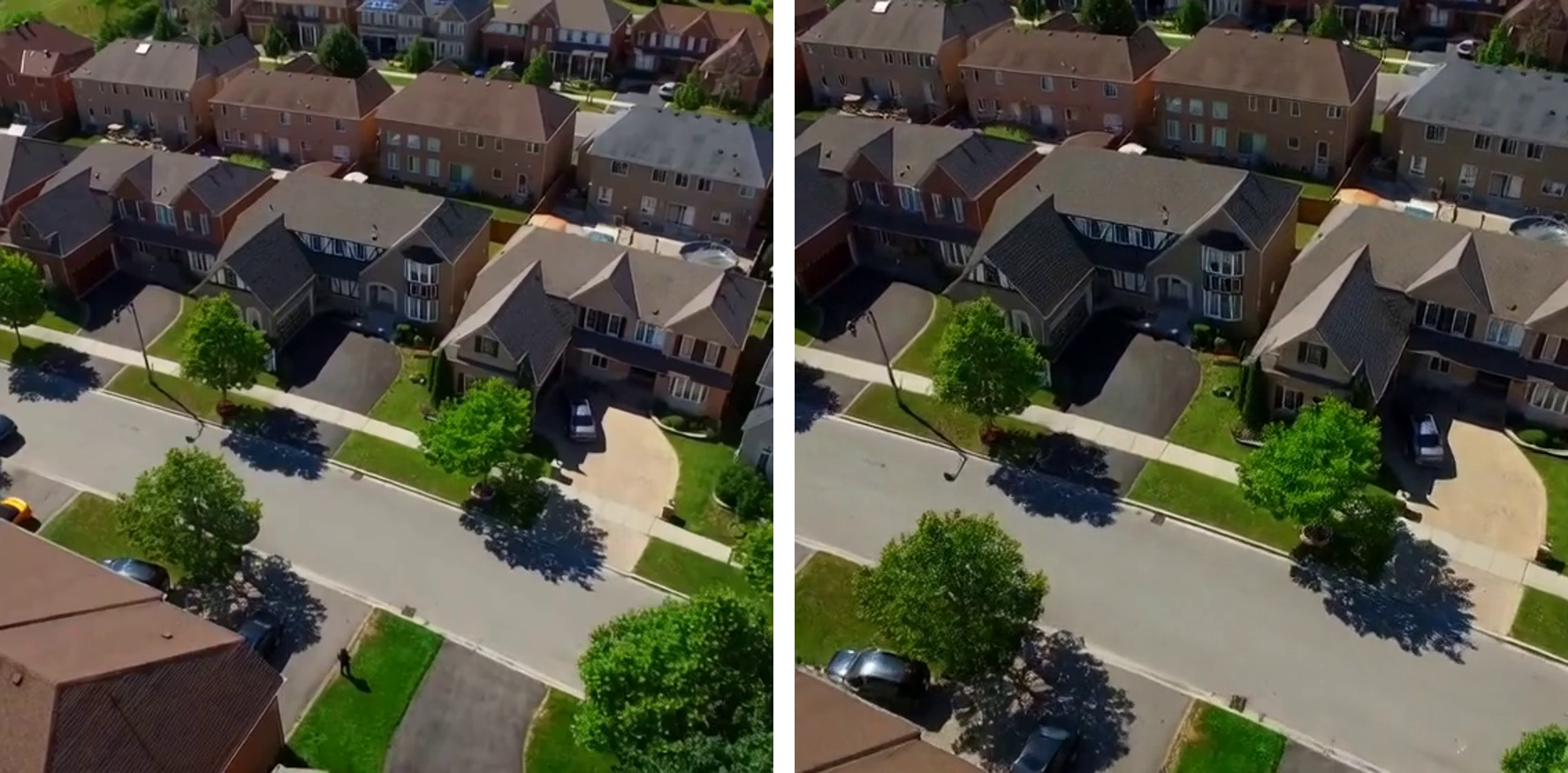}
        \end{minipage}
        &
        \includegraphics[width=1.13\linewidth]{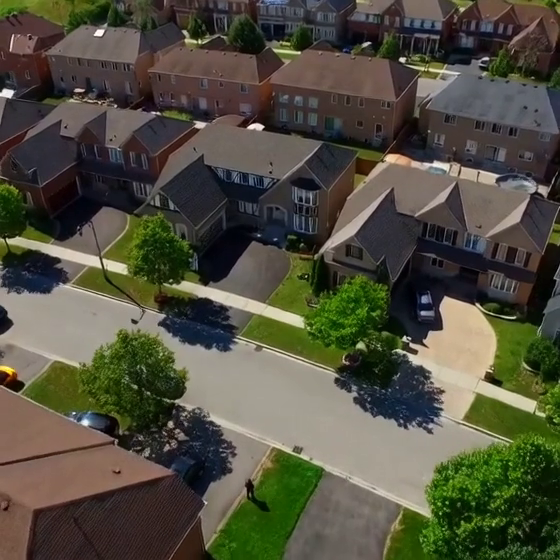} &
        \includegraphics[width=1.13\linewidth]{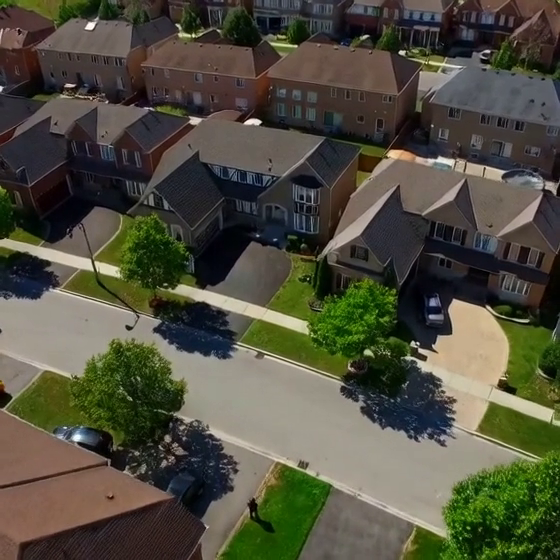} &
        \includegraphics[width=1.13\linewidth]{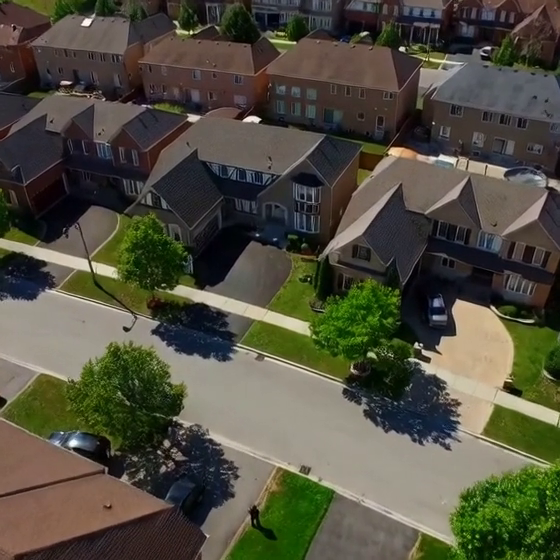} &
        \includegraphics[width=1.13\linewidth]{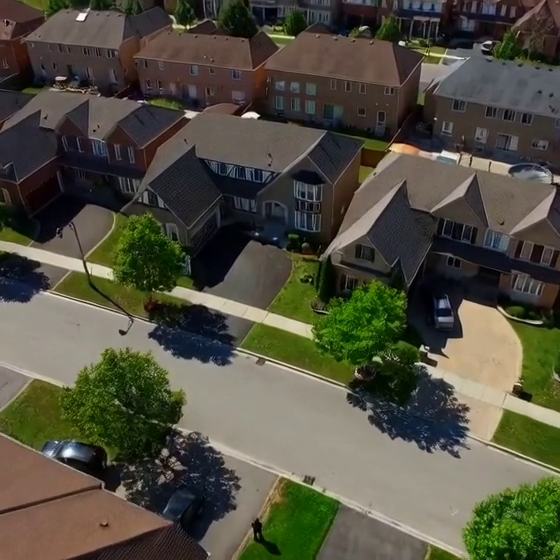} &
        \includegraphics[width=1.13\linewidth]{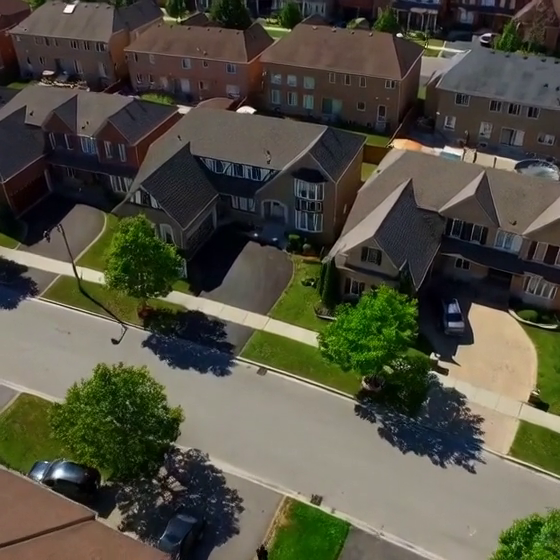} &
        \includegraphics[width=1.05\linewidth]{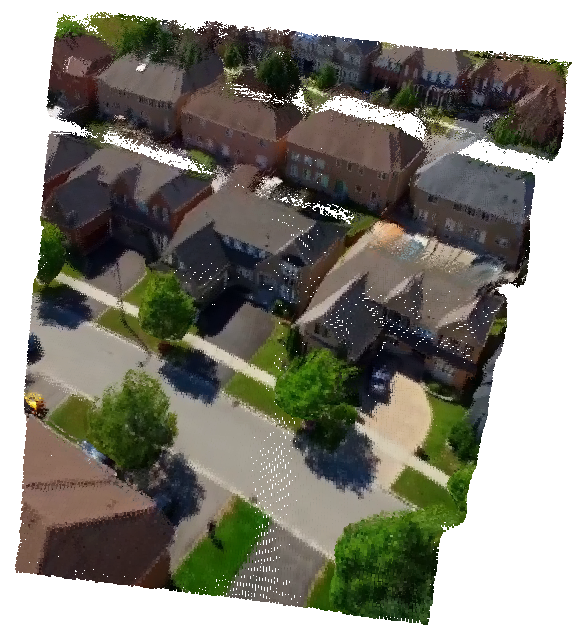}
        \\ [-2.5pt]

        \footnotesize Input &
        \multicolumn{5}{c}{\footnotesize Generated Frames} &
        \footnotesize Point Cloud
    \end{tabular}
    \vspace{-0.15cm}
    \caption{\textbf{More Qualitative Results} in 1-view and 2-view settings \textit{without} camera conditions.}
    \label{fig: more_gen_wo_camera}
    \vspace{-0.2cm}
\end{figure*}

\begin{figure*}[t]
    \centering

    \def\mywidth{2.0cm}
    \begin{tabular}{P{\mywidth}P{\mywidth}P{\mywidth}P{\mywidth}P{\mywidth}P{\mywidth}}

        \includegraphics[width=1.13\linewidth]{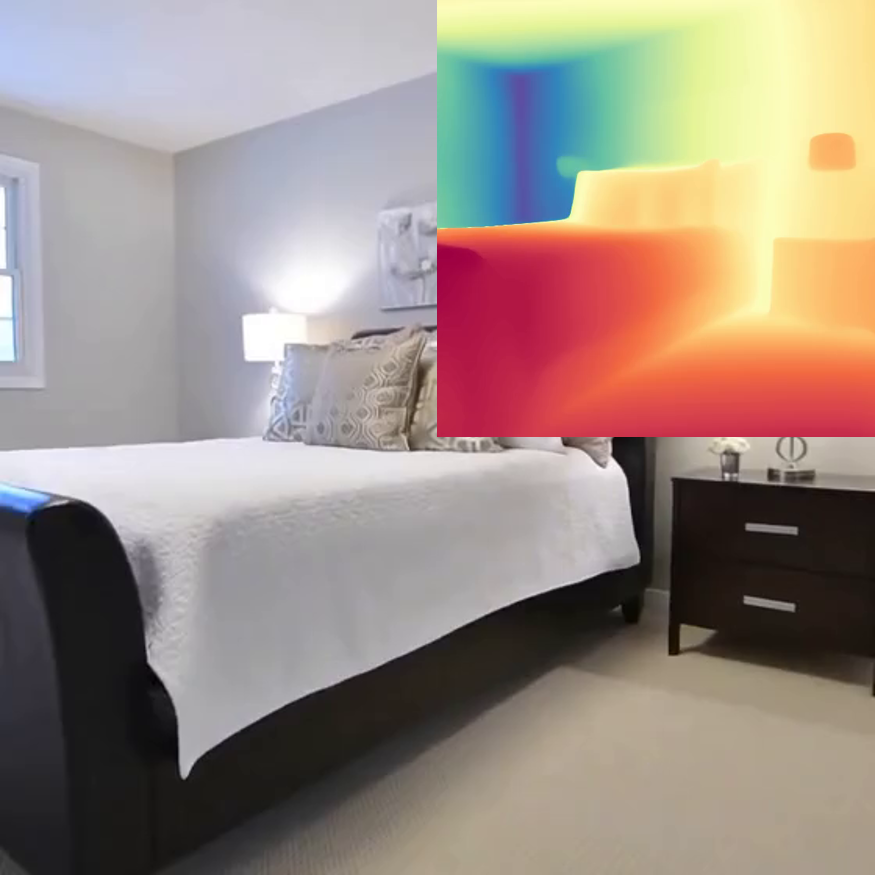} &
        \includegraphics[width=1.13\linewidth]{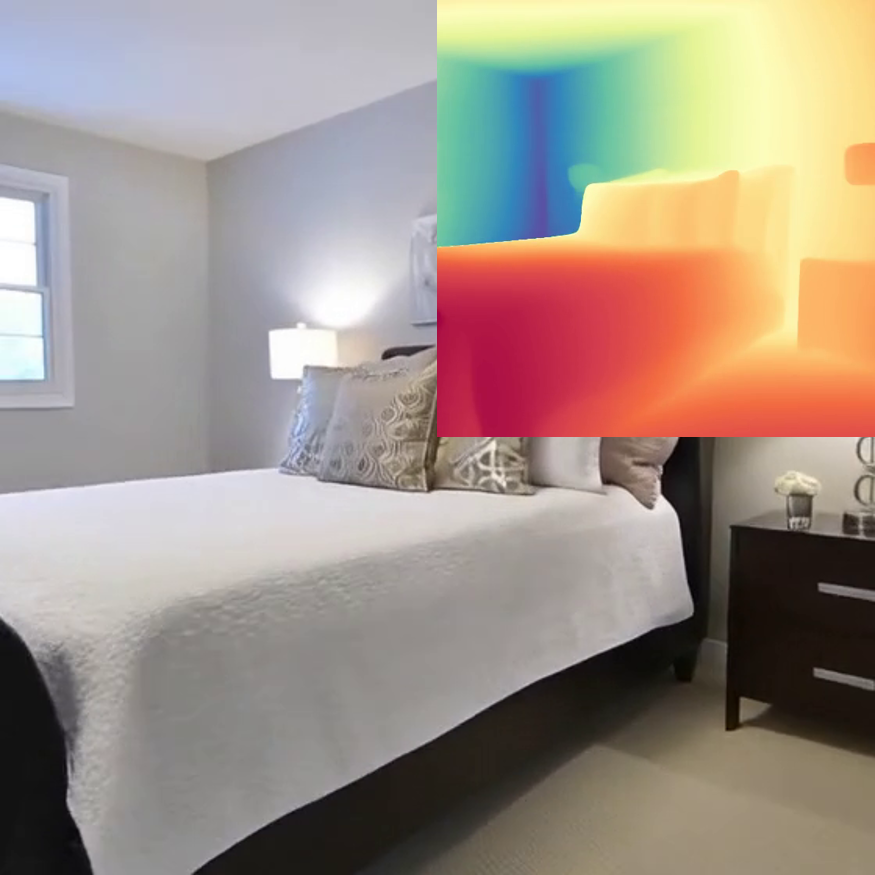} &
        \includegraphics[width=1.13\linewidth]{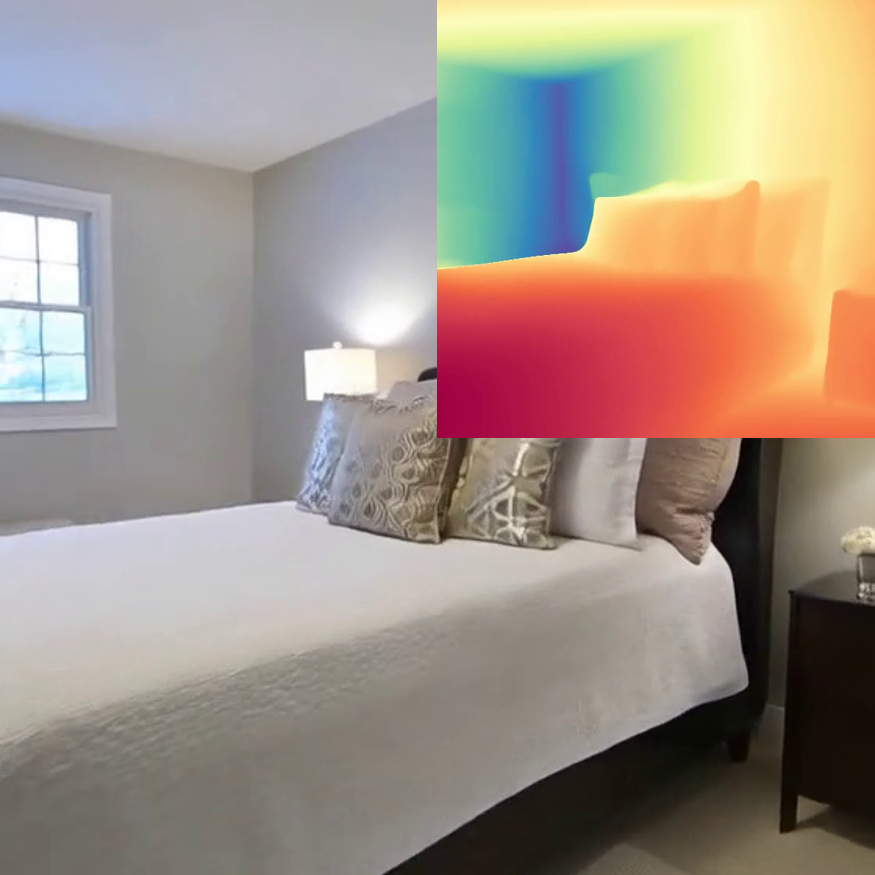} &
        \includegraphics[width=1.13\linewidth]{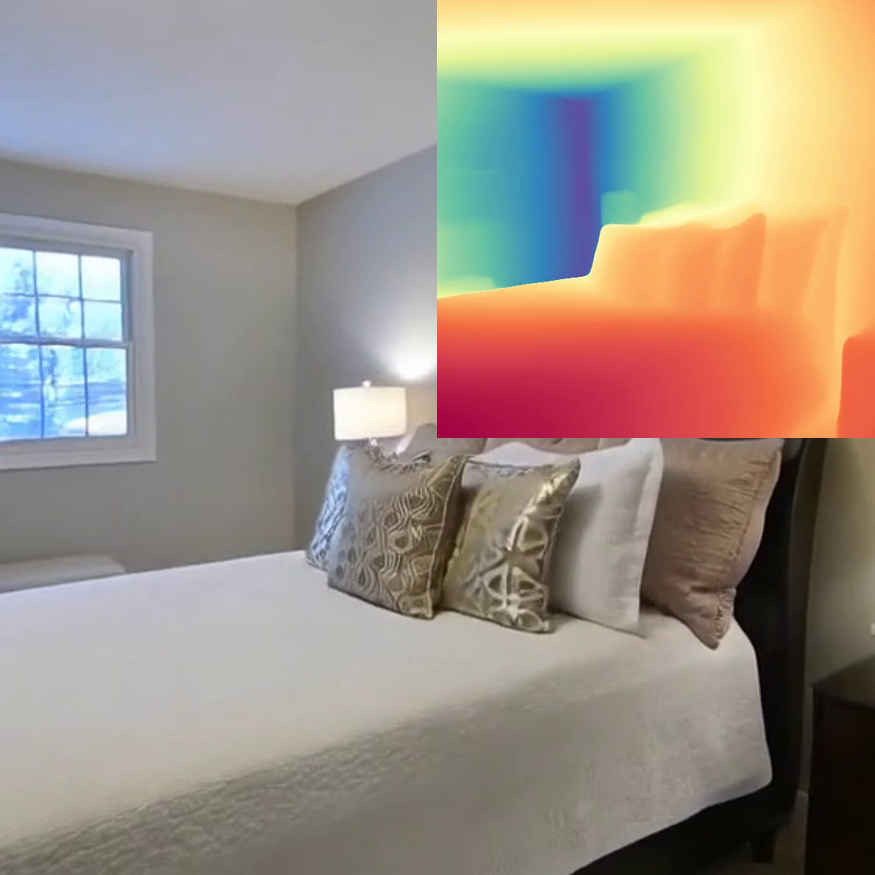} &
        \includegraphics[width=1.13\linewidth]{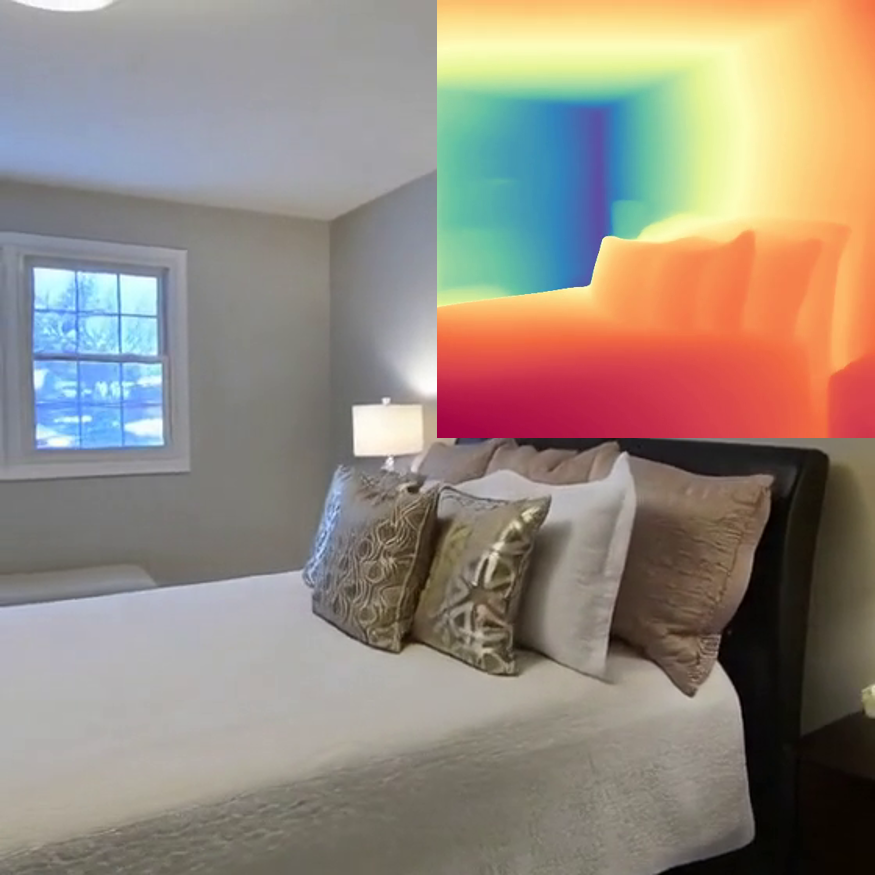} &
        \includegraphics[width=1.13\linewidth]{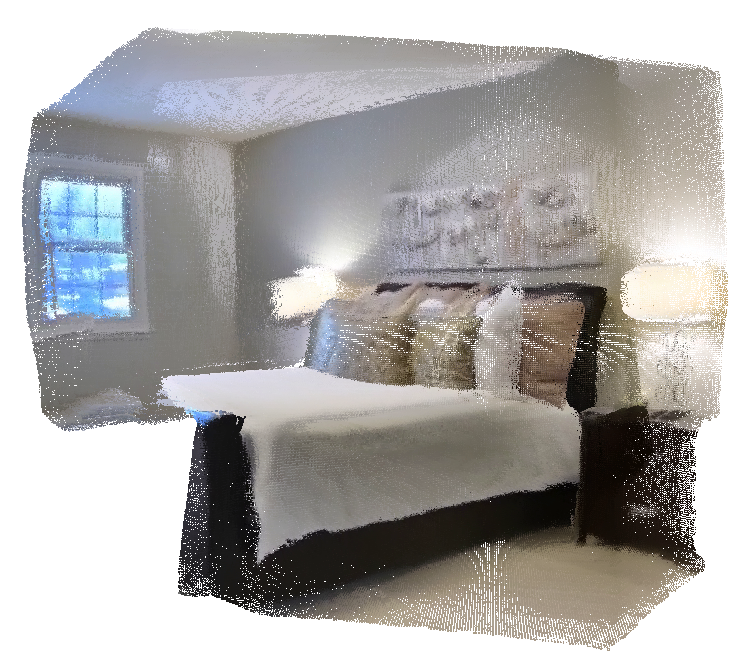}
        \\ [-2.5pt]

        \includegraphics[width=1.13\linewidth]{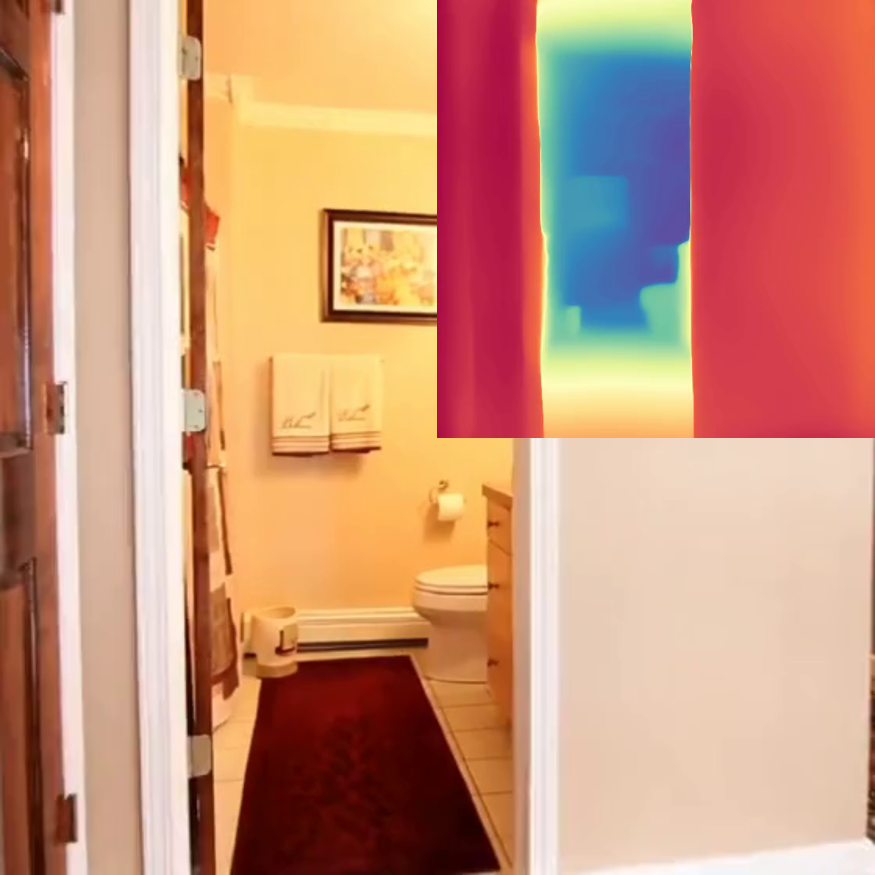} &
        \includegraphics[width=1.13\linewidth]{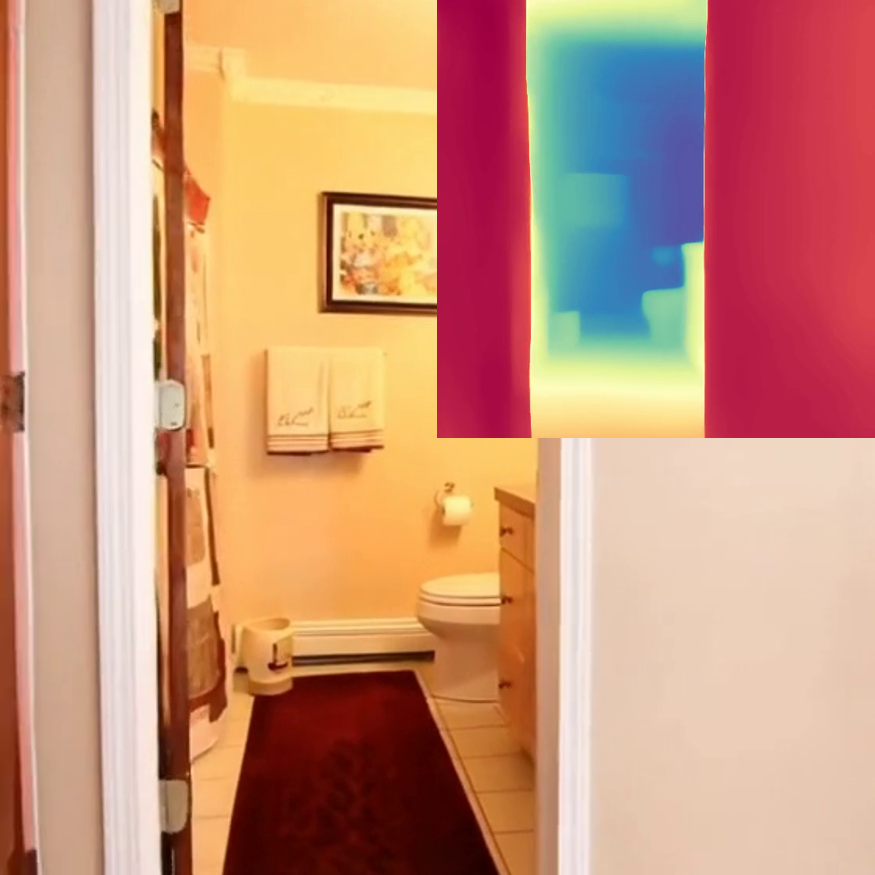} &
        \includegraphics[width=1.13\linewidth]{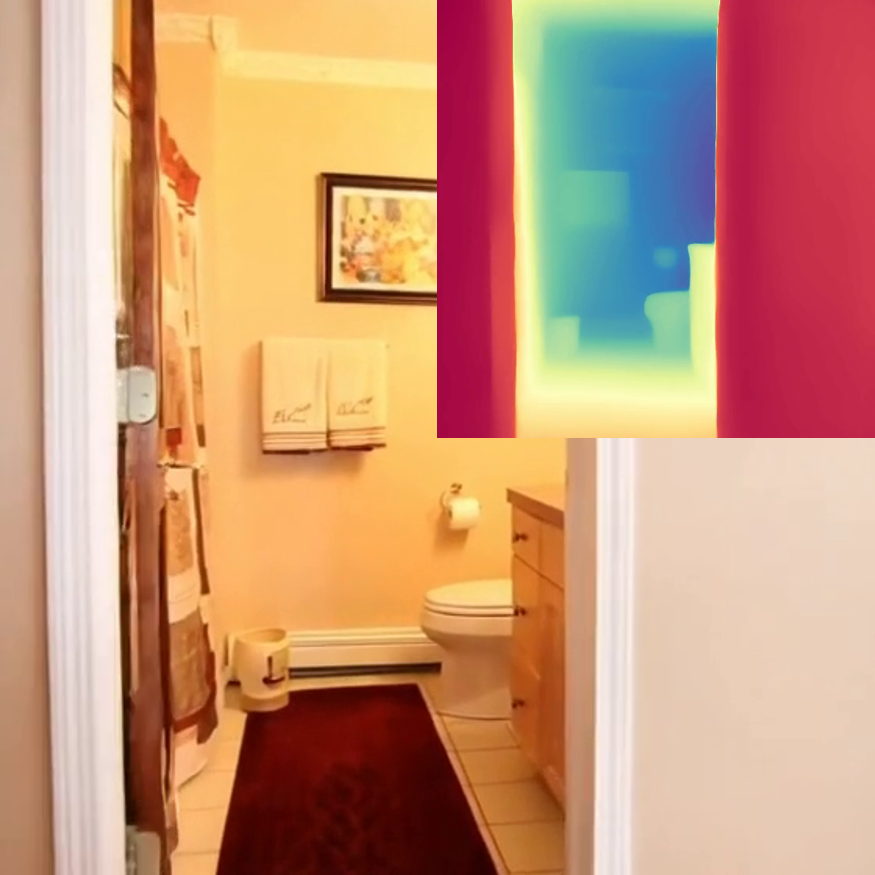} &
        \includegraphics[width=1.13\linewidth]{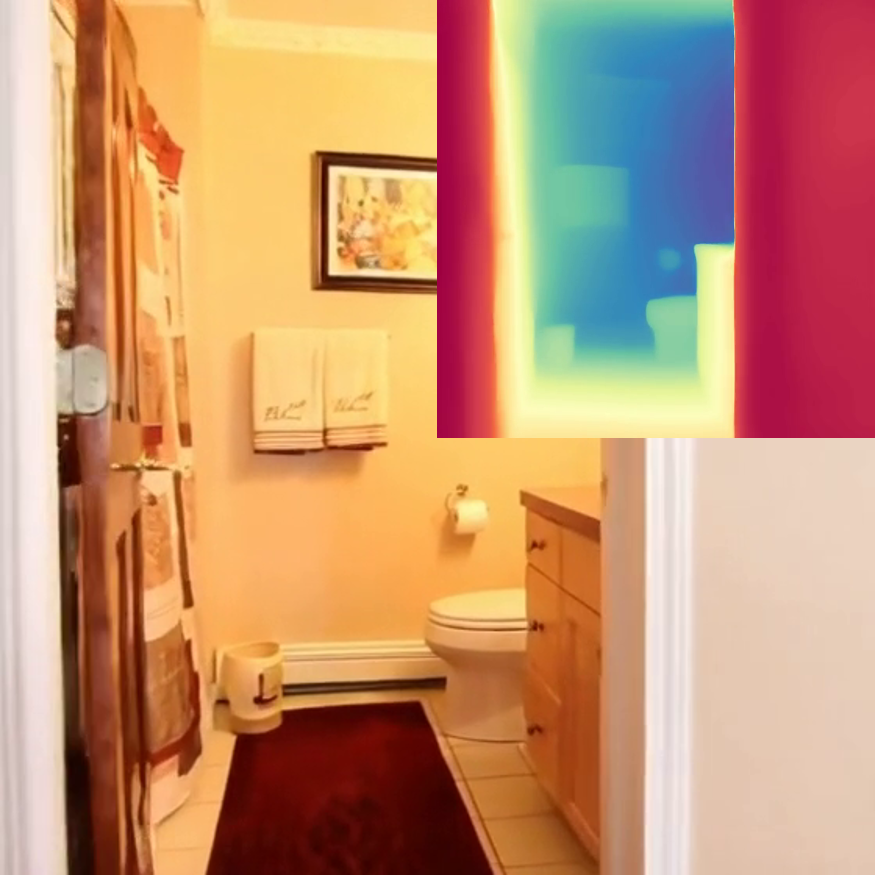} &
        \includegraphics[width=1.13\linewidth]{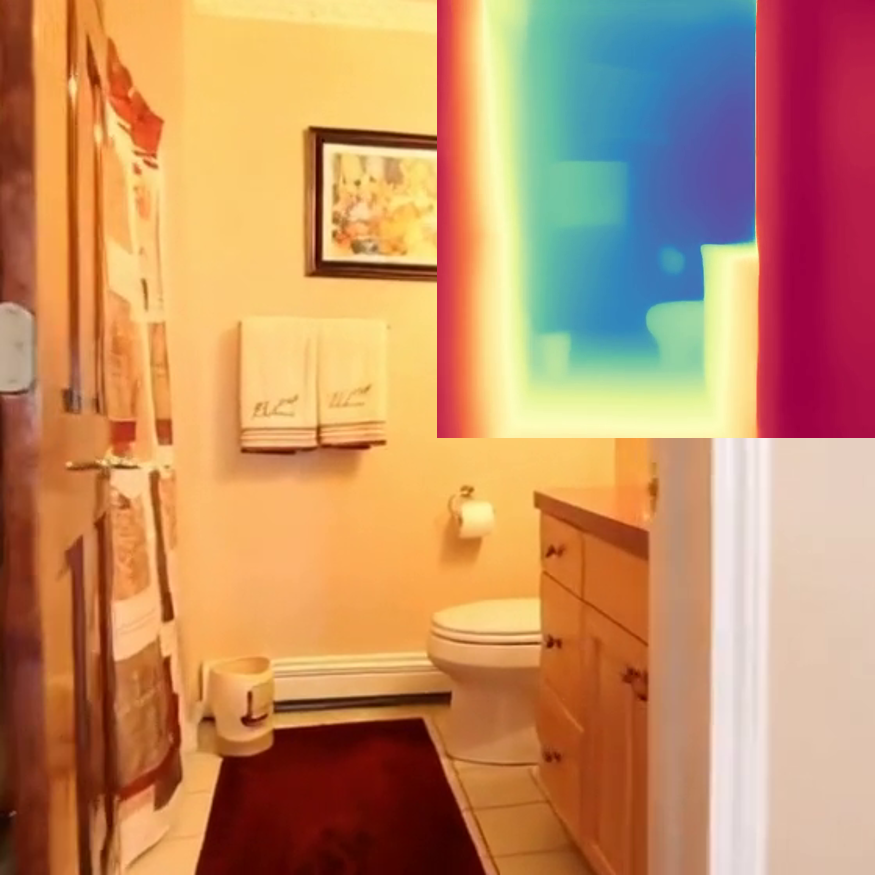} &
        \includegraphics[width=1.\linewidth]{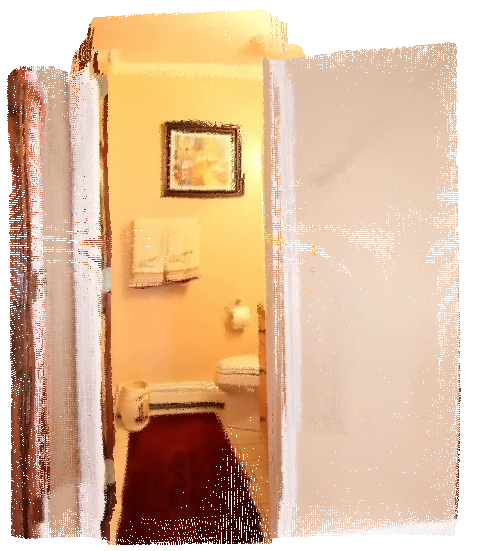}
        \\ [-2.5pt]

        \includegraphics[width=1.13\linewidth]{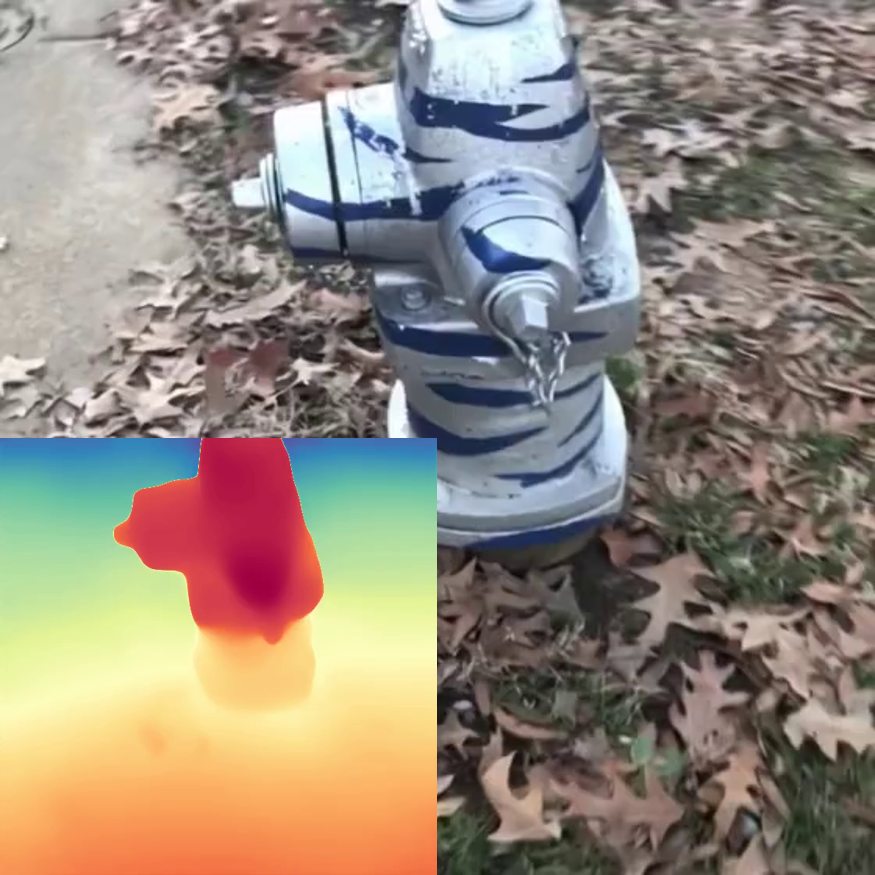} &
        \includegraphics[width=1.13\linewidth]{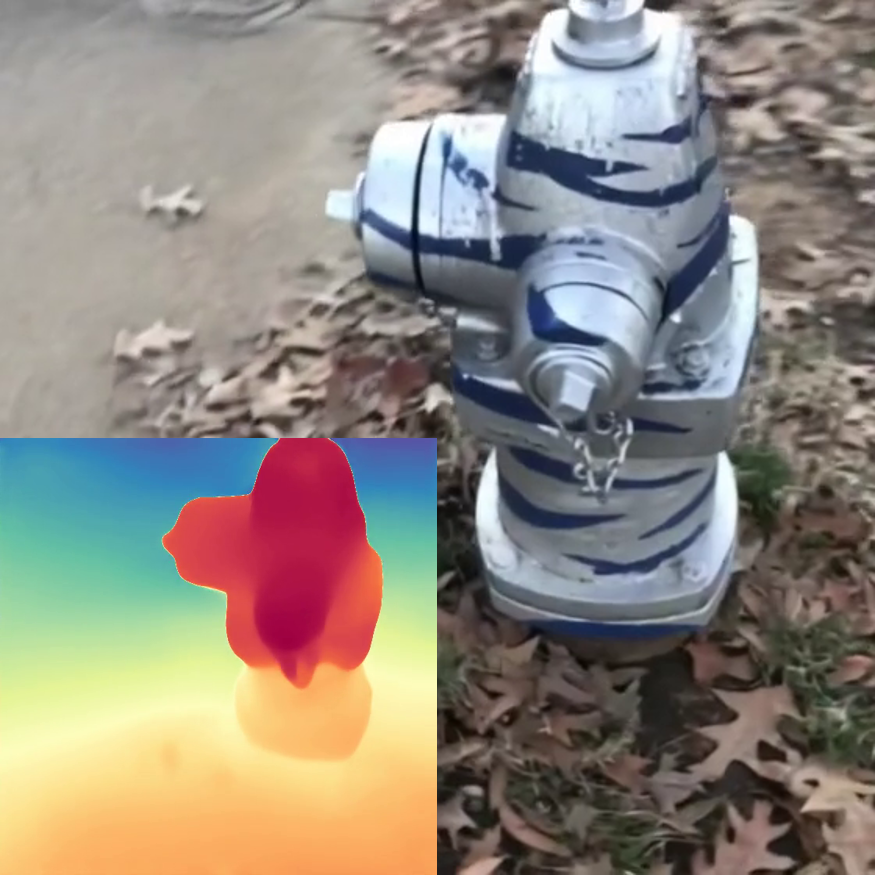} &
        \includegraphics[width=1.13\linewidth]{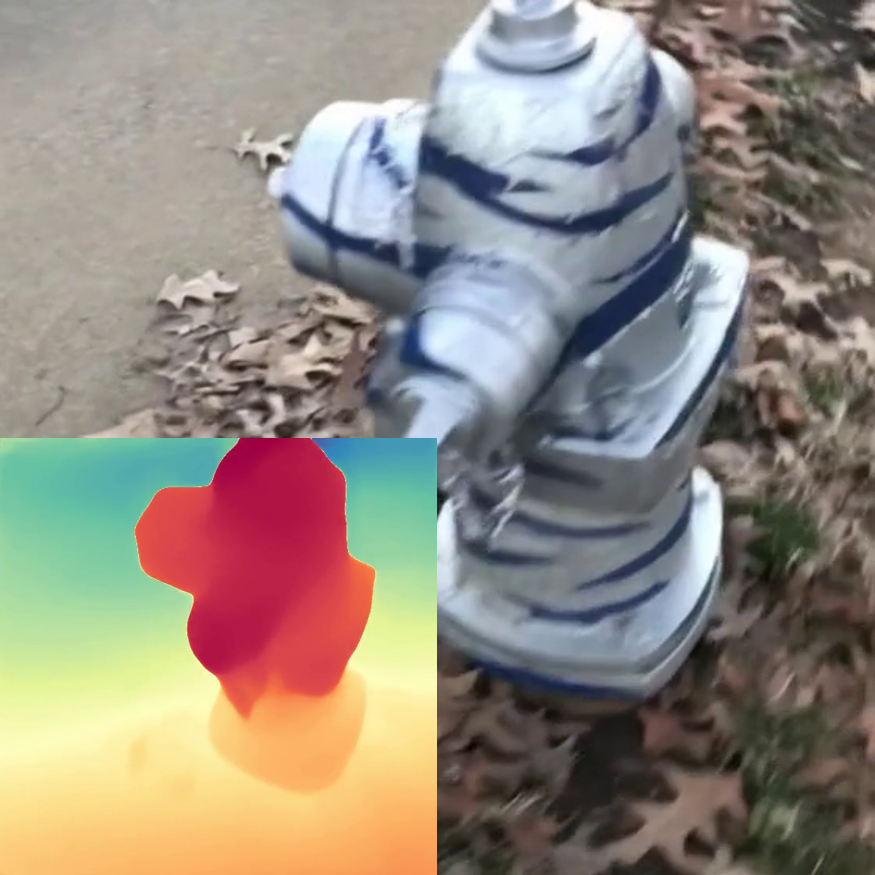} &
        \includegraphics[width=1.13\linewidth]{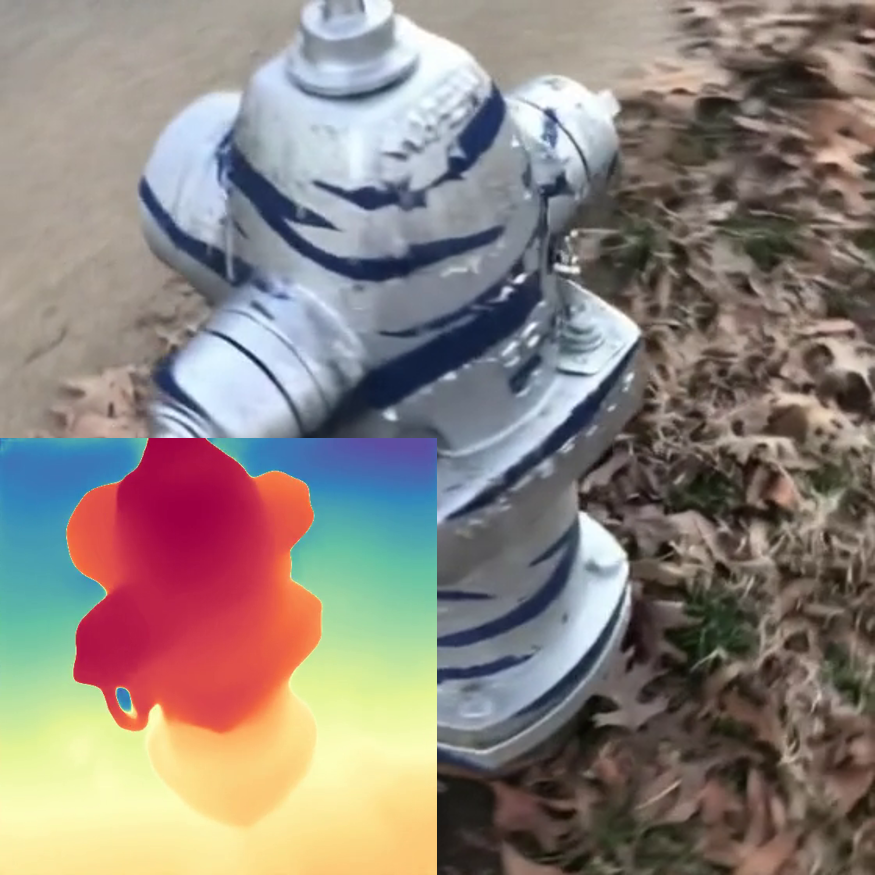} &
        \includegraphics[width=1.13\linewidth]{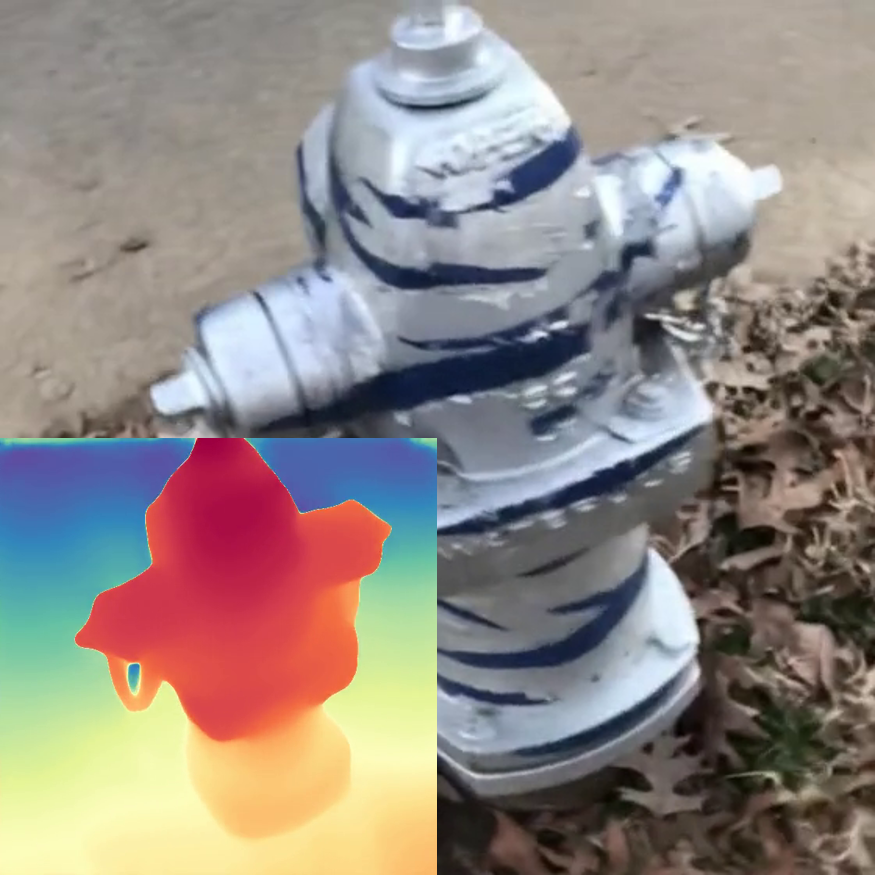} &
        \includegraphics[width=1.13\linewidth]{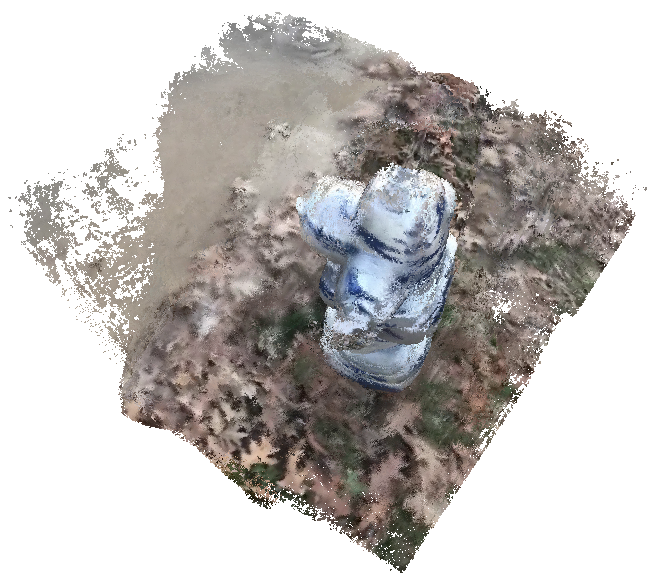}
        \\ [-2.5pt]

        \multicolumn{5}{c}{\footnotesize Generated Frames \& Depths} &
        \footnotesize Point Cloud
    \end{tabular}
    \vspace{-0.15cm}
    \caption{\textbf{More Qualitative Results} of feed-forward reconstruction.}
    \label{fig: more_reconstruction}
    \vspace{-0.25cm}
\end{figure*}

\section{Additional Comparison Results}
\label{sec: additional_results}

\subsection{3D Generation}

\begin{table}[t]
    \centering
    \footnotesize
    \begin{tabular}{lccc}
    \toprule
    \multirow{2}{*}{Method} & \multicolumn{3}{c}{ScanNet++} \\
    \cmidrule(lr){2-4}
     & Accuracy~$\downarrow$ & Completeness~$\downarrow$ & CD~$\downarrow$ \\
    \midrule
    Aether& \tbest{0.3187} & \tbest{0.3022} & \tbest{0.3105} \\
    VGGT  & \best{\textbf{0.1396}} & \sbest{0.1162} & \sbest{0.1279} \\
    Ours  & \sbest{0.1455} & \best{\textbf{0.0963}} & \best{\textbf{0.1209}} \\
    \bottomrule
    \end{tabular}
    \vspace{-0.2cm}
    \caption{\textbf{Quantitative Comparison of Zero-shot Geometry Reconstruction} in ScanNet++ dataset.}
    \label{tab: zero_shot_3d_recon}
\end{table}

\begin{table}[t]
    \centering
    \footnotesize
    \begin{adjustbox}{max width=\linewidth}
    \begin{tabular}{lcccccc}
    \toprule
    \multirow{2}{*}{Method} & \multicolumn{3}{c}{RealEstate10K} & \multicolumn{3}{c}{DL3DV-10K} \\
    \cmidrule(lr){2-4} \cmidrule(lr){5-7}
    & PSNR~$\uparrow$ & SSIM~$\uparrow$ & LPIPS~$\downarrow$ & PSNR~$\uparrow$ & SSIM~$\uparrow$ & LPIPS~$\downarrow$ \\
    \midrule
    VGGT\textsuperscript{*}~\cite{vggt} & 23.3927 & 0.8346 & 0.2341 & 22.6958 & 0.7557 & 0.2910 \\
    RGB VAE~\cite{wan} & 37.5770 & 0.9819 & 0.0288 & 32.7673 & 0.9057 & 0.1031 \\
    \bottomrule
    \end{tabular}
    \end{adjustbox}
    \vspace{-0.2cm}
    \caption{\textbf{Quantitative Comparison for RGB Reconstruction.} We train an RGB head for VGGT to reconstruct images from geometry tokens. * indicates our implementation.}
    \label{tab: ablation_rgb_decoder}
    \vspace{-0.25cm}
\end{table}

\boldparagraph{Comparison on 3D Generation with Camera Conditions}
We provide the full appearance evaluation results on Co3Dv2~\cite{co3d}, WildRGB-D~\cite{wildrgbd} and TartanAir~\cite{tartanair} datasets in~\cref{tab: supp_3d_generation_rgb}. \method{} consistently surpassing existing methods across all metrics and datasets in the 1-view setting, and achieves leading performance in the 2-view setting. Additional qualitative comparisons of 3D generation are shown in~\cref{fig: supp_single_to_3d} and~\cref{fig: supp_first_last_frame_to_3d}. As observed, LVSM~\cite{lvsm}, Aether~\cite{aether} and WVD~\cite{wvd} fail to synthesize images from novel viewpoint in 1-view setting, primarily due to poor camera controllability. While Gen3C~\cite{gen3c} can generate plausible contents, it exhibits notable shifts caused by inaccurate depth estimation. In contrast, our methed produces high-fidelity results that adhere closely to the camera conditions and maintain better 3D structure, as shown in~\cref{fig: supp_first_last_frame_to_3d}.

\boldparagraph{Comparison on 3D Generation without Camera Conditions}
We further demonstrate our capability to generate 3D scenes from images without camera conditions. To assess this, we report the VBench Score~\cite{vbench, vbenchpp}, focusing on I2V Subject~(I2V Subj.), I2V Background~(I2V BG), Aesthetic Quality~(Aes.Q.), Imaging Quality~(I.Q.) and Motion Smoothness~(M.S.) on RealEstate10K~\cite{re10k} and DL3DV-10K~\cite{dl3dv} datasets. As shown in~\cref{tab: supp_wo_camera}, our method clearly outperforms Aether~\cite{aether} and WVD~\cite{wvd}, illustrating its superior ability in generating high-quality 3D scenes.

\boldparagraph{Comparison on 3D Generation on Out-of-Distribution Datasets} 
We report appearance generation results in 1-view and 2-view settings on LLFF~\cite{llff}, Mip-NeRF 360~\cite{mipnerf360} and ScanNet++~\cite{scannetpp} test sets, which are excluded from the training datasets. As shown in~\cref{tab: supp_ood_3d_generation_rgb}, our method handles these unseen scenes well, and yields consistent conclusions with the main paper.

\subsection{Feed-forward 3D Reconstruction}
\boldparagraph{Comparison on Camera Pose Estimation}
We evaluate our method on RealEstate10K and WildRGB-D datasets for camera pose estimation, as reported in~\cref{tab: feed_forward_cam}. Our approach achieves competitive results compared to VGGT, while notably surpassing Aether, showing the versatility and robustness of our model.

\boldparagraph{Comparison on Geometry Reconstruction}
We provide additional qualitative results of feed-forward 3D reconstruction compared with VGGT~\cite{vggt} in~\cref{fig: supp_reconstruction}. It can be observed that VGGT produces noticeable floaters in the reconstructed point clouds, while our method generates significantly cleaner geometry.

In addition, to evaluate zero-shot generalization, we compare~\method{} with VGGT on ScanNet++~\cite{scannetpp} test split. As shown in~\cref{tab: zero_shot_3d_recon}, VGGT slightly outperforms \method~in accuracy; however, our method achieves better completeness and better chamfer distance, indicating that our method generalizes reasonably to unseen scenes. 

\subsection{Ablation Study}
\boldparagraph{RGB Head for VGGT}
To validate the effectiveness of our joint latents design, we train an RGB head for VGGT to enable direct RGB reconstruction from its geometry tokens \(\mathcal{V}\). We then compare its RGB reconstruction quality with that of Wan's RGB VAE~\cite{wan}. The results are presented in~\cref{tab: ablation_rgb_decoder}. RGB VAE significantly outperforms VGGT\textsuperscript{*}, as VGGT is designed primarily for geometry modeling and lacks sufficient capacity for RGB feature extraction and high-fidelity appearance reconstruction. This observation also motivates our choice to decode appearance and geometry separately. By combining the strengths of both pretrained models, we achieve photorealistic video generation together with high-quality 3D structure.

\section{More Results of \method{}}
\label{sec: more_results}
We present additional qualitative results for both 3D generation and feed-forward 3D reconstruction in this section, including: 
\textit{1) 3D Generation with Camera Conditions}~(see~\cref{fig: more_1view_gen_w_camera} and~\cref{fig: more_2view_gen_w_camera}); 
\textit{2) Feed-Forward 3D Reconstruction}~(see~\cref{fig: more_reconstruction}); 
and \textit{3) 3D Generation without Camera Conditions}~(see~\cref{fig: more_gen_wo_camera}).
We visualize the generated frames, depth maps of the sequences, and the global point clouds of the scenes. 

Our method synthesizes globally consistent and photorealistic 3D scenes under diverse input conditions and effectively handles a wide range of scenarios, including indoor scenes, outdoor environments, and object-centric cases. Thanks to our design, the model exhibits strong camera controllability under conditioned settings, while also enabling free scene navigation in the absence of camera inputs. Combined with support for multiple output modalities, \method{} provides fine-grained and coherent 3D scene generation across both constrained and unconstrained regimes.

\end{document}